\documentclass[acmtog, nonacm]{acmart}

\AtBeginDocument{%
  }

\makeatletter
\def\@authorsaddresses{}
\makeatother

\usepackage{xcolor}
\usepackage{tabularx}
\usepackage{array}

\definecolor{darkgreen}{RGB}{0,100,0}

\begin{document}

\title{\centering Semantic Browsing: Controllable Diversity for Image Generation}
\subtitle{%
\centerline{Sara Dorfman$^{*}$, Maya Vishnevsky$^{*}$, Omer Dahary, Or Patashnik, Daniel Cohen-Or}
\centerline{Tel Aviv University}
}

\begin{abstract}

Modern text-to-image models excel in visual fidelity and prompt adherence. However, this strict adherence comes at the cost of diversity: generated samples tend to collapse into a single visual interpretation. Existing methods to improve diversity produce outputs driven by incidental variations rather than meaningful design choices. This motivates a new variant of the diversity task where structure is enforced on the generated samples.

We introduce a method for controlled diversity that enables \textit{Semantic Browsing}, where users can navigate structured image galleries and experience creative exploration through a systematic traversal of meaningful, interpretable axes of variation.
Achieving this level of semantic control requires a deep understanding of the scene. We exploit the fact that recent text-to-image models are trained on elaborated captions, effectively decoupling semantic decision-making from pixel generation. This enables a paradigm shift: instead of relying on stochastic variation within the text-to-image model, we induce diversity directly at the text level. By leveraging rich textual representations, we allow a Vision Language Model (VLM) to operate on the full scene context. To overcome the generic outputs typical of standard VLMs, we employ an \emph{agentic workflow} that explicitly enforces structured variation attuned to the original prompt. We demonstrate that our method produces diverse and navigable design spaces where every variation corresponds to a specific, user-understandable semantic decision. \textit{Project page: \url{https://saradorfman1.github.io/SemanticBrowsing-webpage/}}

\end{abstract}

\begin{teaserfigure}
  \includegraphics[width=\textwidth]{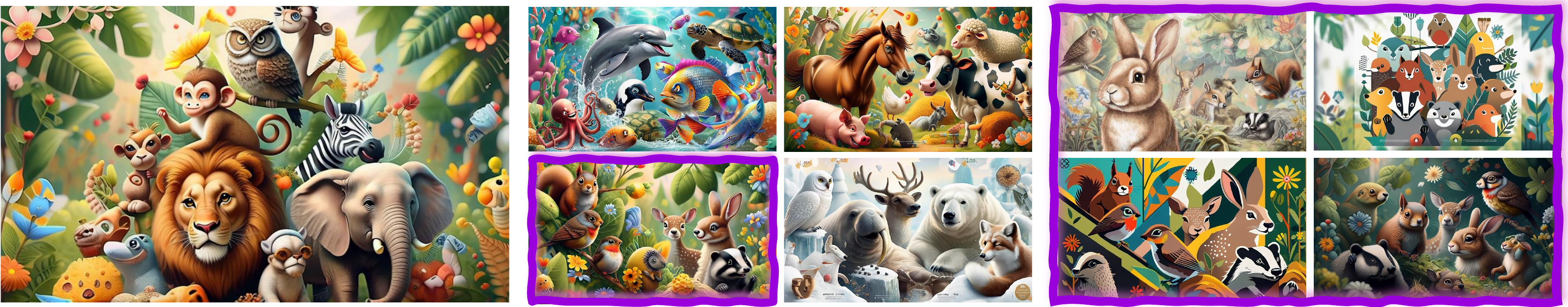}
  \caption{
\textbf{Semantic Browsing for Image Generation.}
From a single text prompt \textit{``A poster featuring animals''}, the system produces a structured gallery of images that explore different meaningful interpretations of the same scene. Rather than random variations, each image reflects a distinct, coherent semantic choice (e.g., changes in character, composition, or style) allowing users to browse a space of alternatives in a deliberate and interpretable way.
In this visualization, the leftmost image serves as the root for the four variations in the center. The variation highlighted with a purple border is then selected as the specific parent for its four children displayed on the right.}
  \label{fig:teaser}
\end{teaserfigure}

\maketitle
\begingroup
\renewcommand{\thefootnote}{*}
\footnotetext{Indicates equal contribution.}
\endgroup

\section{Introduction}

Advancements in generative image models have rapidly transformed the way visual content is created, edited, and explored~\cite{rombach2022highresolution, ho2020denoising, dhariwal2021diffusion}.
Much of the progress in these models has focused on visual fidelity and adherence to input conditioning.
However, as these models become more capable, user expectations have expanded: rather than seeking a single faithful rendering, users often wish to explore multiple plausible outputs, particularly when their desired outcome is still unclear.
This change in user expectations raises the challenge of generating a diverse gallery of outputs from a single input prompt.

\begin{figure}[t]
    \centering
    \setlength{\tabcolsep}{1pt} 
    
    \renewcommand{\arraystretch}{0} 
    
    {\small
    \begin{tabular}{c c c c c}
        \raisebox{2.5em}{\rotatebox{90}{Ours}} & 
        \includegraphics[width=0.22\linewidth]{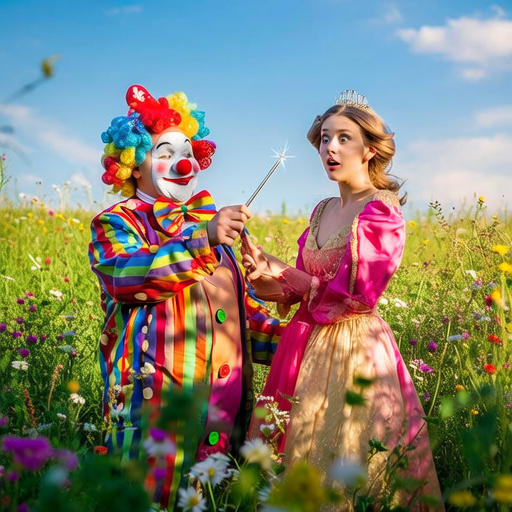} &
        \includegraphics[width=0.22\linewidth]{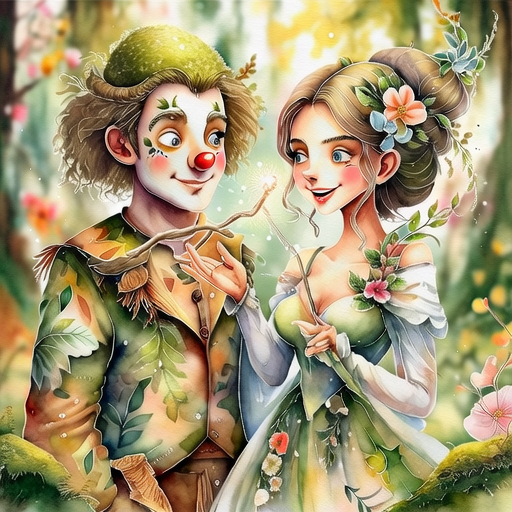} &
        \includegraphics[width=0.22\linewidth]{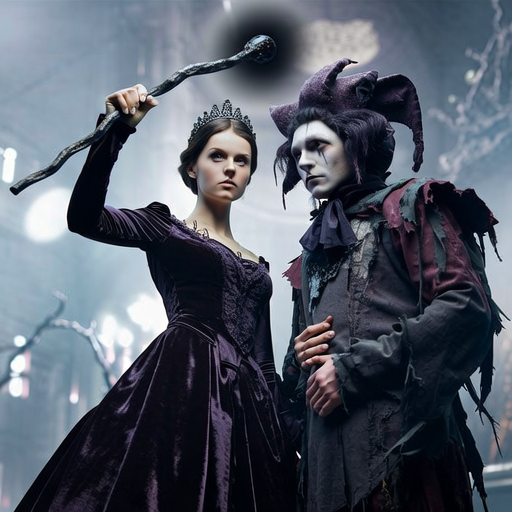} &
        \includegraphics[width=0.22\linewidth]{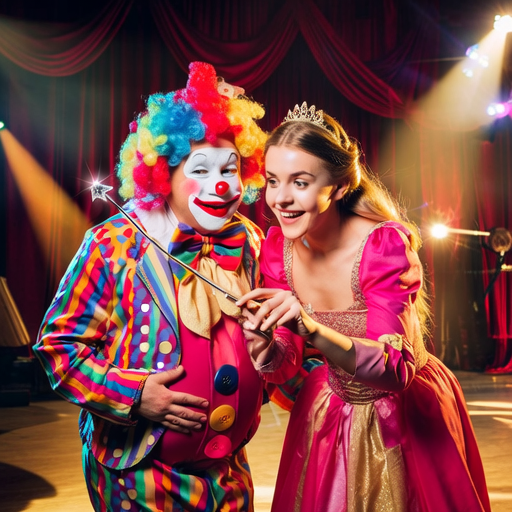} \\[-1pt] 

        \rotatebox{90}{\parbox{2cm}{\centering Standard\\ Sampling}} &
        \includegraphics[width=0.22\linewidth]{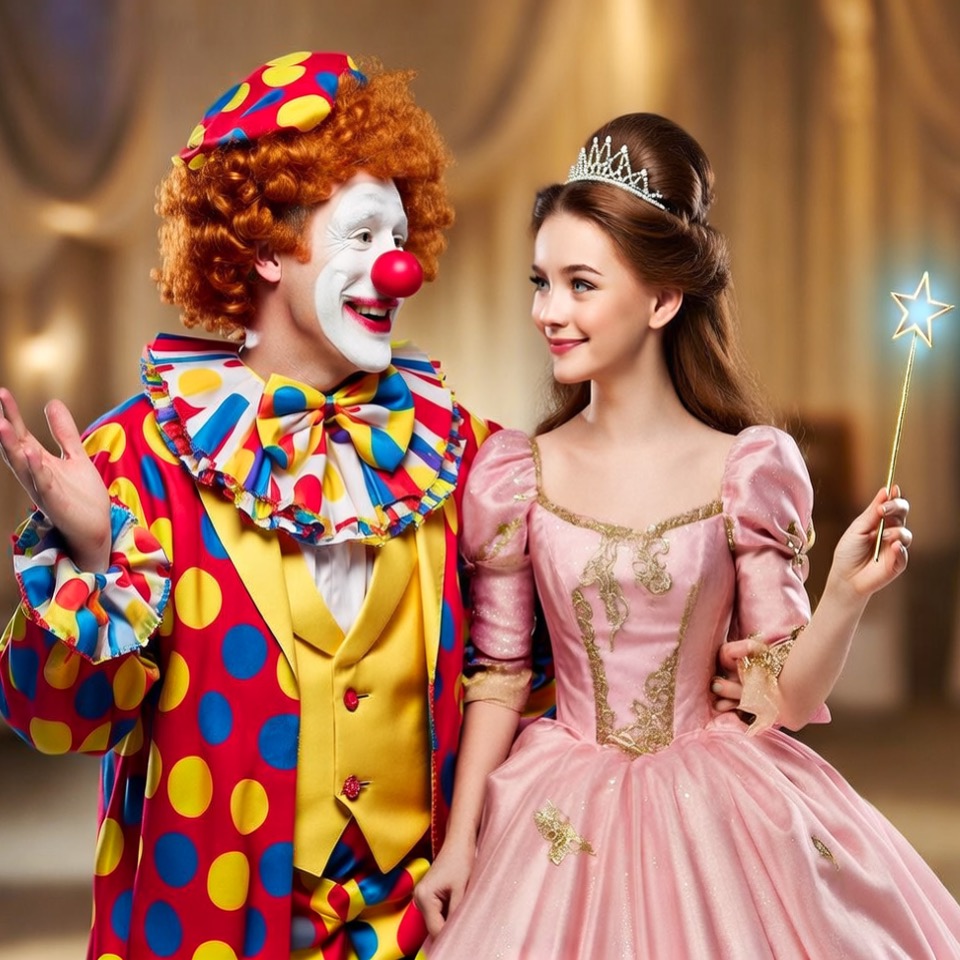} &
        \includegraphics[width=0.22\linewidth]{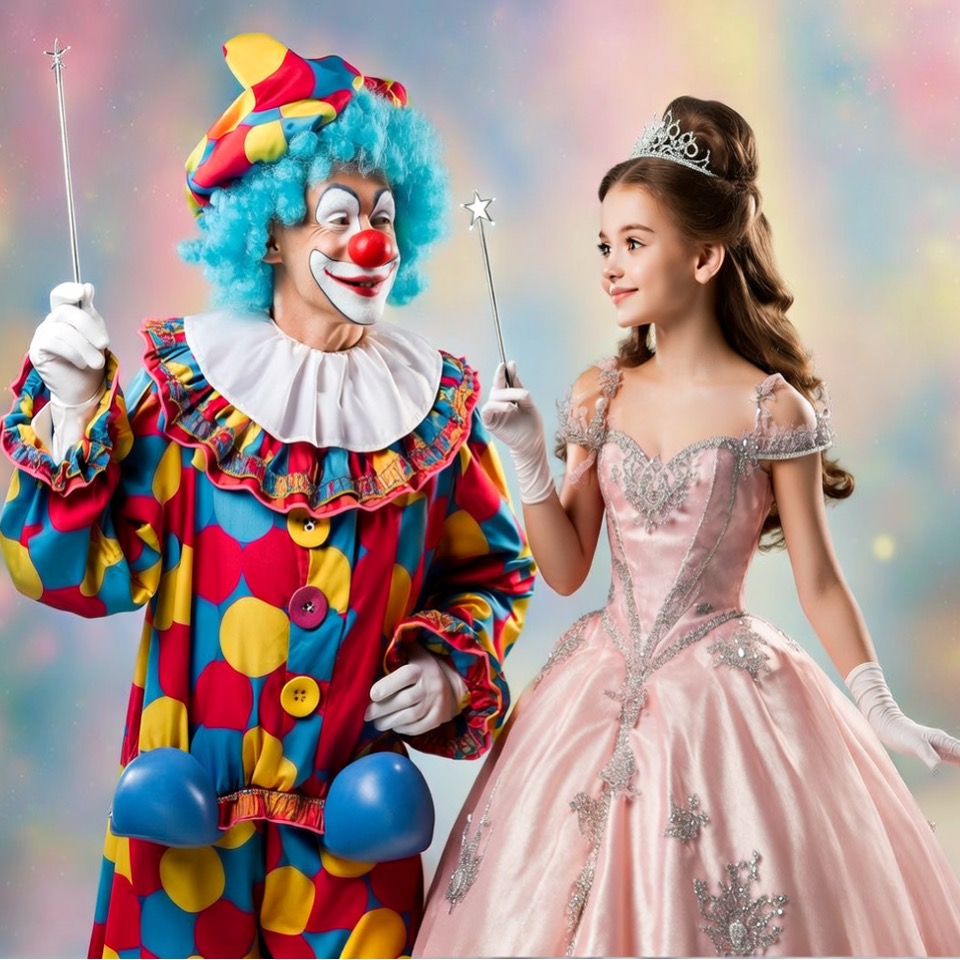} &
        \includegraphics[width=0.22\linewidth]{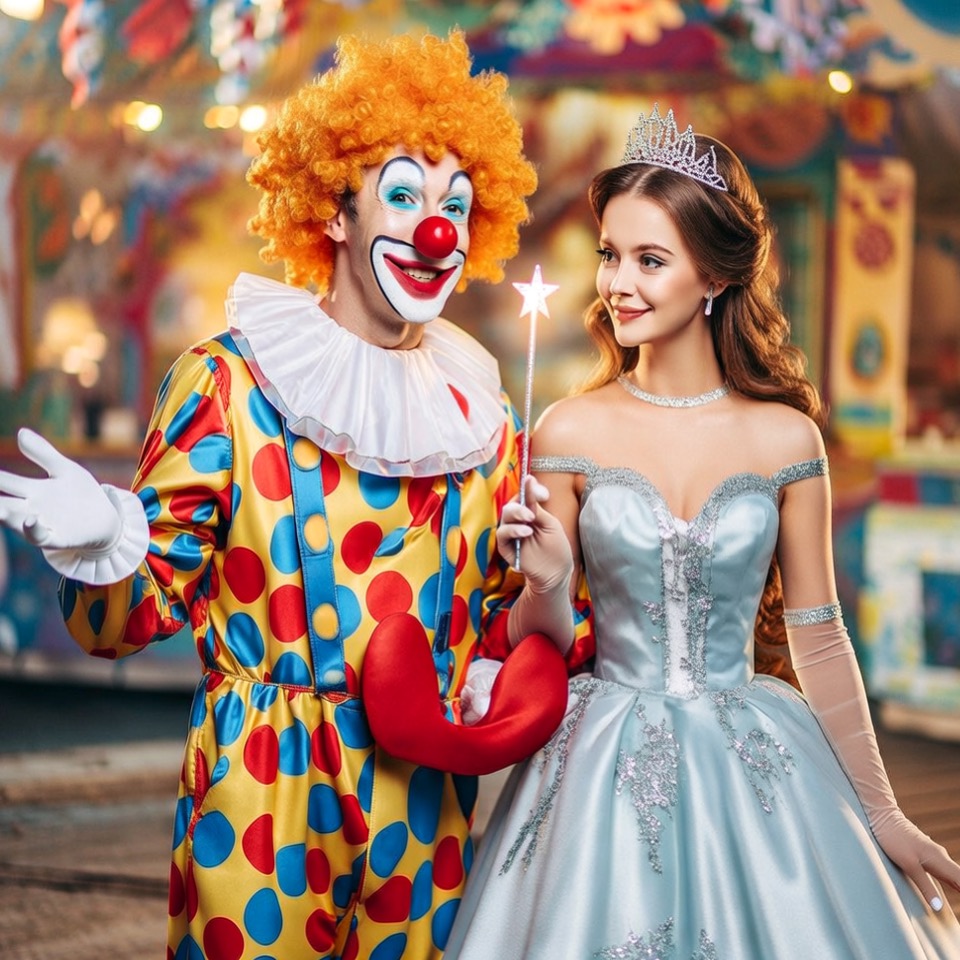} &
        \includegraphics[width=0.22\linewidth]{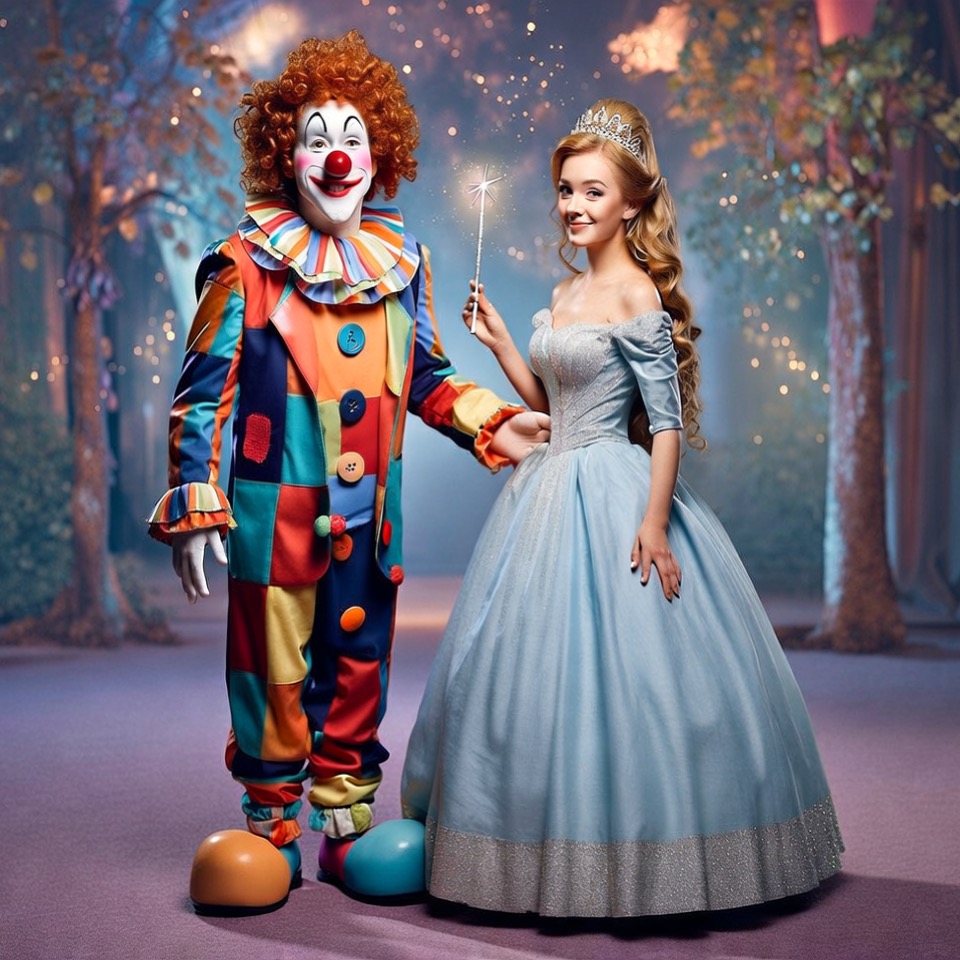} \\
    \end{tabular}
    }
    
    \caption{\textbf{Diversity Collapse in Standard Sampling.} Visual comparison for the prompt: \textit{``A clown and a princess holding a wand.''} While simply changing the random seed (consecutive seeds 0-3 shown in bottom row) results in repetitive layouts~\cite{dahary2025decisive} and limited variation, our method (top row) achieves significant structural and semantic diversity.}
    \label{fig:issue_illustration}
    
\end{figure}

Achieving such diversity is challenging, as recent state-of-the-art text-to-image models often exhibit limited semantic variation across samples generated from the same prompt (Figure~\ref{fig:issue_illustration}).
In particular, even when prompts are underspecified, different generations tend to converge on the same high-level semantic interpretation, differing only in
visually insignificant details, or exhibiting severe biases~\cite{cohen2025minethegapautomaticminingbiases}.
A likely contributing factor to this lack of semantic diversity is the training paradigm of modern text-to-image models, which emphasizes strict adherence to highly detailed captions~\cite{flux, gutflaish2025generating, BetkerImprovingIG}. 
While this design choice substantially improves controllability and prompt faithfulness, it also biases the model toward committing to a single realization of the prompt, leaving little room for semantically diverse outputs.

Prior work has addressed this limitation by perturbing the conditioning signal~\cite{sadat2023cads,um2025minority}, introducing repulsive forces between sampling trajectories~\cite{corso2023particle,dahary2026ontheflyrepulsioncontextualspace}, or generating large candidate pools from which diverse subsets are selected~\cite{parmar2025scaling}.
While successful at increasing diversity, these approaches do not offer explicit user control over the nature of the resulting variations. 
Consequently, differences across samples are driven by stochastic effects rather than explicit semantic factors.

In this work, we introduce the task of controlled semantic diversity, which enables users to explore generated images through meaningful, interpretable variations rather than relying on stochastic sampling.
We refer to this process as \textit{Semantic Browsing}, and view it as a conceptually different approach to diversity, where variations are explicitly specified rather than emergent. 
By semantic variations, we refer to changes in interpretable attributes of the image, such as object attributes or configurations (e.g., pose or spatial arrangement), global appearance factors (e.g., style, color palette, or lighting), or contextual elements (e.g., weather or background), while preserving all other aspects of the prompt, see Figure~\ref{fig:teaser}.

To achieve this controlled diversity, we impose structure on the diversity of generated outputs by leveraging the semantic reasoning capabilities of modern VLMs.
Specifically, we introduce an agentic workflow that expands the user prompt into a richer semantic representation and identifies meaningful dimensions along which variation is both plausible and under-specified.
These dimensions capture alternative semantic interpretations or design choices that are compatible with the original prompt but not explicitly specified by it.
We then organize them into a structured set of prompt alternatives, each corresponding to a distinct semantic choice.

This prompt-based formulation places two key requirements on the underlying image generator.
First, the generator must support fine-grained prompt-level control, so that semantic changes specified by the agentic workflow result in correspondingly precise visual changes. Second, it must preserve all aspects of the image that are not explicitly modified, ensuring that differences across the generated gallery arise solely from the intended semantic variations.
Notably, recent state-of-the-art text-to-image models naturally satisfy these requirements, as they are trained for strict adherence to detailed textual specifications~\cite{flux, gutflaish2025generating}. This makes them well suited to accurately reflect explicit prompt changes while maintaining consistency in attributes that are not mentioned.
This training paradigm is exemplified by FIBO~\cite{gutflaish2025generating}, which trains a text-to-image generator on long, structured captions to improve prompt adherence and controllability.

We evaluate our approach through extensive experiments across state-of-the-art text-to-image models, demonstrating consistent and substantial improvements in diversity over prior methods. Beyond increasing diversity, our method enables explicit control over the nature of the variations, allowing semantic differences to be specified and explored systematically rather than emerging from stochastic sampling. As illustrated in Figure~\ref{fig:teaser}, this results in structured galleries of images in which each output corresponds to a distinct, interpretable semantic alternative.

\begin{figure*}[t] %
    \centering
    
    \includegraphics[width=\textwidth]{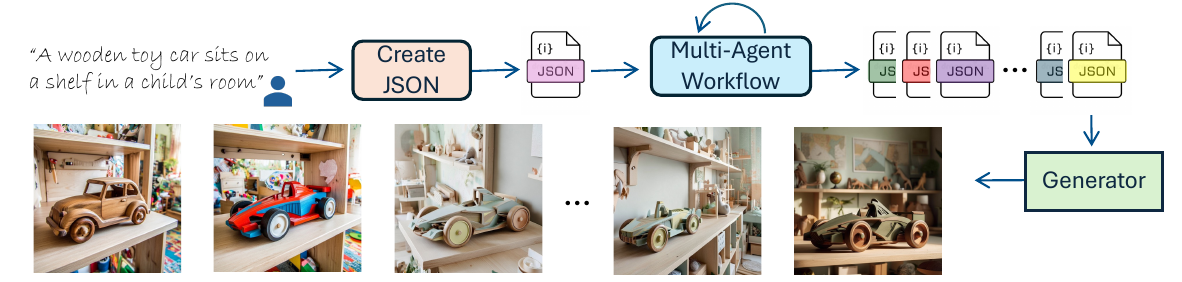}

    \caption{\textbf{Overview of the iterative generation flow.} A user prompt is transformed into a structured JSON format which is iteratively modified by a Multi-Agent workflow. This process creates structured diversity of JSON variations that remain faithful to the initial user intent, driving the generator to produce perceptually distinct images.}
    \label{fig:overview}
    
\end{figure*}

\section{Related Work}

\paragraph{Diversity in Text-to-Image Generation.}
Maintaining output diversity in Text-to-Image (T2I) systems is a persistent challenge, as common techniques like Classifier-Free Guidance (CFG)~\cite{ho2022classifier} often prioritize aesthetic fidelity at the cost of variety.
Recent work~\cite{jin2025stage} investigates the stage-wise dynamics of CFG, demonstrating how it suppresses diversity. This diversity collapse is further compounded in fast distilled diffusion models, a phenomenon directly linked to early generation dynamics~\cite{gandikota2025distillingdiversitycontroldiffusion}.

To mitigate
the CFG trade-off,
Autoguidance~\cite{karras2024guiding} replaces the unconditional model in CFG with a weaker variant, effectively restoring diversity while maintaining image quality. However, this requires the computationally intensive training of a separate weak model. Although recent works~\cite{gu2025insituautoguidanceelicitingselfcorrection,yehezkel2025navigating} propose lightweight alternatives to address this burden, the approach has demonstrated limited reliability in practice.

CADS~\cite{sadat2023cads} and Guidance Interval~\cite{kynkaanniemi2024applying} modulate the conditioning signal during denoising. While these methods improve sample variety, they can significantly degrade prompt alignment by relaxing guidance constraints. Other approaches, such as Particle Guidance~\cite{corso2023particle} and MinorityPrompt~\cite{um2025minority}, manipulate the sampling process through latent repulsion or loss-based optimization at the latent level. However, because these methods operate primarily in the latent space, they lack the semantic granularity necessary for rich conceptual variety. Similarly, SGI~\cite{parmar2025scaling} starts with a large pool of initial seeds and filters them during generation to reduce redundancy. While effective for batch variety, SGI is ultimately limited by the intrinsic diversity of the base generative model.
To overcome these limitations, Contextual Repulsion~\cite{dahary2026ontheflyrepulsioncontextualspace} applies repulsion within the contextual attention space. Although this shift improves semantic awareness and sample variety, the method still relies on stochastic diversity without explicit control over specific semantic axes.

A more recent approach to prompt-level variety is PAG~\cite{yun2025learning}, which utilizes GFlowNets for diverse sampling. However, PAG is constrained by its reliance on a specific training dataset and lacks a global view over the relationships between generated prompts. In contrast, our approach is training-free and utilizes a hierarchical tree structure of generated images. By reasoning about multiple nodes collectively within the tree, our system ensures semantic diversity through structural inheritance, avoiding the repetitive results that often occur in independent or unstructured generation. 

Beyond text-to-image generation, meaningful diversity~\cite{ICLR2024_19e2ed0e} and hierarchical exploration~\cite{NEURIPS2024_e262fc23} have also been studied in image restoration; our work instead uses hierarchy to organize explicit semantic alternatives for generation.

\paragraph{Creative Generation and Exploration.} Our framework operates at the intersection of structured diversity and open-ended creative exploration. While methods like ConceptLab~\cite{richardson2024conceptlab} and adaptive negative prompting~\cite{golan2025vlm} focus on exploring creative sub-categories of single objects, other approaches decompose and merge existing visual concepts for inspiration~\cite{vinker2023conceptdecompositionvisualexploration,goldberg2026inspiration}. In contrast, our method explicitly explores creative variations within the semantic space itself to organize alternative directions for generation.

\paragraph{Multi-Agent Systems for Controllable Generation.}
Current research has increasingly focused on utilizing specialized agents to enhance user control and refine the generation process. Maestro~\cite{wan2025maestro} employs a self-improving loop where multimodal agents act as critics to identify visual under-specification and iteratively refine the output for higher precision. Similarly, PromptSculptor~\cite{xiang2025promptsculptor} is a multi-agent framework that decomposes complex user queries into detailed, semantically rich descriptions to ensure the model captures every aspect of the user's intent. Proactive T2I Agents~\cite{hahn2024proactive} further improve control by leveraging belief graphs to actively clarify ambiguous instructions through dialogue.
Twin-Co~\cite{wang2025twin} follows a comparable strategy, employing an agentic feedback loop to systematically eliminate uncertainty in the prompt.

While these agent-based systems significantly enhance intent alignment and visual fidelity, they are fundamentally designed to converge on a single "best" version of the user's prompt. Since they focus on maximizing control over one optimal result, they ignore the many different ways a prompt could be interpreted. Our work departs from these by using multiple agents to drive exploration instead of just narrowing down intent. By organizing generations into a hierarchical tree, we ensure the system produces a wide range of creative results rather than settling on a single interpretation.

\section{Method}
\begin{figure*}
    \includegraphics[width=\textwidth]{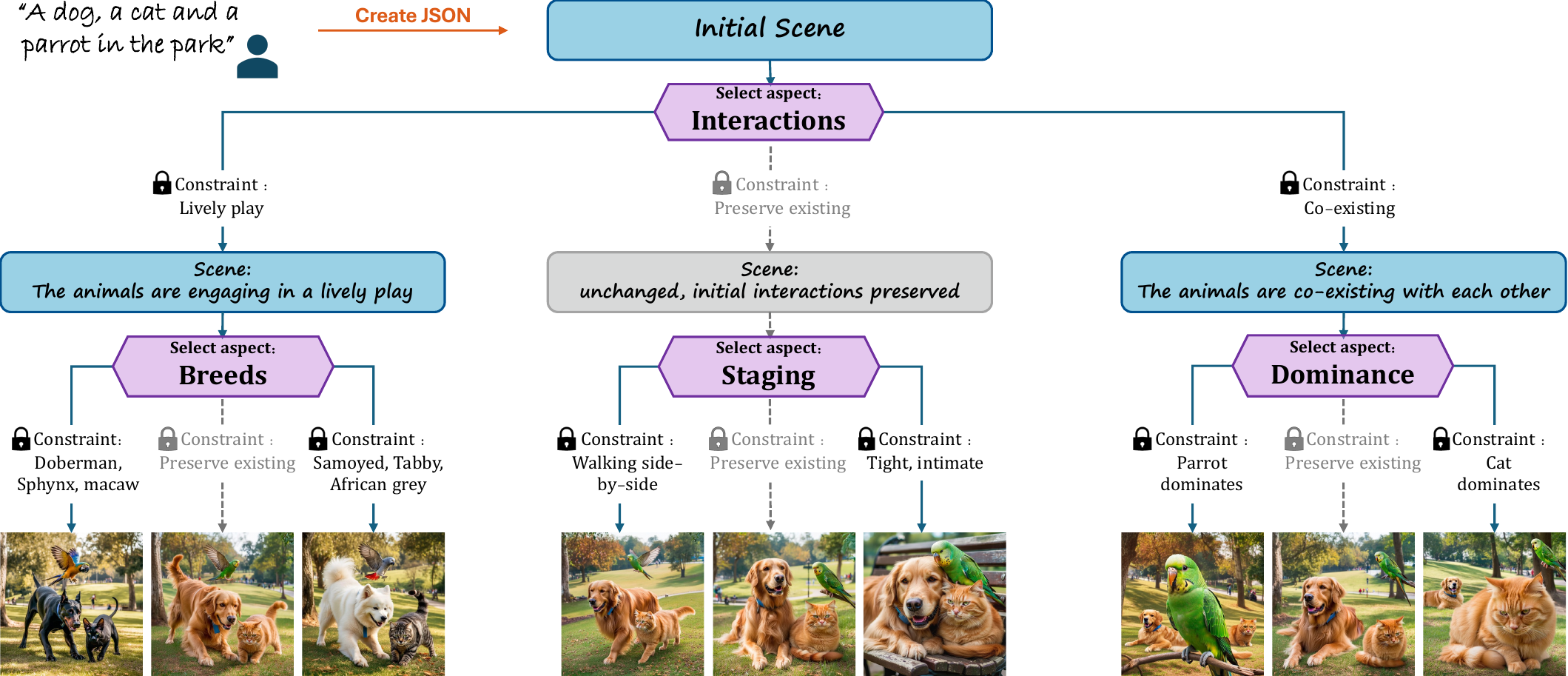}
    \caption{
    \textbf{Example of semantic browsing produced by our method.} Starting from an initial scene interpretation inferred from the user prompt, the method explores alternative realizations by committing explicit semantic constraints at each step. Each branching point corresponds to alternative realizations of a single semantic aspect, while previously fixed constraints are preserved. Branching points also include an option to preserve the current value of the selected aspect, allowing exploration to continue along other semantic dimensions. Every node is a fully specified, renderable scene; `preserve' branches propagate these states to the final level, ensuring the leaf nodes contain all generated representations ready for rendering.
    }
    \label{fig:browsing}
\end{figure*}

To enable controlled semantic exploration, we formalize the generation process as the construction of a hierarchical interpretative tree within a structured scene space. An overview of our method is demonstrated in Figure~\ref{fig:overview}, with a concrete example of a generated tree shown in Figure \ref{fig:browsing}.
This section details our notation, the fundamental requirements for navigable diversity, and the multi-agent workflow that iteratively expands this tree through reasoned semantic refinements.

\subsection{Setting}

Our method operates within the space $\mathcal{S}$ of fully specified scene interpretations, encoded as structured JSONs. This format allows for fine-grained control over objects, attributes, and global scene properties. Given a user prompt $p$, we first expand it into an initial scene interpretation $s_0 \in \mathcal{S}$ using a VLM. This root scene represents one complete, plausible specification of the prompt. The output of our method is a rooted tree $(V, E)$, where each node is a scene interpretation $s \in V \subset \mathcal{S}$.

In this structure, edges represent the atomic unit of semantic exploration. For any edge $(s_1, s_2) \in E$, there exists a semantic constraint $c$ that transforms $s_1$ into $s_2$. Each constraint $c$ is defined to be a specific instantiation of a broader semantic aspect $a$ (e.g., \textit{subject interactions}, \textit{scene composition}, or \textit{style}). For example, given a root scene $s_0$ derived from the prompt ``A dog, a cat and a parrot'' (Figure~\ref{fig:browsing}), a constraint $c$ might instruct that the animals' \textit{Interactions} are depicted as \textit{Lively play}. This results in a child $s_1$ adhering to this behavior while preserving the remaining context of $s_0$. Practically, this transition is executed by a VLM-based scene refiner $R$ such that $s_1 = R(s_0, c)$, ensuring that every step in the tree is both traceable and grounded in the preceding scene.

Subsequently, we can render each node $s$ using a modern prompt-adherent generator to produce a tree of images, enabling structured Semantic Browsing (Fig. ~\ref{fig:browsing}).

\subsection{Tree Requirements}
\label{Tree Requirements}
To ensure the tree remains both diverse and navigable, we require that for any node $s$ with a set of children, 
the applied set of semantic constraints must satisfy three interdependent requirements:
(i) \textit{Semantic Structuring}: All children of a parent node must be derived from a shared semantic aspect $a$. This property is essential for structured browsing, as it ensures that the branching at each level explores variations along a single, semantically meaningful dimension. For instance, in Figure~\ref{fig:browsing}, the children of the root node vary strictly based on the \textit{Interactions} between the animals, while the children of the rightmost branch vary based on the \textit{Dominance} in the scene.
(ii) \textit{Heterogeneity}: Each constraint $c$ must realize the common aspect $a$ in a unique manner. For example, under the \textit{Interactions} aspect shown in Figure~\ref{fig:browsing}, one branch instantiates the scenario of \textit{Lively play}, while its sibling instantiates a \textit{Co-existing} dynamic. This is the primary driver of diversity, forcing the model to explore different conceptual directions within the same shared aspect.
(iii) \textit{Plausibility}: Each constraint $c$ must be logically consistent with the original prompt $p$ and the preceding constraints in its branch. Plausibility acts as a filter for Heterogeneity: it ensures that while branches differ, they remain faithful to the parent scene's established context. Consider the rightmost branch in Figure~\ref{fig:browsing}: since it establishes that the animals are \textit{Co-existing}, the subsequent \textit{Cat dominates} constraint must be realized without aggression to avoid contradicting the parent state.

While these requirements define the target structure of the tree, balancing them simultaneously is a non-trivial reasoning task. We therefore employ a multi-agent workflow that serves as the engine for tree growth.
\subsection{Agentic Workflow}
Rather than generating the tree in a single pass, we expand it one node at a time. When the system expands a node $s$, our agentic workflow is triggered to generate its children through a staged process. The agentic workflow first identifies all details in the scene that remain flexible for change to ensure \textit{Plausibility}, then combines these details into a single coherent aspect $a$ to ensure \textit{Structuring}, and finally proposes and critiques a set of candidate refinements to maximize \textit{Heterogeneity}. This iterative, node-wise application ensures that every new set of children maintains the structural integrity and diversity required for effective Semantic Browsing.

Concretely, for every node $s$, we define the trajectory $C_s = (c_1, \dots, c_n)$ as the ordered sequence of constraints applied along the path from the root $s_0$ to $s$. The workflow uses $s$, $p$, and $C_s$ as context to ensure that new branches respect these previously fixed semantic decisions.

\begin{figure*}
\vspace{-8pt}
    \includegraphics[width=\textwidth]{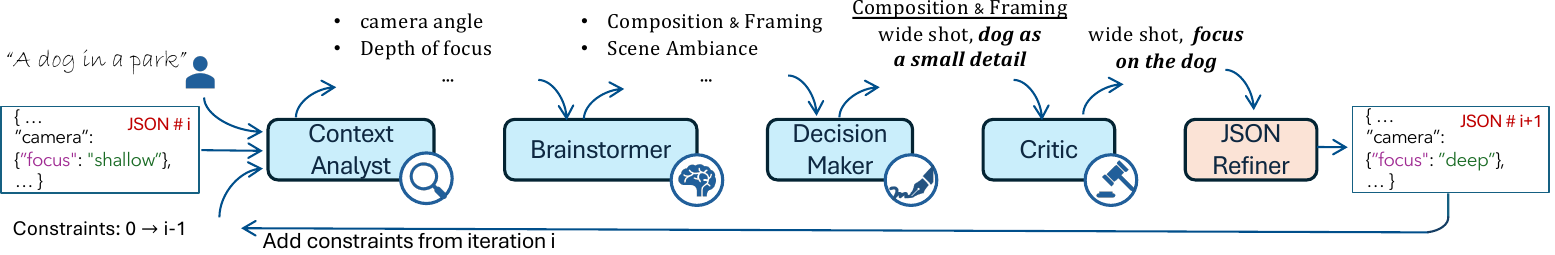}
    \caption{\textbf{Multi-Agent workflow guiding an iterative JSON generation process.} The pipeline takes the current JSON configuration and a history of constraints derived from previous modifications (including the user prompt) as inputs. A sequence of agents—\emph{Context Analyst}, \emph{Brainstormer,} \emph{Decision Maker,} and \emph{Critic}—analyzes these inputs to select an aspect to modify and formulate specific instructions. The JSON Refiner then translates these instructions into an updated JSON configuration, and the new modifications are added to the constraint set for subsequent iterations.}
    \label{fig:method}
    \vspace{-10pt}
\end{figure*}

Next, we describe each component of the agentic workflow in detail. An illustration of their interactions is shown in Figure~\ref{fig:method}.

\paragraph{Context Analyst.}
The Context Analyst is tasked with defining the admissible search space for modification by identifying granular, low-level details, directly addressing the \textit{Plausibility} requirement of the tree.
It operates on the insight that a generated scene $s$ is a composite of explicit specifications (enforced by the prompt $p$ or the accumulated constraints $C_s$) and unconstrained details (filled in by the VLM to complete the scene), which we consider eligible for mutation.
By explicitly distinguishing these, the Context Analyst isolates the set of mutable details $\{d_i\}$—such as specific colors, textures, or object sub-types—ensuring that subsequent changes target only the flexible components of the scene without violating its established logical coherence.
For example, in the scene from Figure~\ref{fig:browsing}, the Context Analyst identifies that while "a dog, a cat, and a parrot" must exist, their specific biological varieties (e.g., Doberman vs. Samoyed) are unconstrained details eligible for mutation.

However, once a particular breed is added to the constraint set, the corresponding scene details become fixed for rest of the subtree.
\paragraph{Brainstormer.}
The Brainstormer is responsible for laying the groundwork for meaningful \textit{Semantic Structuring}, ensuring that the tree evolves through clear, meaningful concepts.

Given the initial prompt $p$ and the accumulated constraints $C_s$, along with the set of low-level mutable details $\{d_i\}$ from the Context Analyst, the agent is tasked with identifying high-potential avenues for exploration.
It applies inductive reasoning to synthesize semantic aspects $\{a_i\}$ by aggregating several low-level details into one high-level aspect.
For instance, in the left branching in Figure~\ref{fig:browsing}, rather than varying the specific dog, cat, and bird species independently, the Brainstormer groups them under the cohesive aspect "Breeds,'' enabling coordinated modifications.

Crucially, it evaluates the potential of varying these candidates, explicitly assessing the magnitude of change (high, medium, or low) that varying each dimension would induce in the scene's \textit{narrative}, \textit{layout} and \textit{style}.
By prioritizing high-impact dimensions, the Brainstormer ensures that the tree evolves through significant conceptual shifts rather than trivial variations.

\paragraph{Decision Maker.}
The Decision Maker serves as the primary driver of \textit{Heterogeneity}.
By reasoning over the original prompt $p$, the current scene $s$, and the accumulated constraints $C_s$, the agent evaluates the candidate aspects $\{a_i\}$ suggested by the Brainstormer to identify prompt-dependent (see Appendix~\ref{generic_diversity}) dimensions that offer the richest potential for variation.
Operating strictly within this provided search space, it selects a single impactful dimension $a^*$ and instantiates it into a set of alternative semantic constraints $\{c_i\}$.
To ensure clear separation between sibling nodes, the Decision Maker actively reasons about the semantic boundaries of the scene, formulating constraints that offer widely divergent interpretations of $a^*$ rather than incremental adjustments.

\paragraph{Critic.}
Finally, the Critic acts as the validation layer, primarily enforcing \textit{Plausibility}.
It reasons over the proposed constraints against the original prompt $p$ and the accumulated constraints $C_s$, identifying potential contradictions or ambiguities that may have emerged during the creative process.
The Critic validates that the proposals faithfully realize the intended concept while maintaining strict alignment with the prompt $p$ and the accumulated context $C_s$.
Aligning with self-correction strategies~\cite{madaan2023self, du2023improvingfactualityreasoninglanguage}, it then refines the candidate set into precise, executable instructions, ensuring that the final branches are not only semantically distinct but are robustly formulated to produce high-fidelity generations.\\

Recent work demonstrates that prompting models to explicitly articulate their reasoning significantly enhances performance across various tasks~\cite{wei2023chainofthoughtpromptingelicitsreasoning, yao2023reactsynergizingreasoningacting}. Building on this literature, we design our agents to explicitly reason over their decisions before finalizing any action.

\begin{figure}
    \includegraphics[width=\columnwidth]{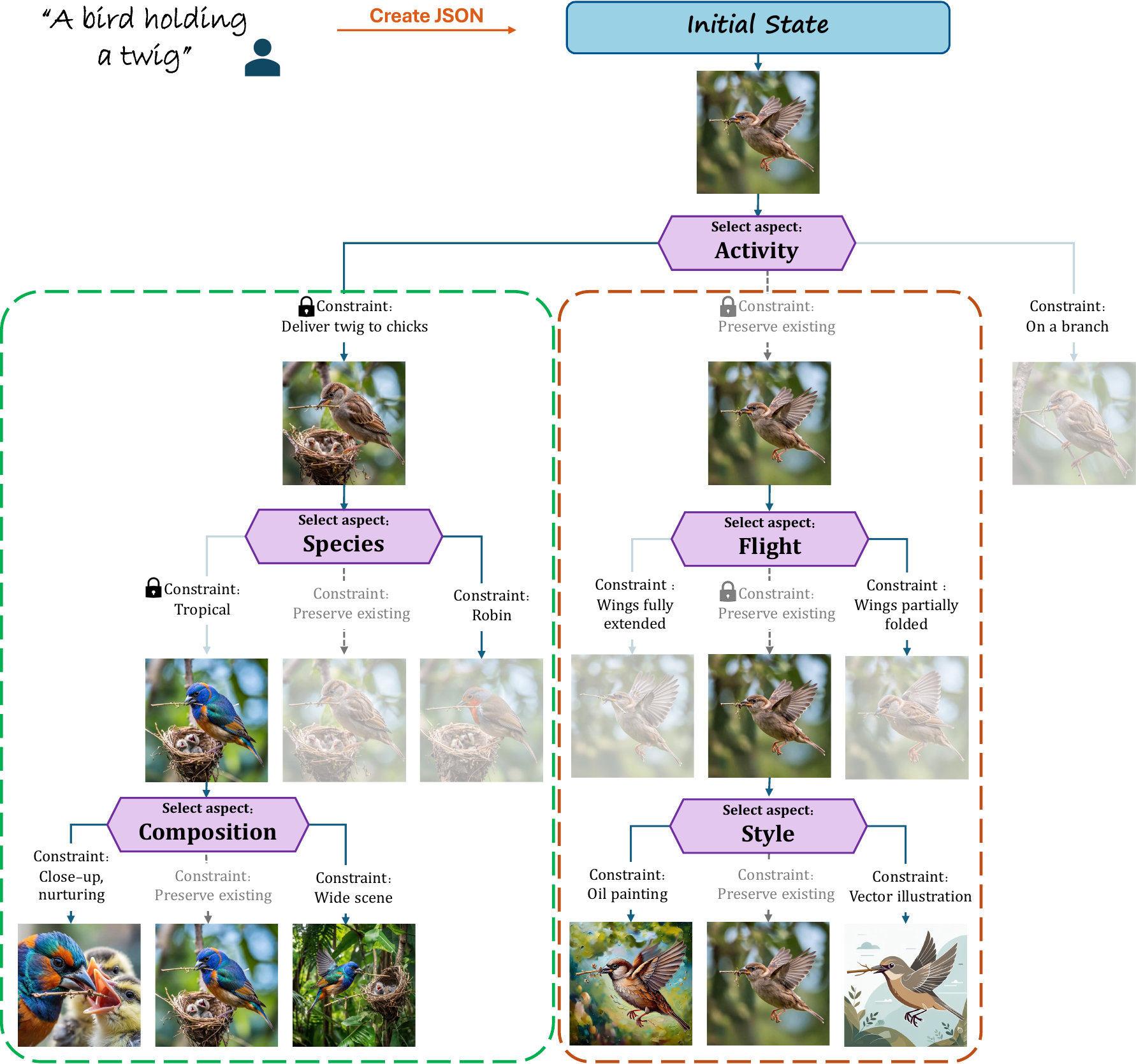}
    \caption{
    \textbf{Example of interactive semantic browsing.} At each node, users may either commit to a new realization of the selected semantic aspect and continue refining that interpretation (green), or preserve the current realization and explore other semantic aspects from the same state (orange). All nodes correspond to valid intermediate states that can be further expanded.}
    \label{fig:interactive_browsing}

\end{figure} 

\subsection{Interactive Browsing}

Our design inherently supports \textit{Interactive Browsing}: while we describe an automatic expansion strategy, the workflow allows a user to manually select any node of interest to trigger further generation, effectively continuing the exploration along a desired path, as demonstrated in Figure~\ref{fig:interactive_browsing}.

\section{Experiments}

In this section, we evaluate Semantic Browsing from three complementary perspectives. First, we demonstrate that our approach significantly enhances output diversity without compromising image quality or prompt alignment, benchmarking against established baselines designed to maximize diversity. Second, as Structured Diversity is a novel task whose hierarchical properties are not captured by existing diversity metrics, we introduce dedicated evaluations that measure the semantic and logical consistency of the generated hierarchy. Finally, we analyze the contribution of each component of our multi-agent workflow through ablation studies. Additional analyses, including a scaling ablation across tree depth and branching factor (Appendix~\ref{scaling_ablation}) and a sensitivity study of VLM choice (Appendix~\ref{vlm_sensitivity}), are provided in the appendix.

\begin{figure*}
    \centering
    
    {\raggedright \small \textbf{User Prompt:} A group of people doing yoga. \par}
    \vspace{0.2em}
    \includegraphics[width=\textwidth]{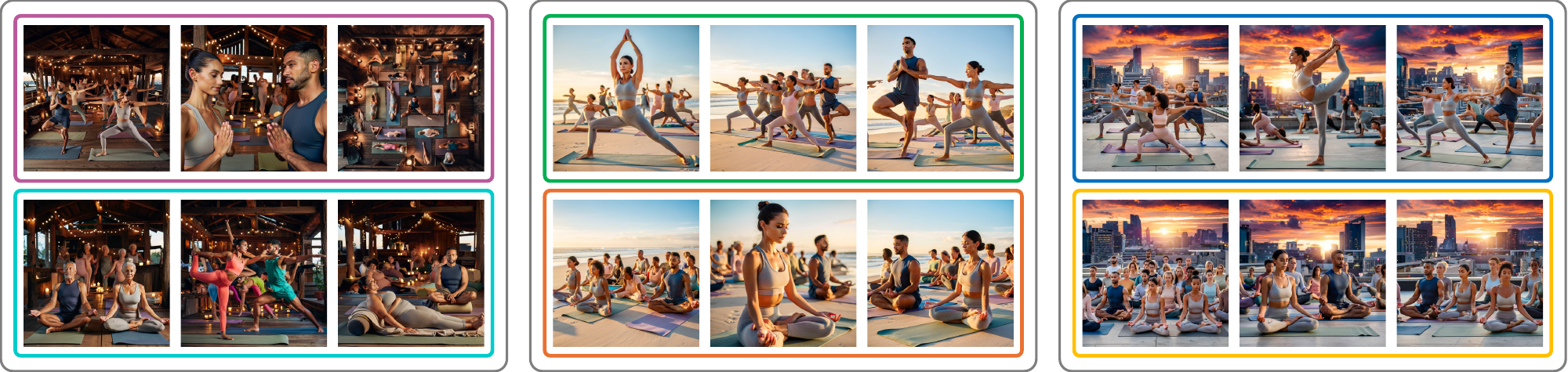}

    \vspace{0.9em}
    \hrule
    \vspace{1em}

    {\raggedright \small \textbf{User Prompt:} A cat and a goldfish bowl. \par}
    \vspace{0.2em}
    \includegraphics[width=\textwidth]{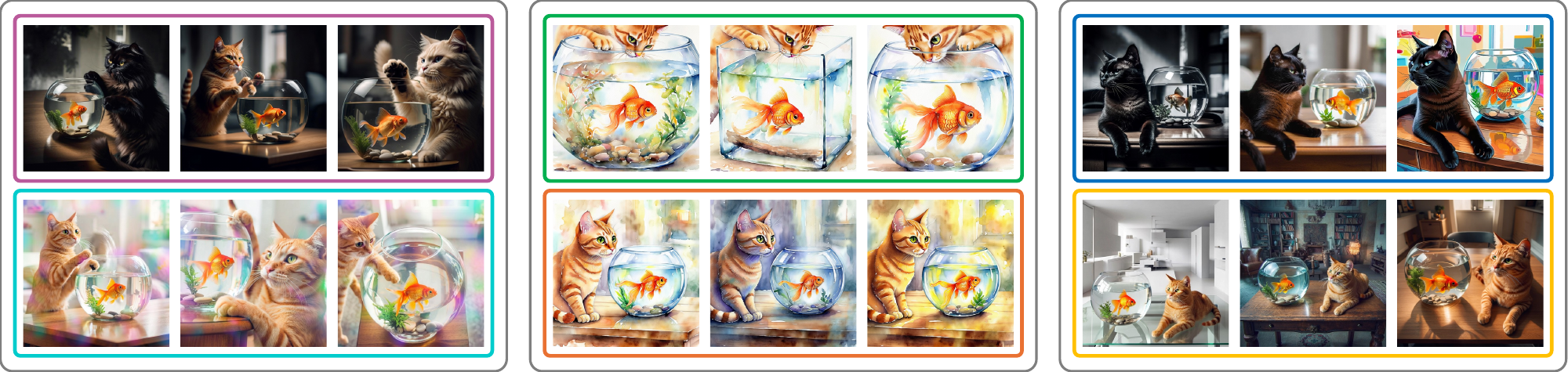}
    
    \caption{
    \textbf{Structured diversity results.} 
    All images shown are derived from a single initial scene. The outer gray groupings organize results that share a direct common ancestor scene. Inside, the colored boxes distinguish sibling branches (parallel variations that share the same parent but differ from one another by a single semantic aspect). This demonstrates how our method introduces meaningful diversity while preserving the coherence of the original user prompt.
    }
    \label{fig:results_main}
\end{figure*}

\begin{figure*}[t]
    \centering
    \setlength{\tabcolsep}{0.003\textwidth}
    {\small
    \begin{tabular}{c c c c c c c}
        Ours & VLM Seeding & Post-Hoc Opt. & Post-Hoc Opt. Temp. & CADS & Guidance Interval & Power-Law CFG \\
        
        \includegraphics[width=0.135\textwidth]{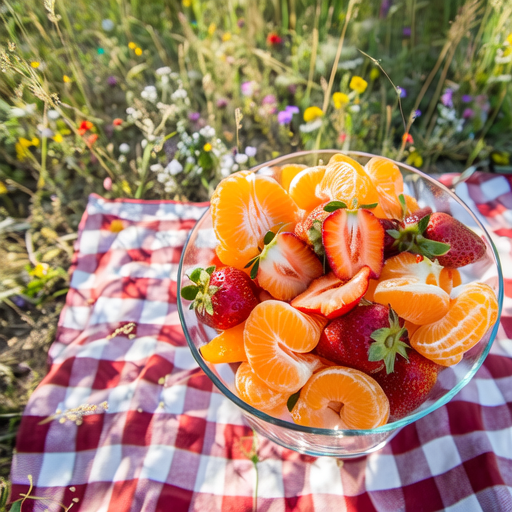} &
        \includegraphics[width=0.135\textwidth]{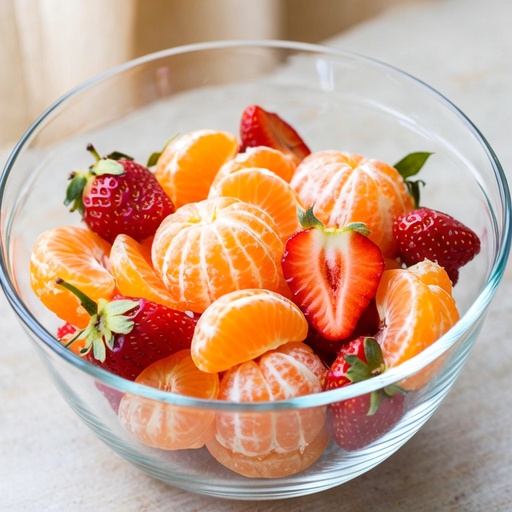} &
        \includegraphics[width=0.135\textwidth]{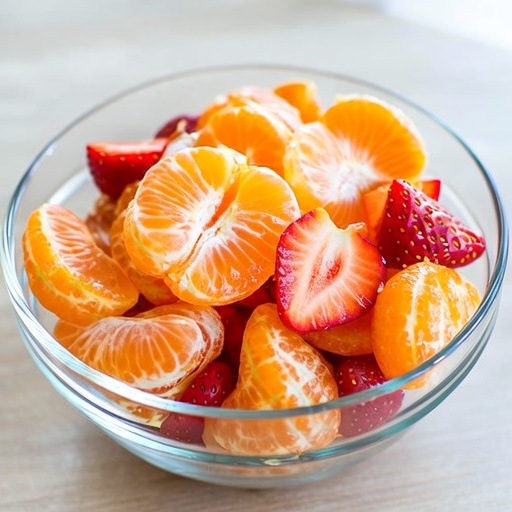} &
        \includegraphics[width=0.135\textwidth]{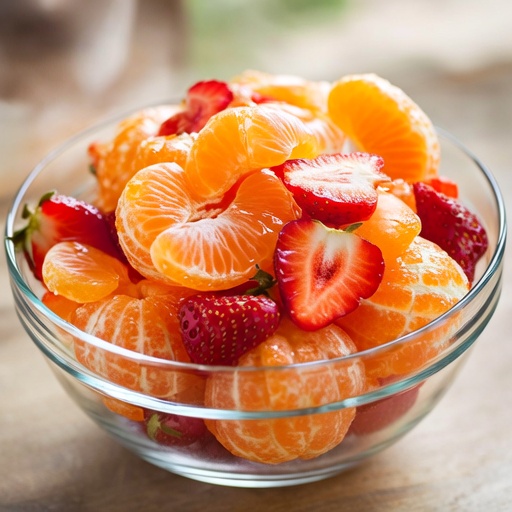} &
        \includegraphics[width=0.135\textwidth]{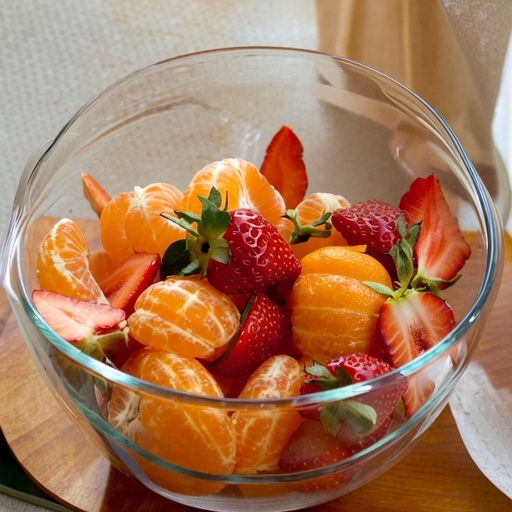} &
        \includegraphics[width=0.135\textwidth]{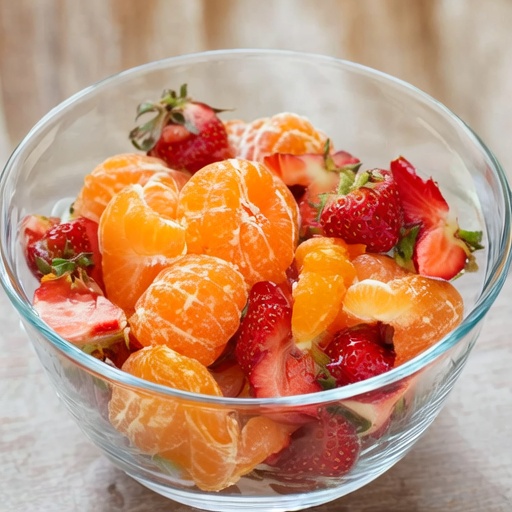} &
        \includegraphics[width=0.135\textwidth]{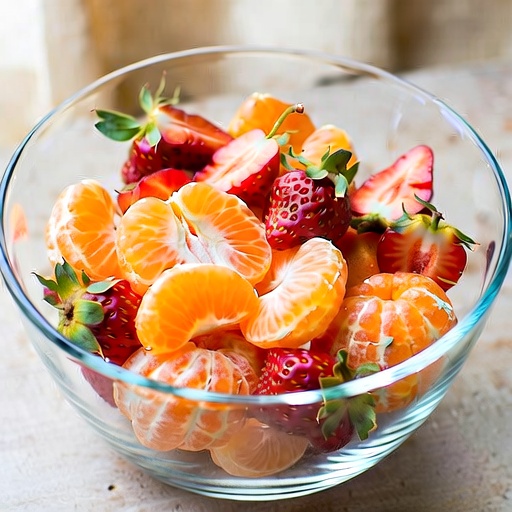} \\

        \includegraphics[width=0.135\textwidth]{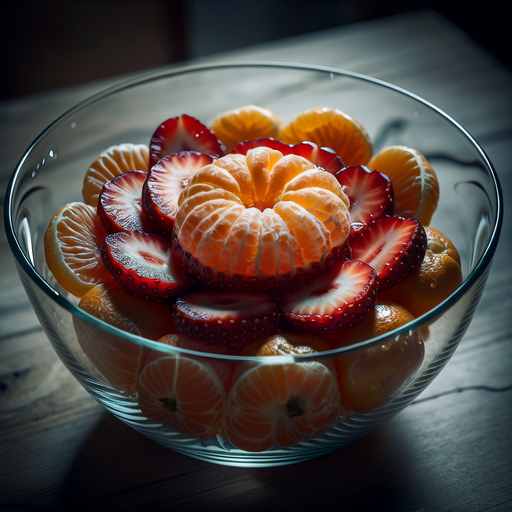} &
        \includegraphics[width=0.135\textwidth]{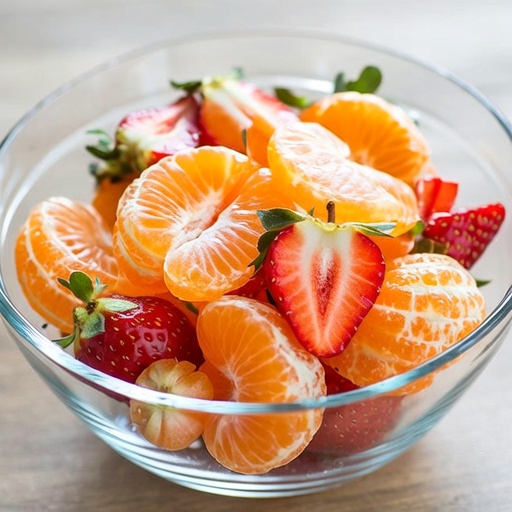} &
        \includegraphics[width=0.135\textwidth]{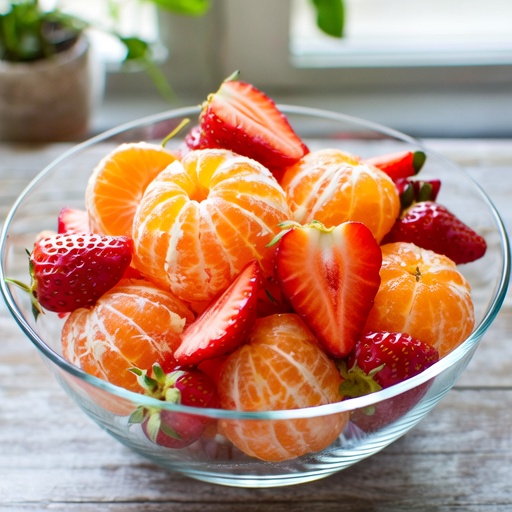} &
        \includegraphics[width=0.135\textwidth]{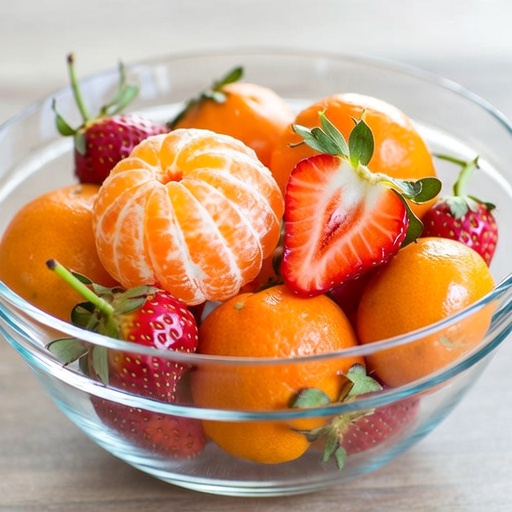} &
        \includegraphics[width=0.135\textwidth]{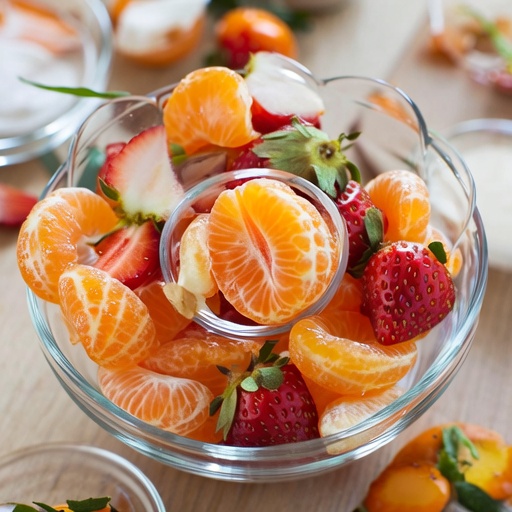} &
        \includegraphics[width=0.135\textwidth]{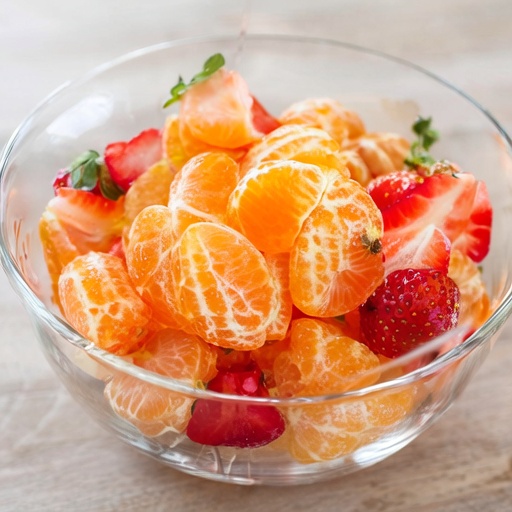} &
        \includegraphics[width=0.135\textwidth]{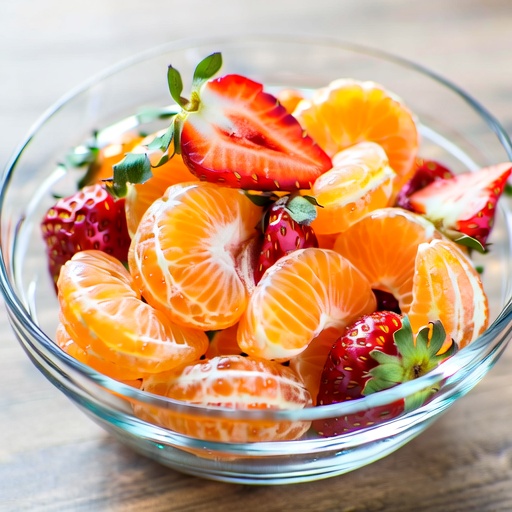} \\

        \includegraphics[width=0.135\textwidth]{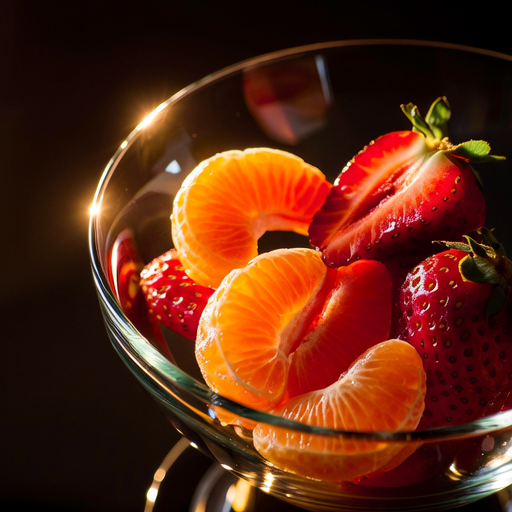} &
        \includegraphics[width=0.135\textwidth]{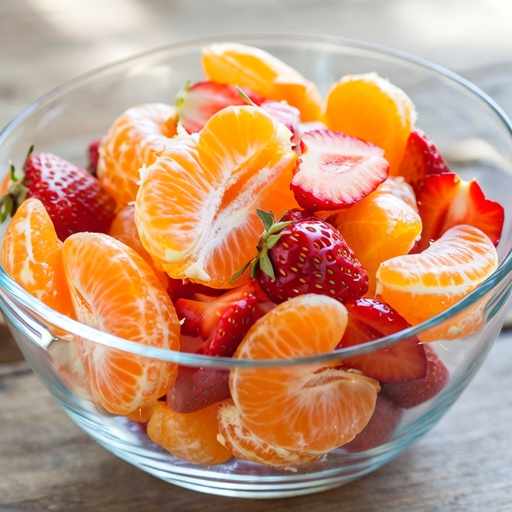} &
        \includegraphics[width=0.135\textwidth]{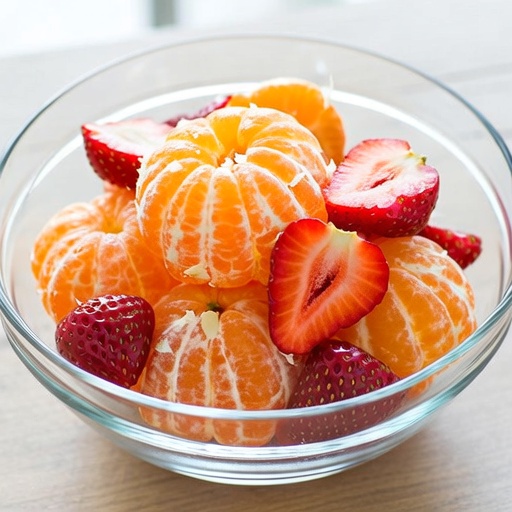} &
        \includegraphics[width=0.135\textwidth]{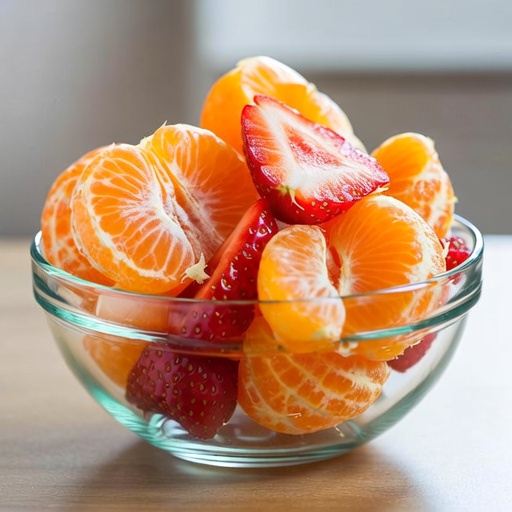} &
        \includegraphics[width=0.135\textwidth]{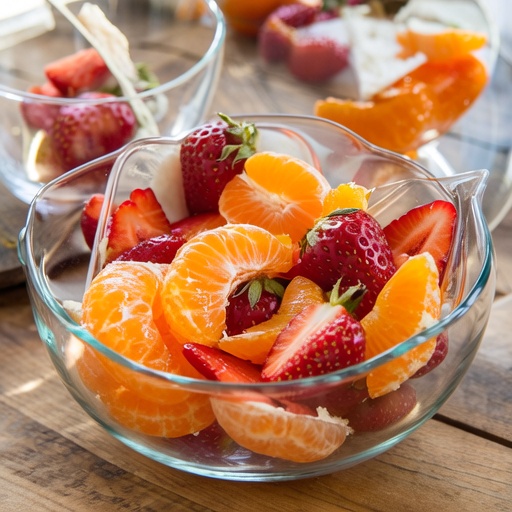} &
        \includegraphics[width=0.135\textwidth]{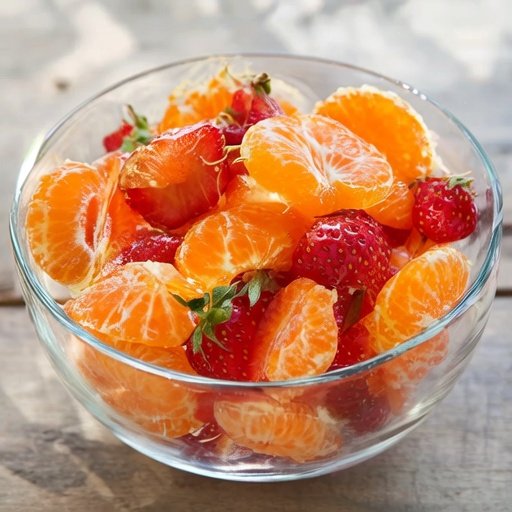} &
        \includegraphics[width=0.135\textwidth]{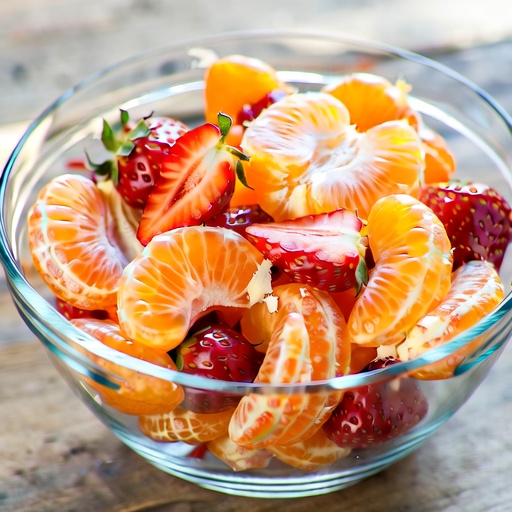} \\

        \includegraphics[width=0.135\textwidth]{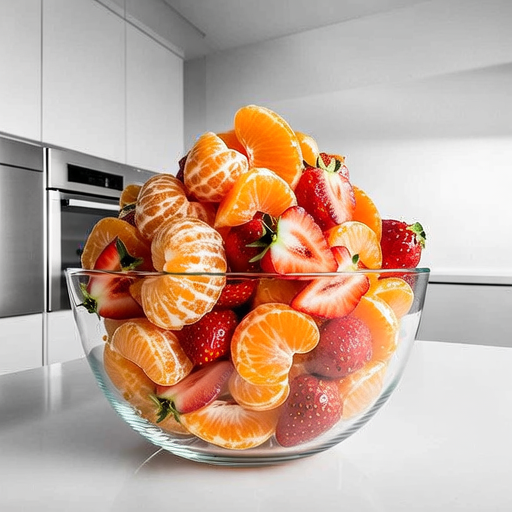} &
        \includegraphics[width=0.135\textwidth]{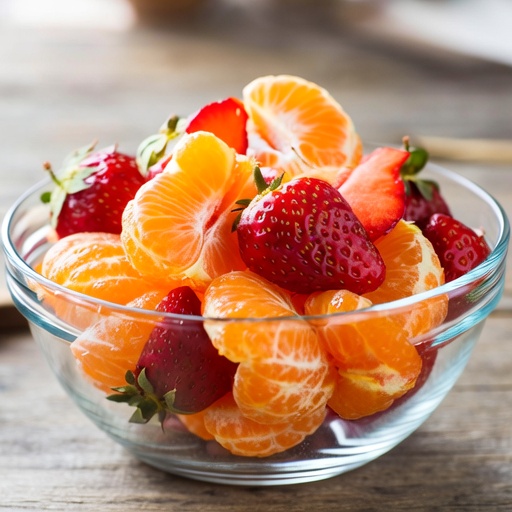} &
        \includegraphics[width=0.135\textwidth]{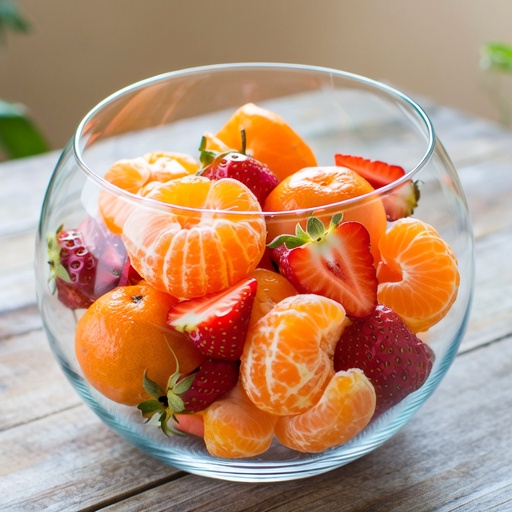} &
        \includegraphics[width=0.135\textwidth]{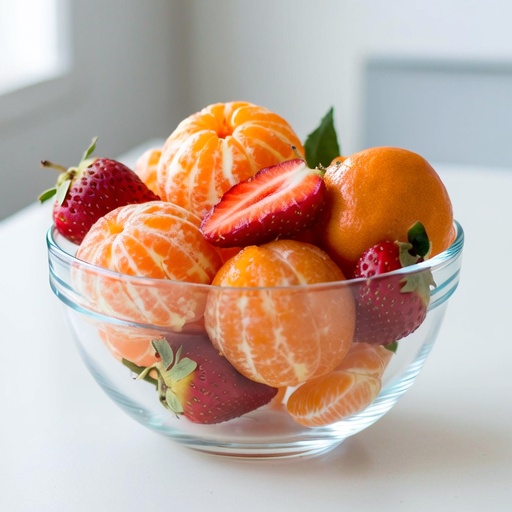} &
        \includegraphics[width=0.135\textwidth]{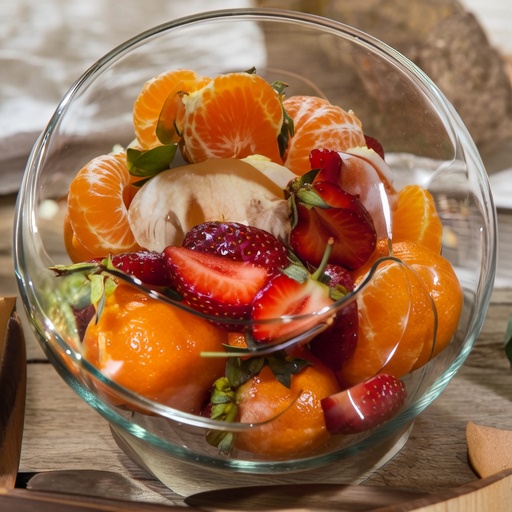} &
        \includegraphics[width=0.135\textwidth]{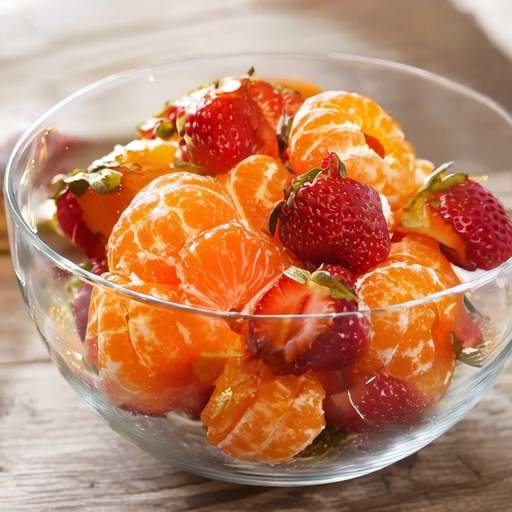} &
        \includegraphics[width=0.135\textwidth]{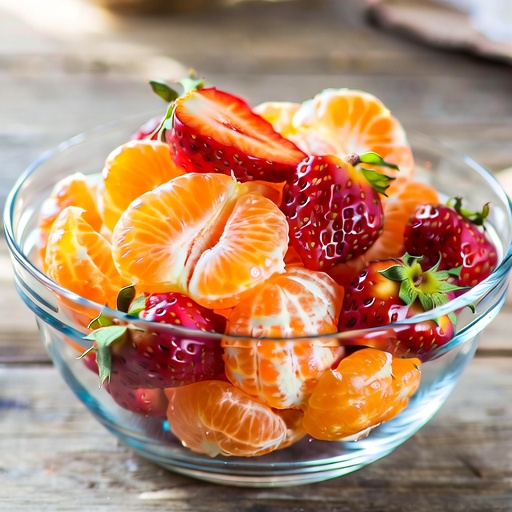} \\

    \end{tabular}
    \vspace{-6pt}
    }
    \captionof{figure}{\textbf{Qualitative comparison on the prompt:} \textit{“A glass bowl contains peeled tangerines and cut strawberries.”} Columns 2 and 5-7 report results using consecutive seeds with hyperparameters optimized for diversity. Columns 3-4 display the most diverse subset of four images selected from a larger candidate pool.
     While baseline methods exhibit limited variation, our method (column 1) successfully presents distinct and coherent interpretations.
     Examples include modifying the overall scene context, such as relocating the bowl to an outdoor picnic setting (row 1) or to a modern kitchen (row 4), the ordering and arrangement of the fruit (row 2), and the lighting conditions (row 3).}
    \vspace{-8pt}
    \label{fig:comparisons_fruit}
\end{figure*}

\paragraph{Experimental Setup.}
We implement our method using the FIBO framework~\cite{gutflaish2025generating}, leveraging its native prompt expander, refiner, and T2I generation modules. 
Additional details regarding the agent configuration and system prompts are provided in Appendix~\ref{implementation}.

\begin{figure*}
    \centering
    {\raggedright \small \textbf{User Prompt:} A dancer performing a dance. \par}
    \vspace{0.2em}
    \includegraphics[width=\textwidth]{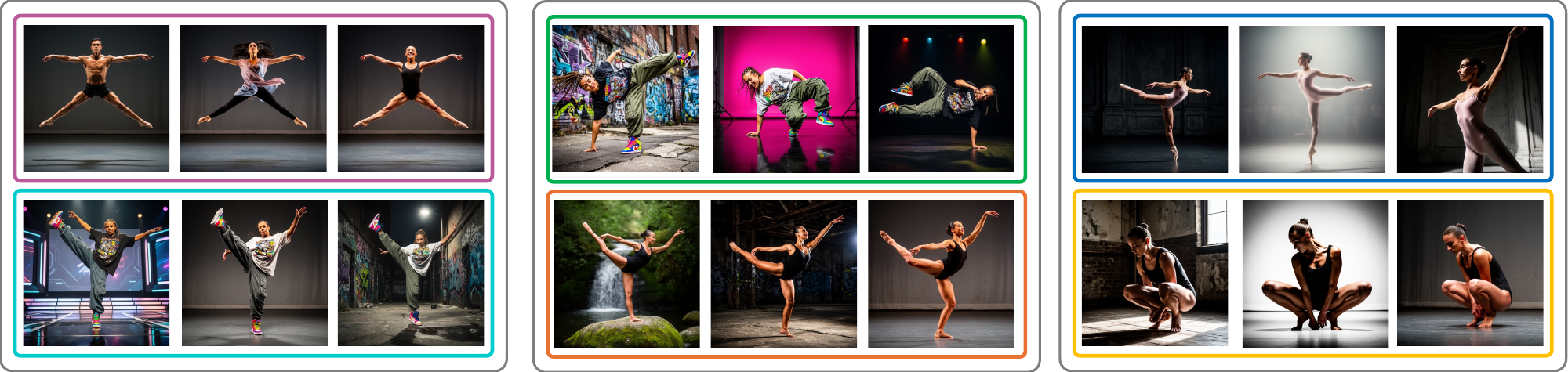}

    \vspace{0.9em}
    \hrule
    \vspace{1em}

    {\raggedright \small \textbf{User Prompt:} A red fox and a white fox playing a video game. \par}
    \vspace{0.2em}
    \includegraphics[width=\textwidth]{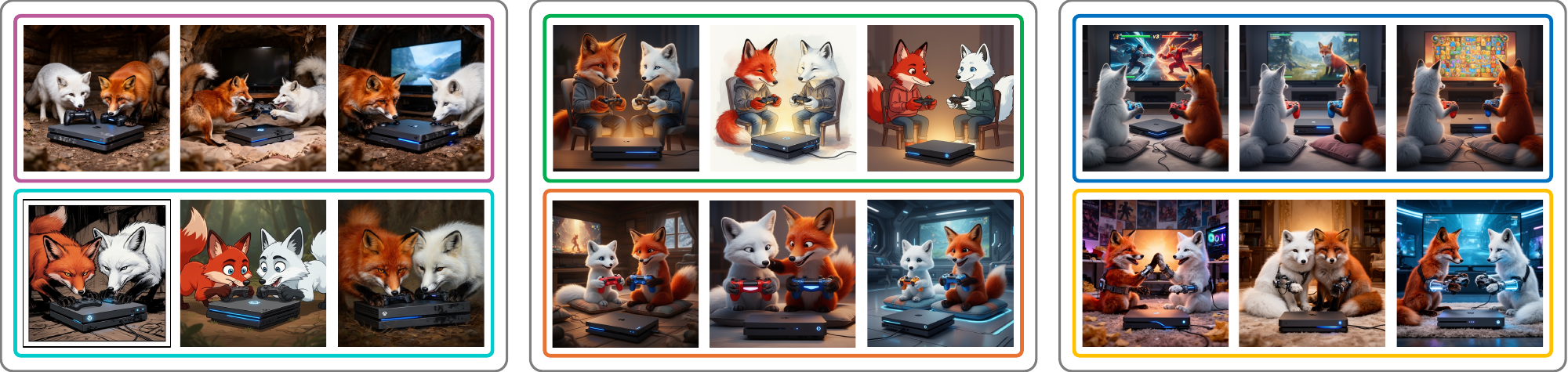}
    \caption{
    \textbf{Model-Agnostic Generation (FLUX.2)}. Qualitative results demonstrating the transferability of our framework to the FLUX.2 architecture. By utilizing our agentic flow solely for scene generation and FLUX.2 as the rendering backbone, we achieve consistent structured diversity.
    }
    \label{fig:flux}
\end{figure*}

\paragraph{Model-Agnostic Design.}
While our experimental results are obtained using FIBO’s generation pipeline, the proposed framework itself is model-agnostic and decoupled from the underlying rendering backbone.
To demonstrate this, we utilize FIBO's {VLM}-based modules for prompt enhancement and scene refinement, while employing a distinct architecture, FLUX.2~\cite{flux-2-2025}, to render the final images.
As shown in Figure~\ref{fig:flux}, our framework successfully separates semantic control from rendering, ensuring consistent performance across different backbones.

\paragraph{Gallery Generation Strategy}
To construct the final structured gallery, we employ our recursive tree expansion with a branching factor of three. At each node, the Decision Maker agent generates two distinct modification instructions, while a third branch retains the parent-node's JSON (identity mapping), ensuring that intermediate nodes are propagated unchanged to the leaf level. We expand this tree for three iterations, resulting in a final set of 27 leaf nodes ($3^3$), which constitutes our structured image gallery.

\paragraph{Baselines}
To rigorously evaluate the effectiveness of our approach, we compare it against several methods. For fair comparison, all baselines were implemented using the same underlying generation model (FIBO), and their hyperparameters were optimized to maximize diversity specifically in this setting. Full implementation details are provided in Appendix~\ref{Baselines}.

We employ three VLM-stochasticity-based baselines: \emph{Stochastic VLM Seeding} generates the target gallery by simply varying the random seed of the VLM to leverage inherent model stochasticity; \emph{Post-Hoc Diversity Optimization} applies a `generate-and-select' strategy, filtering a pool of 79 candidates generated with different VLM seeds (strictly matching our method's total VLM call budget) to retain the subset that explicitly maximizes diversity; and \emph{High-Temperature Post-Hoc Diversity Optimization}, which additionally increases sampling entropy of the VLM to force the selection of lower-probability tokens.

Furthermore, we evaluate established generator-level methods that induce diversity directly within the denoising process: \emph{CADS}~\cite{sadat2023cads}, \emph{Guidance Interval}~\cite{kynkaanniemi2024applying}, and \emph{Power-Law CFG}~\cite{pavasovic2025classifierfreeguidancehighdimensionalanalysis}. To test whether these inference techniques provide additive diversity beyond simple random initialization, we applied them in conjunction with Stochastic VLM Seeding to generate the full gallery of 27 images.

\subsection{Qualitative Evaluation}

We begin by presenting a visual overview of our generated outputs in Figure~\ref{fig:results_main}, with additional examples shown in Figures~\ref{fig:results_additional} and~\ref{fig:results_sup} (Appendix). These examples demonstrate the framework's ability to synthesize a rich variety of semantic interpretations from a single input prompt, spanning the full spectrum from granular entity adjustments to holistic changes in setting and mood. Notably, the results are structured into triplets, where each group stems from a shared unique ancestor node, highlighting how early branching decisions propagate into distinct yet internally consistent variations. Crucially, this expansion in diversity does not come at the cost of visual quality; the generated images consistently exhibit high fidelity and aesthetic coherence, validating our approach's ability to balance broad semantic exploration with high-quality generation.

To validate diversity against existing baselines, Figure~\ref{fig:comparisons_fruit} compares results for the prompt: \textit{“A glass bowl contains peeled tangerines and cut strawberries.”} 
While baseline methods converge on a single mode, our approach uncovers distinct and plausible interpretations. As shown in the first column, our framework successfully modifies the overall scene context, such as relocating the bowl to an outdoor picnic setting (row 1) or to a modern kitchen (row 4). We also vary the ordering and arrangement of the fruit (row 2) and the lighting conditions (row 3). This confirms our ability to retrieve heterogeneous, high-fidelity alternatives without compromising prompt adherence. Additional qualitative comparisons are provided {in Figures~\ref{fig:comparisons_bath}--\ref{fig:comparisons_woman}.}

\subsection{Quantitative Evaluation}

\begin{table}[t]
    \centering
    \renewcommand{\arraystretch}{0.9}

    \caption{\textbf{Comparison to Baselines.} Our method achieves top diversity (Vendi, DINO) while maintaining competitive Aesthetic scores; although lower on VQAScore, the result still reflects strong prompt adherence.}
    \label{tab:comparison}
    
    \resizebox{\columnwidth}{!}{%
    \begin{tabular}{l c c c c}
        \toprule
        \textbf{Method} & \textbf{Vendi}~$\uparrow$ & \textbf{DINO Sim.}~$\downarrow$ & \textbf{Aesthetic}~$\uparrow$ & \textbf{VQAScore}~$\uparrow$ \\
        \midrule
        Semantic Browsing (Ours) & \textbf{3.34} & \textbf{0.61} & 6.52 & 0.90 \\
        Stochastic VLM Seeding  & 2.60 & 0.76 & 6.53 & \textbf{0.93} \\
        Post-Hoc Diversity Opt.  & 2.79 & 0.67 & 6.51 & 0.92 \\
         High-Temp. Post-Hoc Opt.  & 2.85 & 0.66 & \textbf{6.56} & 0.92 \\
        CADS & 3.29 & 0.67 & 6.30 & 0.89 \\
        Guidance Interval. & 2.96 & 0.71 & 6.42 & 0.92 \\
        Power-Law CFG & 2.75 & 0.74 & 6.28 & 0.93 \\
        \bottomrule
    \end{tabular}%
    }
    
\end{table}

\paragraph{Dataset}
We conduct our evaluation on a subset of 50 prompts randomly sampled from the MS-COCO dataset~\cite{lin2015microsoftcococommonobjects}.

\paragraph{Metrics}
To provide a comprehensive assessment of our method, we report metrics across three dimensions: diversity, image quality, and prompt adherence.
\emph{Diversity} is quantified via the Vendi Score~\cite{friedman2023vendiscorediversityevaluation} in Inception space~\cite{szegedy2015rethinkinginceptionarchitecturecomputer} and pairwise DINO~\cite{oquab2024dinov2learningrobustvisual} similarity, capturing the extent of semantic variation across the gallery.
To evaluate \emph{quality} and validate diversity-enhancing mechanisms do not degrade visual fidelity, we report the Aesthetic Score~\cite{schuhmann2022aesthetic} (utilizing the LAION-based predictor~\cite{schuhmann2022laion5bopenlargescaledataset}).
For \emph{prompt adherence}
we utilize VQAScore~\cite{lin2024evaluatingtexttovisualgenerationimagetotext}.

Table~\ref{tab:comparison} presents the quantitative results against all baselines. Our method achieves superior diversity, securing the highest Vendi Score (3.34) and the lowest DINO Similarity (0.61), significantly outperforming all competing approaches. Crucially, this substantial expansion in semantic coverage is achieved while maintaining comparable Aesthetic Scores (6.52). This confirms that our framework successfully balances high-variance exploration with high image quality, avoiding the significant degradation often associated with maximizing diversity. While we observe a decrease in VQAScore, this may reflect inherent model biases within the evaluators, which often favor conventional, low-variance compositions rather than a true decline in prompt adherence. Regardless, the observed difference remains practically negligible.

We additionally report the computational overhead of our agentic workflow in Appendix~\ref{efficiency}. Despite the added structure, Semantic Browsing remains competitive in cost with baseline methods.

\paragraph{User Study}

\begin{figure}
    \centering
    \vspace{-10pt}
    
    \includegraphics[width=\linewidth]{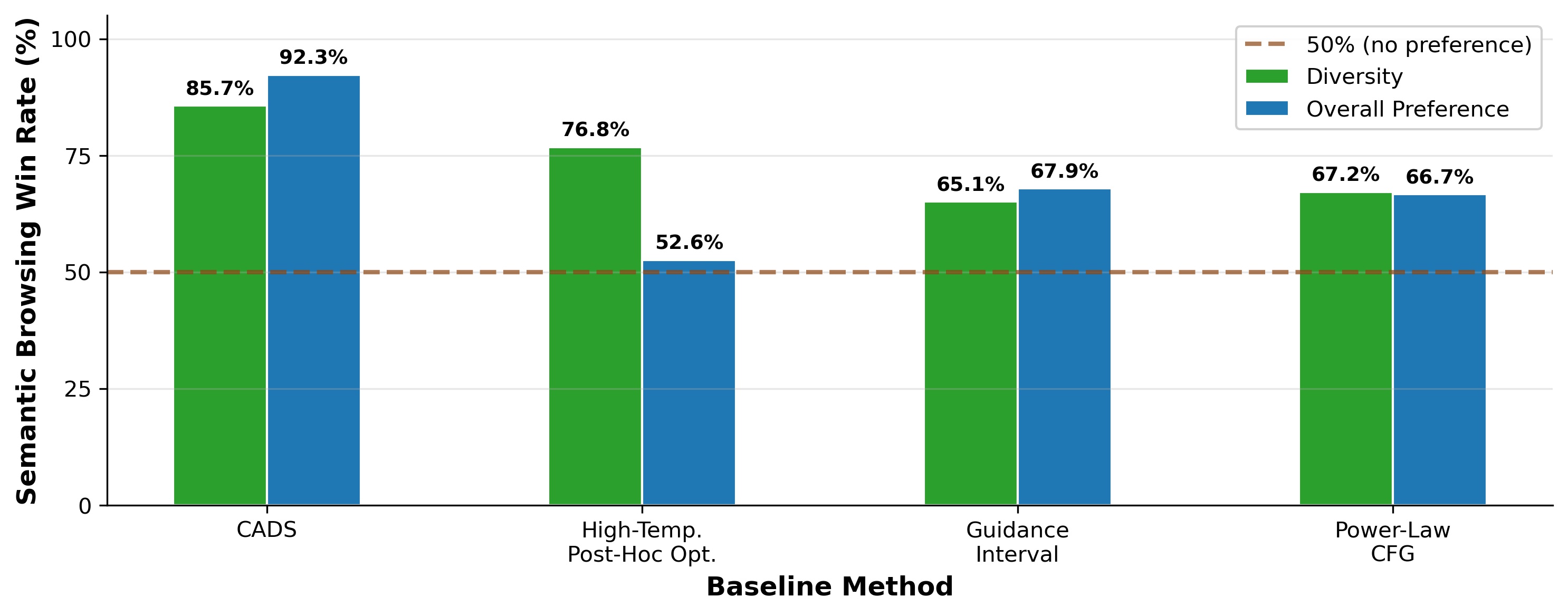}
    
    \vspace{-12pt}
    
    \caption{\textbf{Human Preference Study.} Our method (Semantic Browsing) dominates in Diversity across all comparisons while consistently outperforming baselines in Overall Preference.}
    \label{fig:user_study}
    
\end{figure}

Standard quantitative metrics often rely on dataset biases that penalize the very diversity our method aims to achieve. To directly assess perceptual quality and diversity, we conducted a head-to-head human evaluation with 25 participants, utilizing 12 randomly selected prompts for each baseline comparison. We compared Semantic Browsing against CADS, Guidance Interval, Power-Law CFG, and Post-Hoc Optimization. For the Post-Hoc baseline, we used the High-Temperature configuration as it yielded the optimal metric performance in Table \ref{tab:comparison}. As shown in Figure \ref{fig:user_study}, our method outperforms all baselines, achieving substantial win rates for superior diversity while consistently securing the majority vote for overall preference.

\subsection{Structure Evaluation}

Since Structured Diversity is a novel task, standard metrics are ill-equipped to capture the relational properties of the generated gallery. While adequate for quantifying the Heterogeneity requirement (Section \ref{Tree Requirements}), 
these metrics treat outputs as independent samples, ignoring the hierarchical dependencies that are unique to our method. Therefore, we introduce two evaluation protocols to specifically validate structural integrity and logical consistency.

For these structural evaluations, we deviate from the gallery generation process described previously. Instead of inspecting only the final leaf nodes (which include identity-mapped copies), we evaluate the complete set of unique nodes within the tree to accurately assess the internal coherence of the generation hierarchy.

\begin{figure}[t]  %
    \centering
    \includegraphics[width=\linewidth]{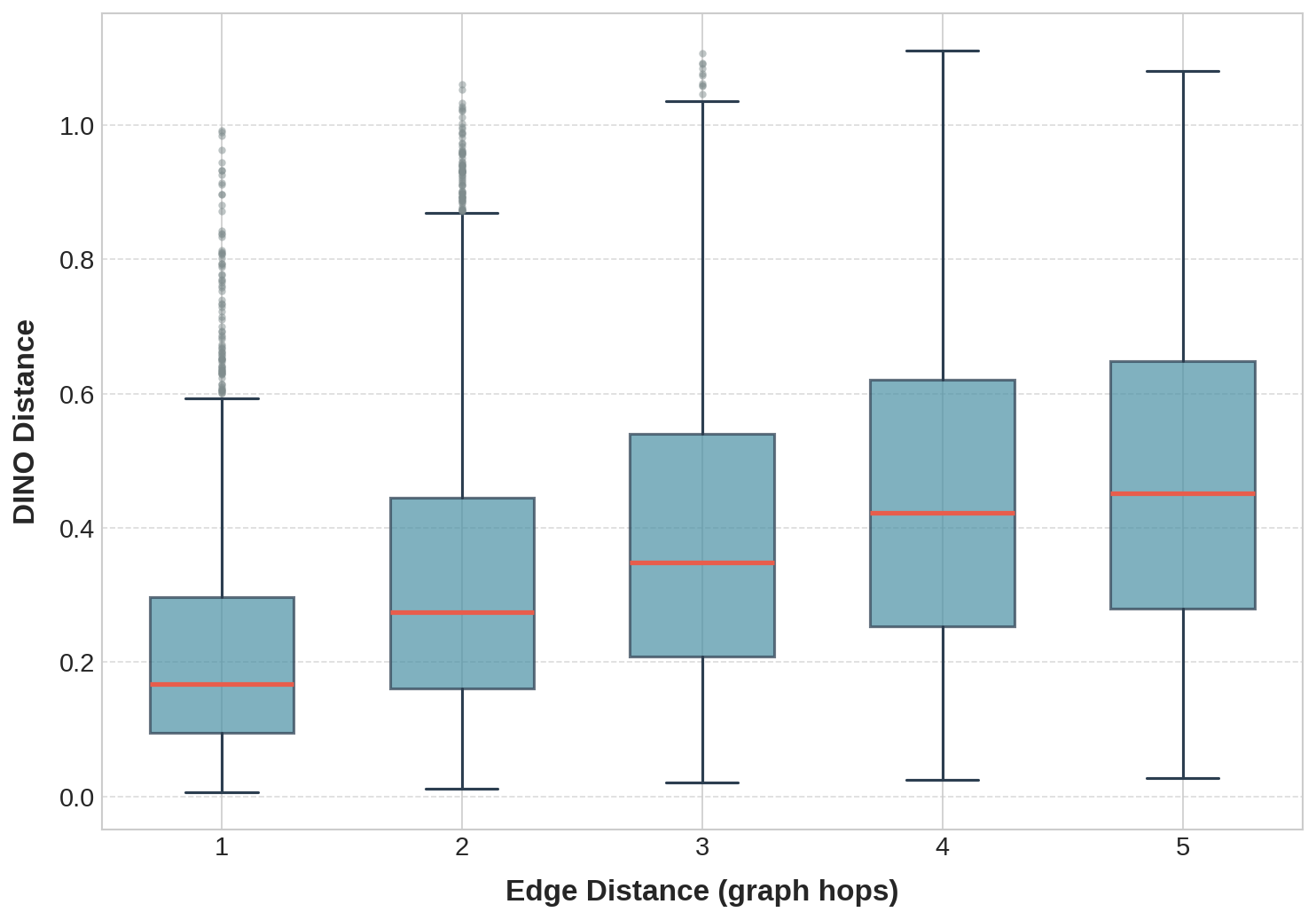}
    \caption{
     \textbf{Semantic-Topological Correlation.} Box plot showing the distribution of Pairwise DINO Distances as a function of graph distance (number of edge hops between nodes). The clear upward trend validates that our generation tree creates a coherent semantic space, where topological proximity translates to semantic similarity.
    }
    \label{fig:edge_vs_dino}
\end{figure}

\paragraph{Semantic-Topological Correlation.}
We hypothesize that the semantic distance between two images should correlate with their topological distance in the generation tree. This relationship is a direct consequence of the \emph{Semantic Structuring} requirement (Section~\ref{Tree Requirements}), which enforces that exactly one aspect changes between parent and child nodes. To verify this, we analyze Pairwise DINO Distance as a function of graph distance (path length between nodes).
Figure~\ref{fig:edge_vs_dino} presents the results of this analysis, aggregated across 50 generated trees (17,550 total pairs). We observe a strong positive correlation between the topological distance in the tree and the semantic distance in the image space. As the number of graph hops between two nodes increases, the median pairwise DINO distance rises monotonically (ranging from 0.168 at 1 hop to 0.452 at 5 hops). This confirms that our framework successfully satisfies the \emph{Semantic Structuring} requirement; the hierarchy effectively encodes semantic relationships, where neighboring nodes share visual characteristics and distant nodes exhibit greater semantic divergence.

\paragraph{Hierarchical Consistency.}
To validate that the tree maintains logical continuity, we utilize LLM-as-a-judge ~\cite{lee2024prometheusvisionvisionlanguagemodeljudge} to measure the alignment between a generated node and the constraints inherited from its ancestors. This metric explicitly validates the \emph{Plausibility} requirement (Section \ref{Tree Requirements}) by ensuring that diversity modifications do not violate established context.
Our framework achieves a high consistency score of 0.87 (out of 1.0), demonstrating that in the vast majority of cases, generated nodes successfully adhere to the cumulative constraints derived from the full root-to-node path. We note that this metric penalizes only the first violation of a constraint; we do not cumulatively penalize a child node if it remains consistent with a parent that has already violated a constraint, allowing us to isolate exactly where divergences occur.

\subsection{Ablation Study} 
To validate the architectural design of our framework, we conduct an ablation study to isolate the specific contribution of each agent. Our analysis confirms that the multi-agent decomposition is essential, as each component plays a distinct and necessary role in the generation pipeline.
To verify that constraints are not violated—an essential aspect of Plausibility—we utilize a VLM-as-a-judge~\cite{lee2024prometheusvisionvisionlanguagemodeljudge} to measure the alignment between a generated node and the constraints inherited from its ancestors. We refer to the average of this score as \textit{Hierarchical Consistency}.

\paragraph{Context Analyst}
When the Context Analyst is removed, the burden of interpreting the semantic gap between the high-level user prompt and the low-level JSON scene representation falls entirely on the Brainstormer. Without explicitly enforcing the internalization of this gap, non-admissible details change, resulting in a significant drop in Plausibility.
Table~\ref{tab:critic_context_analyst_ablation} (w/o Context Analyst) quantifies this impact. While the VQAScore remains stable (0.90), indicating that prompt adherence is preserved, the Hierarchical Consistency drops from 0.87 to 0.82. This divergence confirms that the Context Analyst is specifically required to maintain contextual continuity, directly associated with the Plausibility requirement.

\paragraph{Brainstormer and Decision Maker}
We evaluate the impact of merging the Brainstormer and Decision Maker into a single, unified agent. Since the Brainstormer is responsible for Semantic Structuring and the Decision Maker for Heterogeneity, the unified agent struggles to optimize for both tasks simultaneously, leading to a degradation in tree quality.
Separating these roles increases the global mean DINO distance from 0.362 (unified) to 0.389 (separated), representing a 7.2\% relative improvement in overall diversity. 

Table~\ref{tab:ablation_dino} decomposes this improvement by topological distance. The full workflow consistently exhibits larger DINO distances across all edge distances, confirming that the specialized role separation yields significantly greater structured diversity compared to the unified ablation.

This demonstrates the critical roles of these agents in maintaining Heterogeneity and Semantic Structure, validating that distinct architectural components are required to satisfy these dual objectives.

\paragraph{Critic} The Critic acts as the final safeguard for prompt adherence and logical consistency.
Table~\ref{tab:critic_context_analyst_ablation} (w/o Critic) shows that ablating this agent reduces VQAScore from 0.90 to 0.87, while Hierarchical Consistency remains stable. This suggests that while upstream agents mostly maintain internal constraint consistency, the Critic remains a necessary component to catch rare violations that do occur. The drop in VQAScore confirms that the Critic is essential for preventing semantic drift, ensuring that the output remains prompt adherent.
\begin{table}[t]
    \centering
    \small %
    \renewcommand{\arraystretch}{0.9} %
    
    \caption{\textbf{Brainstormer / Decision Maker Ablation.} Comparison of mean pairwise DINO distance. The full workflow consistently yields higher diversity across all graph hops.}
    \label{tab:ablation_dino}
    
    \begin{tabular*}{\columnwidth}{@{\extracolsep{\fill}} c c c}
        \toprule
        \textbf{Edge Dist.} & \textbf{Full Workflow} & \textbf{Unified Agent} \\
        \midrule
        1 & 0.221 & 0.218 \\
        2 & 0.326 & 0.313 \\
        3 & 0.392 & 0.365 \\
        4 & 0.450 & 0.411 \\
        5 & 0.473 & 0.439 \\
        \bottomrule
    \end{tabular*}
\end{table}

\begin{table}[t]
    \centering
    \caption{\textbf{Ablation of Agents Responsible for Plausibility.} We demonstrate the complementary roles of the Context Analyst and the Critic. The Context Analyst is essential for internal logical continuity (Hierarchical Consistency), while the Critic safeguards prompt adherence (VQAScore), confirming that both are necessary to maintain the full scope of plausibility.}
    \label{tab:critic_context_analyst_ablation}

    \resizebox{\columnwidth}{!}{%
    \begin{tabular}{l c c c}
        \toprule
        \textbf{Metric} & \textbf{Full Workflow} & \textbf{w/o Context Analyst} & \textbf{w/o Critic} \\
        \midrule
        VQAScore & 0.90 & 0.90 & 0.87 \\
        Hierarchical Consistency & 0.87 & 0.82 & 0.87 \\
        \bottomrule
    \end{tabular}%
    }
    
\end{table}

\section{Conclusions, Limitations and Future Work}

We have presented a structured approach for generating semantic diversity in text-to-image models. At a high level, this work adopts a perspective in which diversity arises from explicit semantic decision-making rather than from stochastic variation alone. By committing to concrete semantic choices during generation, differences between outputs become interpretable and persistent rather than incidental. Consequently, the generated results form not just a collection of images, but a structured and navigable semantic space of alternative scene interpretations.

This perspective was enabled by recent text-to-image models that exhibit strong prompt adherence, which we treated not as a limitation on diversity but as an enabler for precise semantic control. Rather than optimizing toward a single refined output, the formulation emphasized exploration, using a multi-agent reasoning process to surface and maintain multiple plausible interpretations of an under-specified prompt. These interpretations were constructed through sequences of inherited semantic commitments, leading to structured semantic variation in which previously fixed decisions remained consistent while new variations were introduced in a controlled and interpretable manner.

The method presented here has several limitations that stem primarily from its current realization. The quality and usefulness of the explored semantic space depend on the underlying generative model’s ability to faithfully realize fine-grained prompt modifications, as well as on the semantic reasoning capabilities of the agents that propose and evaluate variations. In particular, while modern VLMs are effective at maintaining consistency and plausibility, their ability to propose rich and diverse semantic alternatives remains limited relative to the breadth of interpretations one might ultimately wish to explore, which can constrain the scope of the resulting semantic space.

More broadly, although this work focused on image generation, the notion of structuring diversity through explicit semantic decisions suggests a more general paradigm. Looking forward, we believe that structured semantic exploration could extend beyond images to other generative domains, such as video, 3D content, or multimodal generation, offering a principled way to move from isolated outputs toward coherent, navigable spaces of alternatives.

{
\section*{Acknowledgments}
We thank Nir Goren, Saar Huberman, and Shelly Golan for helpful discussions and early feedback on this work. We also thank the ECCV 2026 reviewers for their constructive comments and suggestions. This research was supported in part by the Israel Science Foundation (grants no. 2492/20 and 1473/24), Len Blavatnik, and the Blavatnik Family Foundation. We also thank NVIDIA for their generous support through the NVIDIA Academic Grant Program, which provided GPU hours via Brev for this research.
}

\clearpage
\bibliographystyle{ACM-Reference-Format}
\bibliography{main}

\clearpage
\appendix
\section*{Appendix}
\section{Baselines}
\label{Baselines}
To rigorously evaluate the effectiveness of our approach, we compare it against the following methods. For fair comparison, all baselines were implemented using the same underlying generation model (FIBO), and their hyperparameters were optimized specifically for this setting.

\paragraph{Stochastic VLM Seeding.}
A na\"ive baseline where we generate the target gallery size (27 images) by simply varying the random seed of the initial VLM call (prompt-to-JSON), relying on the model's inherent stochasticity for diversity.

\paragraph{Post-Hoc Diversity Optimization.}
A 'generate-and-select' baseline where we over-generate a pool of 79 candidates and select the optimal subset of 27 images that maximizes pairwise DINO distance via Quadratic Integer Programming (QIP)~\cite{parmar2025scaling}. Due to the high computational cost of this optimization, we impose a strict 300-second time limit per instance. Crucially, the pool size of 79 matches the total number of LLM calls used in our proposed tree-generation method. This ensures a fair comparison under a fixed computational budget, testing whether our structured, hierarchical expansion yields better diversity than simply running the base prompt-to-JSON flow repeatedly.

\paragraph{High-Temperature VLM Seeding.}
A variation of the post-hoc diversity optimization baseline where we maximize the sampling temperature of the initial VLM call. Unlike the standard baseline which operates within a conventional probability distribution, this method forces the selection of lower-probability tokens. We include this to strictly evaluate whether the diversity gap can be closed simply by increasing the entropy of the unstructured generation process, or if our structured intervention is necessary.

\paragraph{CADS (Condition-Annealed Diffusion Sampler).} ~\cite{sadat2023cads}
A method that induces diversity by injecting random noise into the text embeddings within the input space of the text-to-image generator.
We optimized the hyperparameters to maximize diversity and set them to: $\tau_1=0.5$, $\tau_2=0.9$, $s=3$, and $\psi=0.5$ (using notations from the original paper).

\paragraph{Guidance Interval}~\cite{kynkaanniemi2024applying}
A guidance modification where Classifier-Free Guidance (CFG) is applied only during a specific timestep interval in the middle of the denoising process. We note that FIBO demonstrates relatively strong performance even without standard CFG, therefore guidance is applied only across one-fifth of the total timestamp range.

\paragraph{Power-Law CFG}~\cite{pavasovic2025classifierfreeguidancehighdimensionalanalysis}
A gradient scaling technique where the CFG update is multiplied by its norm raised to the power of a pre-determined hyperparameter. We optimized the scaling hyperparameter to maximize diversity and set it to 0.3.

\hfill \break
Since CADS, Guidance Interval, and Power-Law CFG are generator-level methods (modifying the inference process rather than the prompt structure), we applied them in conjunction with Stochastic VLM Seeding to generate the full gallery of 27 images. This ensures we evaluate whether these inference techniques provide additive diversity beyond simple random seeding.

\begin{figure*}
    \centering
    {\raggedright \small \textbf{User Prompt:} A group of people riding on a group of elephants. \par}
    \vspace{0.2em}
    \includegraphics[width=\textwidth]{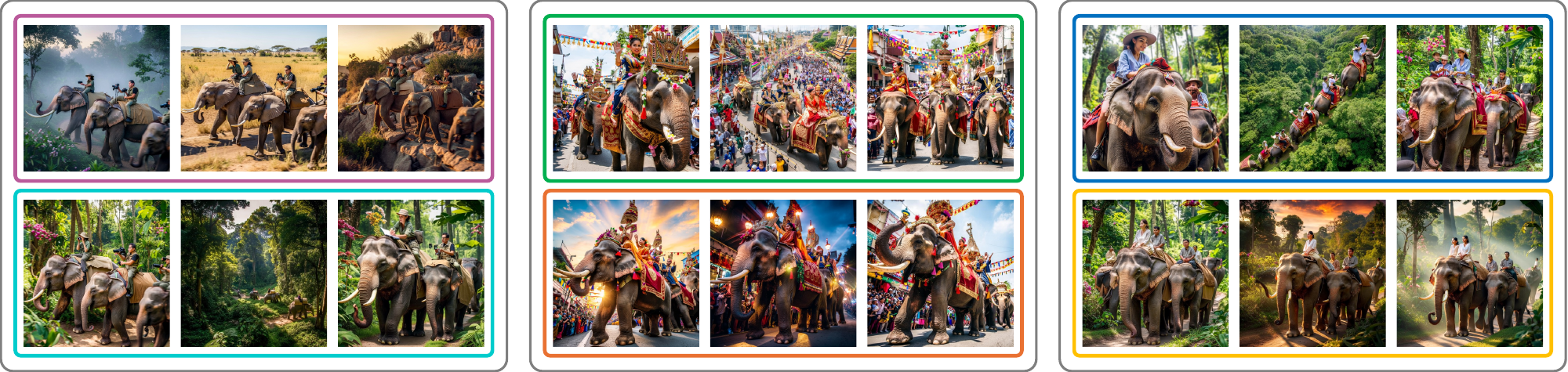}

        \vspace{0.9em}
    \hrule
    \vspace{1em}

    {\raggedright \small \textbf{User Prompt:} A birthday cake. \par}
    \vspace{0.2em}
    \includegraphics[width=\textwidth]{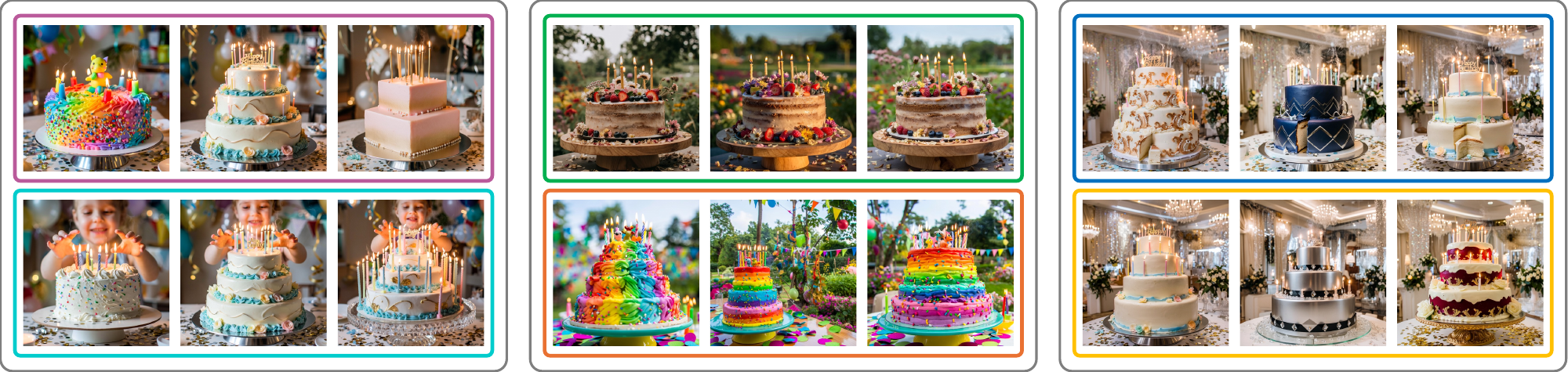}

    \vspace{0.9em}
    \hrule
    \vspace{1em}
    
    {\raggedright \small \textbf{User Prompt:} A family of monkeys. \par}
    \vspace{0.2em}
    \includegraphics[width=\textwidth]{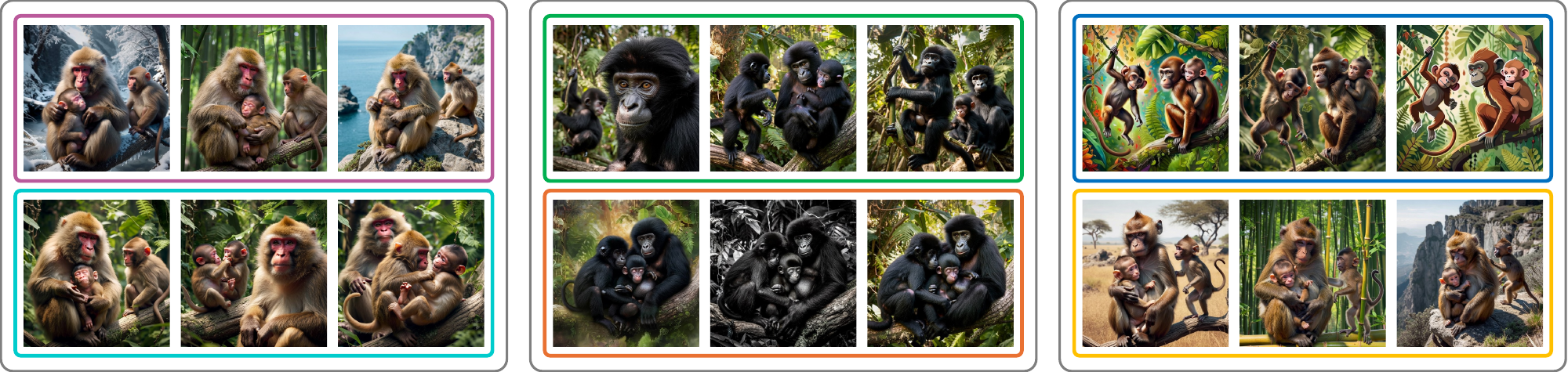}

    \vspace{0.9em}
    \hrule
    \vspace{1em}

    {\raggedright \small \textbf{User Prompt:} A man in uniform riding a horse. \par}
    \vspace{0.2em}
    \includegraphics[width=\textwidth]{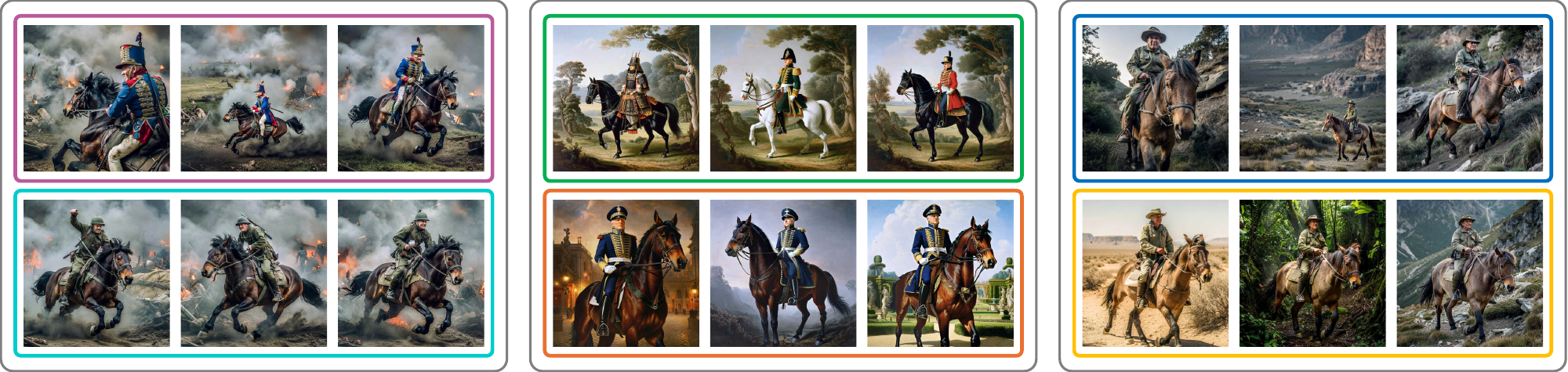}

    \caption{
    \textbf{Additional structured diversity results.} For each user prompt, outer gray panels group images derived from the same initial scene. Colored boxes distinguish sibling branches (parallel variations that share the same parent but differ from one another by a single semantic aspect).}
    \label{fig:results_additional}
\end{figure*} 

\begin{figure*}
    \centering
    {\raggedright \small \textbf{User Prompt:} A group of people at a sports event. \par}
    \vspace{0.2em}
    \includegraphics[width=\textwidth]{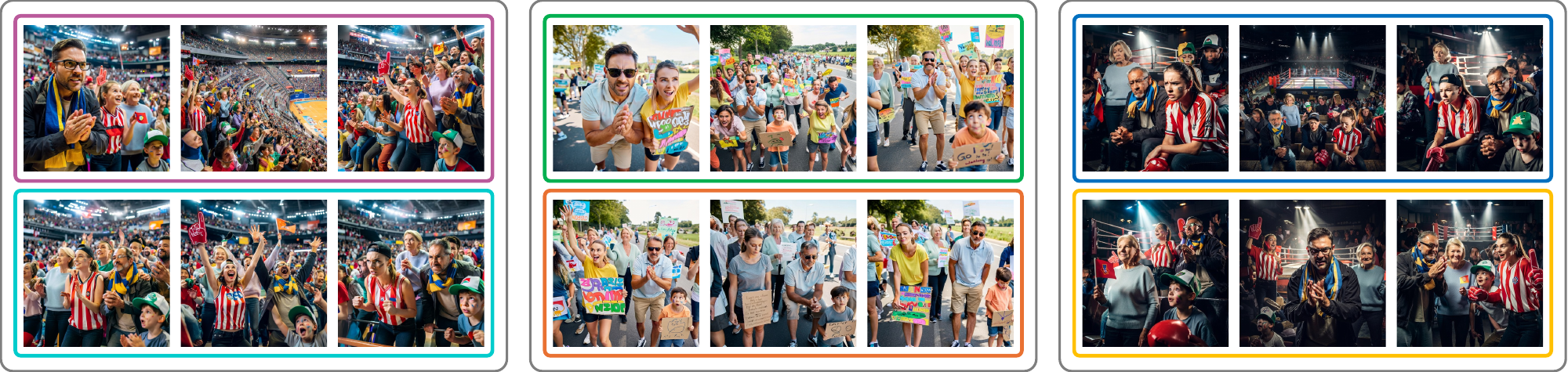}

    \vspace{0.9em}
    \hrule
    \vspace{1em}

    {\raggedright \small \textbf{User Prompt:} A robot and a scarecrow in a field. \par}
    \vspace{0.2em}
    \includegraphics[width=\textwidth]{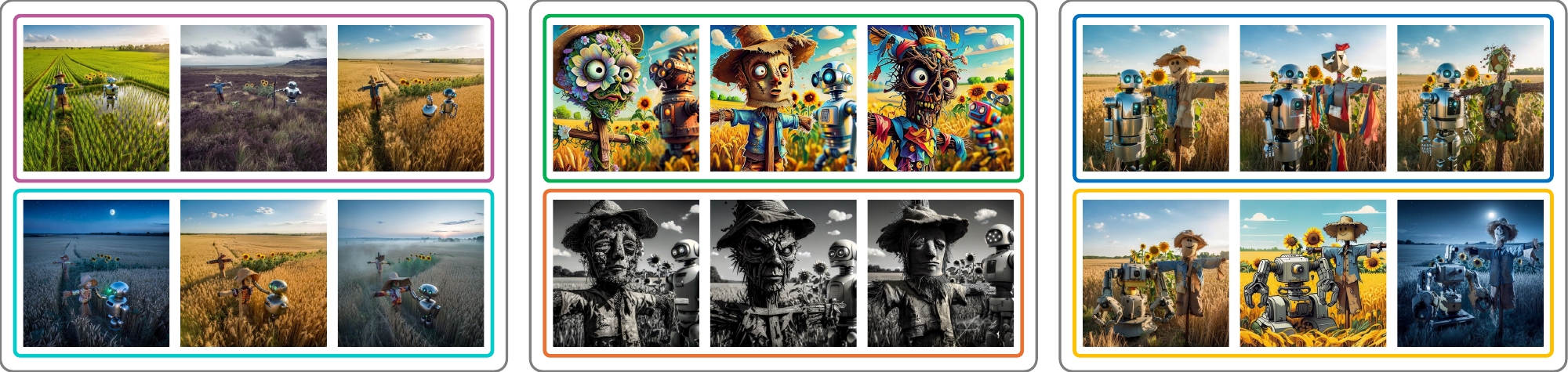}

    \vspace{0.9em}
    \hrule
    \vspace{1em}

    {\raggedright \small \textbf{User Prompt:} A doll on a shelf. \par}
    \vspace{0.2em}
    \includegraphics[width=\textwidth]{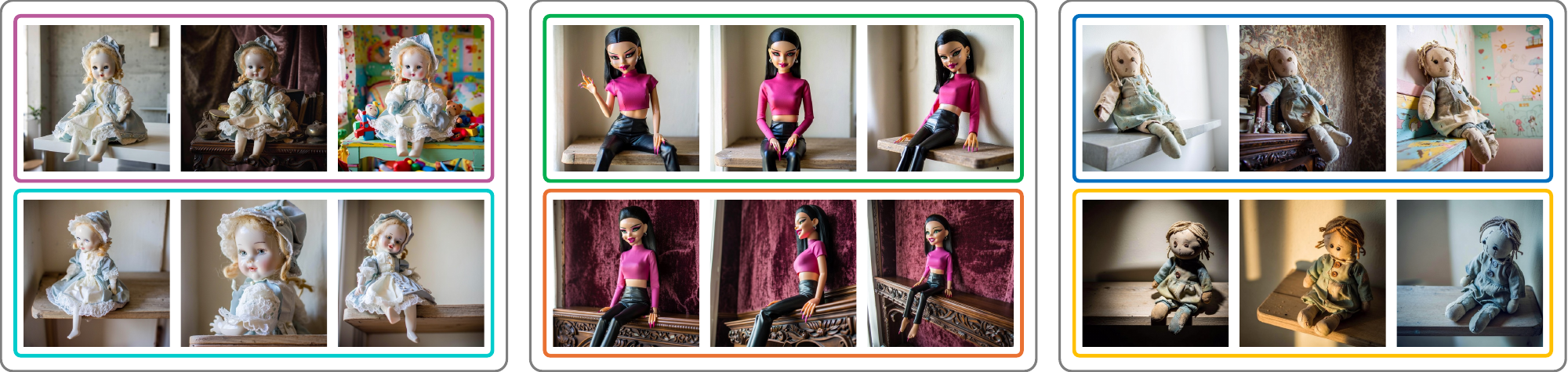}

    \vspace{0.9em}
    \hrule
    \vspace{1em}
    
    {\raggedright \small \textbf{User Prompt:} A boat passes by waterfront houses flanked by trees. \par}
    \vspace{0.2em}
    \includegraphics[width=\textwidth]{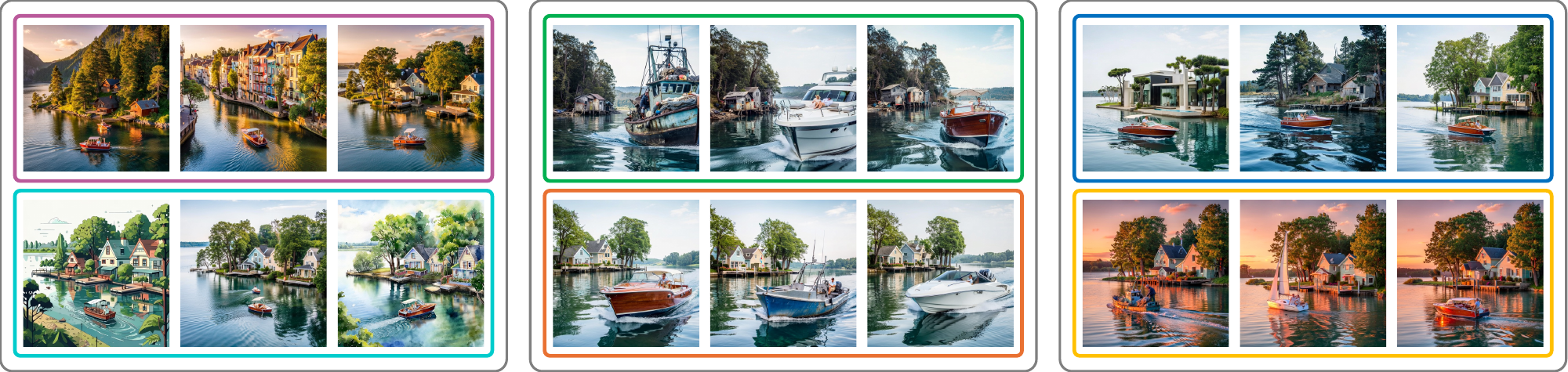}

    \caption{
    \textbf{Additional structured diversity results.} For each user prompt, outer gray panels group images derived from the same initial scene. Colored boxes distinguish sibling branches (parallel variations that share the same parent but differ from one another by a single semantic aspect).}
    \label{fig:results_sup}
\end{figure*} 

\begin{figure*}[t]
    \centering
    \setlength{\tabcolsep}{0.003\textwidth}
    {\small
    \begin{tabular}{c c c c c c c}
        Ours & VLM Seeding & Post-Hoc Opt. & Post-Hoc Opt. Temp. & CADS & Guidance Interval & Power-Law CFG \\
        
        \includegraphics[width=0.133\textwidth]{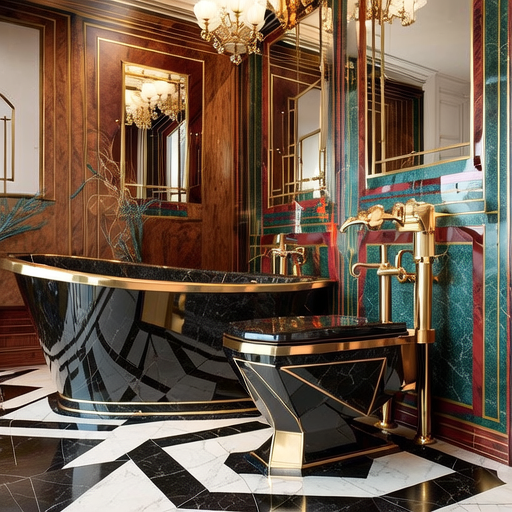} &
        \includegraphics[width=0.133\textwidth]{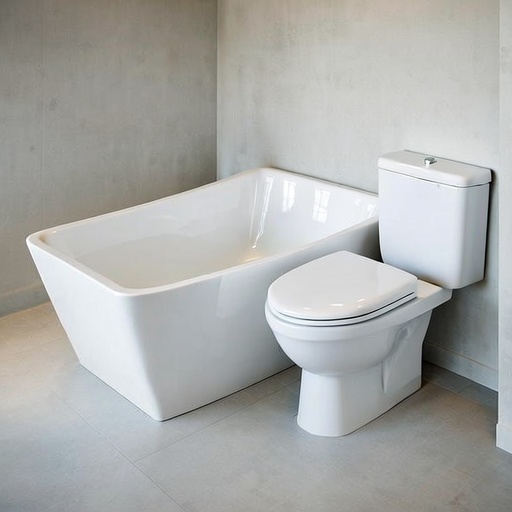} &
        \includegraphics[width=0.133\textwidth]{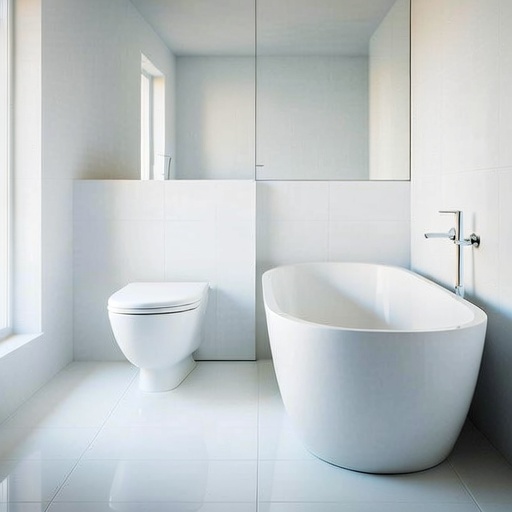} &
        \includegraphics[width=0.133\textwidth]{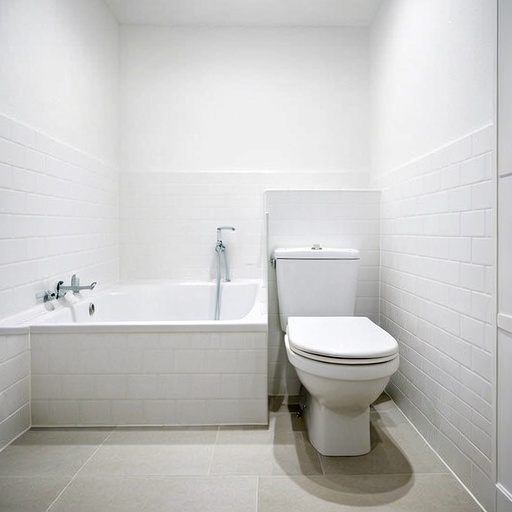} &
        \includegraphics[width=0.133\textwidth]{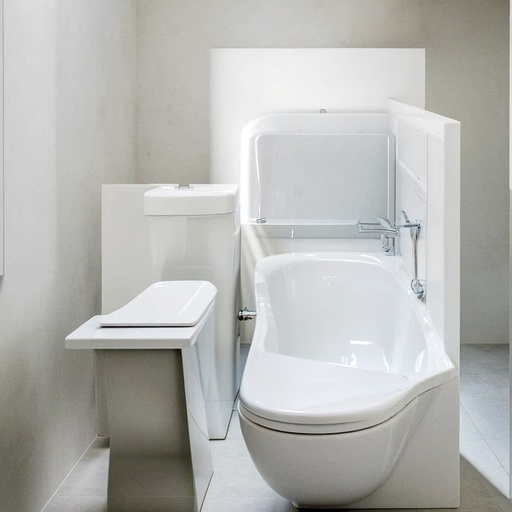} &
        \includegraphics[width=0.133\textwidth]{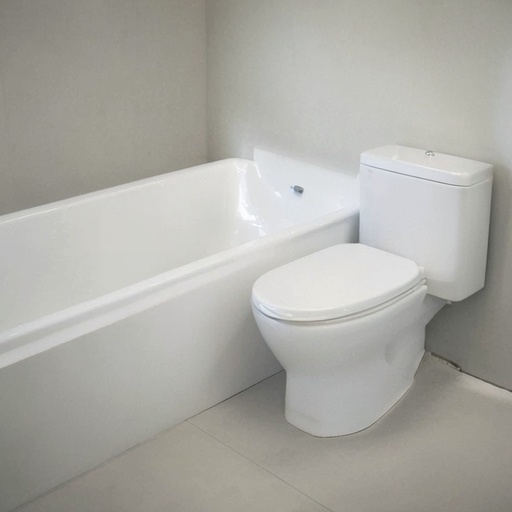} &
        \includegraphics[width=0.133\textwidth]{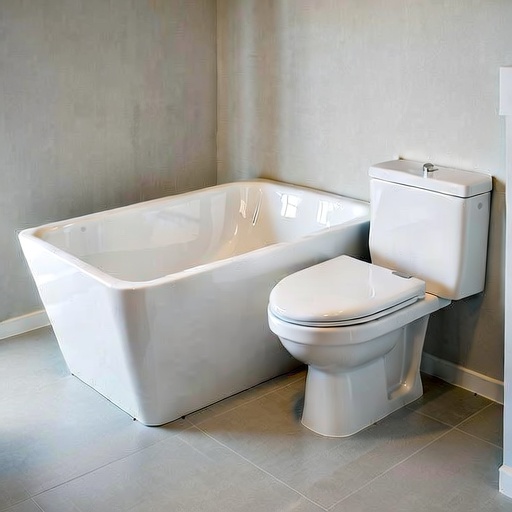} \\

        \includegraphics[width=0.133\textwidth]{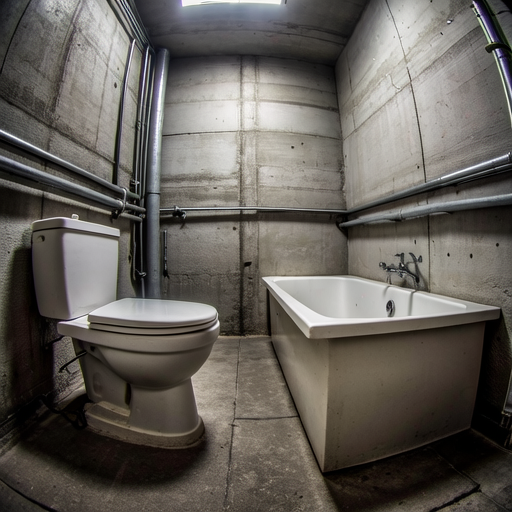} &
        \includegraphics[width=0.133\textwidth]{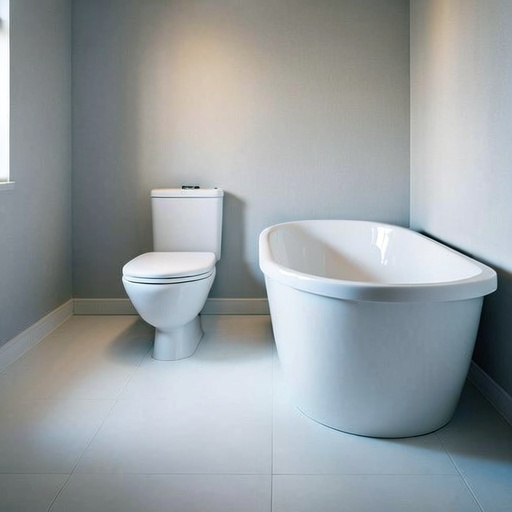} &
        \includegraphics[width=0.133\textwidth]{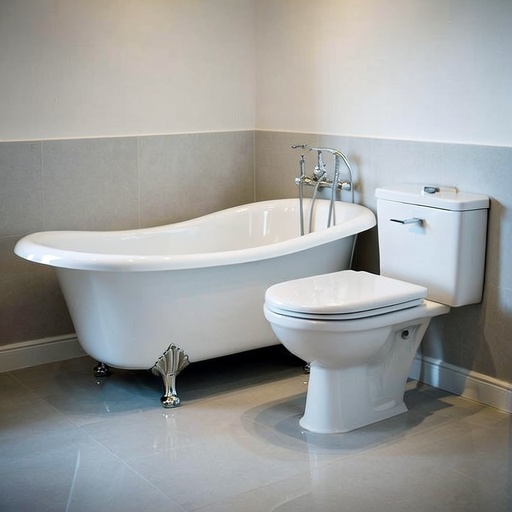} &
        \includegraphics[width=0.133\textwidth]{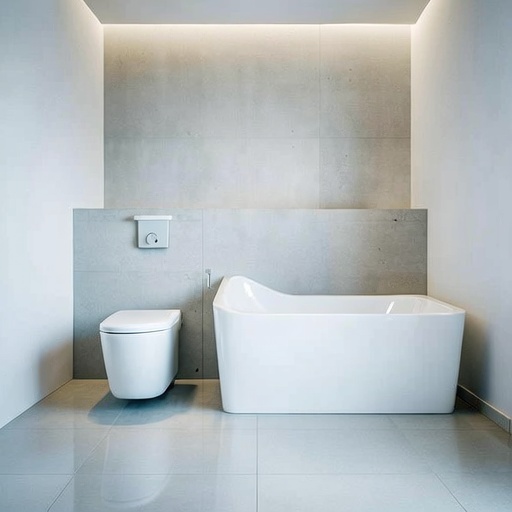} &
        \includegraphics[width=0.133\textwidth]{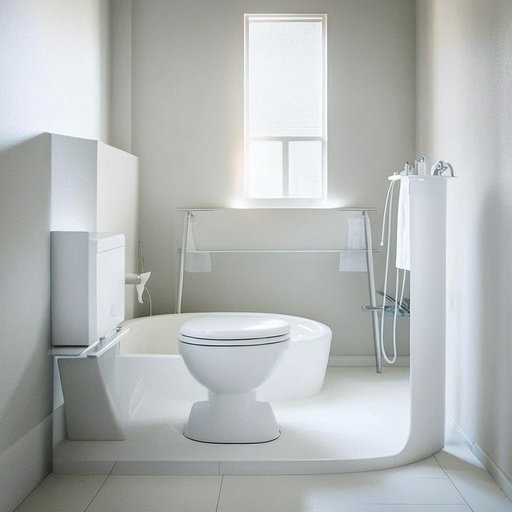} &
        \includegraphics[width=0.133\textwidth]{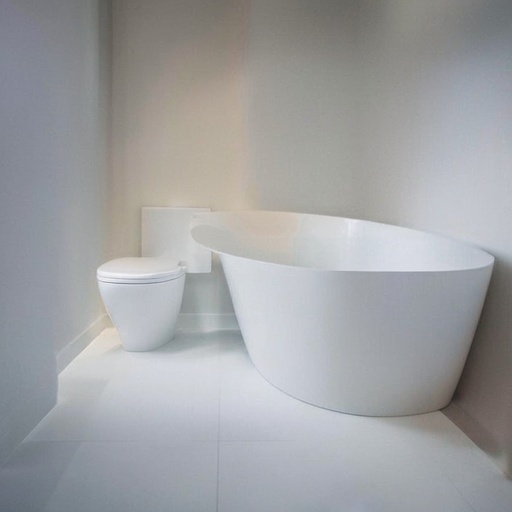} &
        \includegraphics[width=0.133\textwidth]{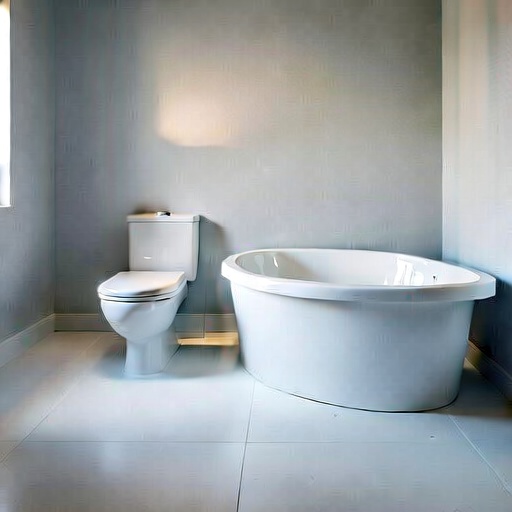} \\

        \includegraphics[width=0.133\textwidth]{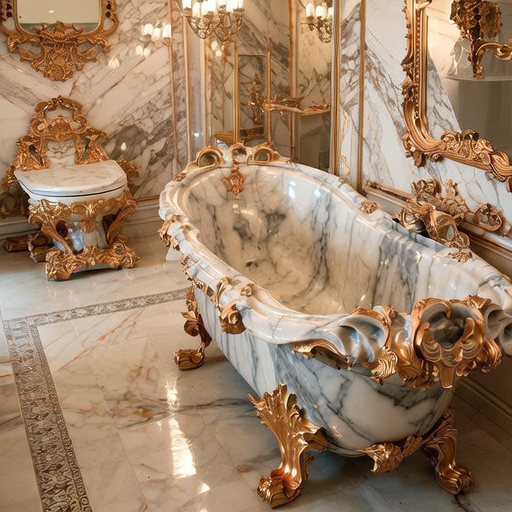} &
        \includegraphics[width=0.133\textwidth]{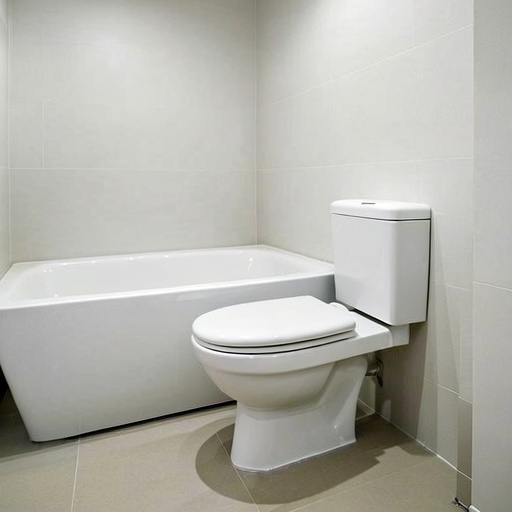} &
        \includegraphics[width=0.133\textwidth]{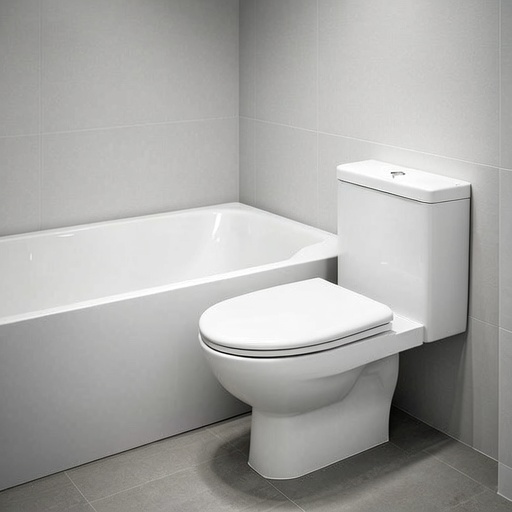} &
        \includegraphics[width=0.133\textwidth]{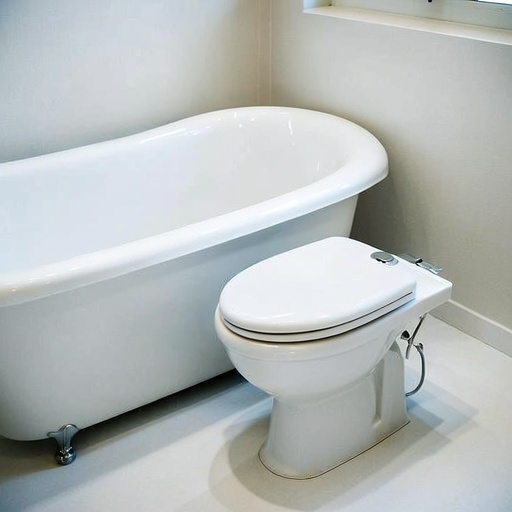} &
        \includegraphics[width=0.133\textwidth]{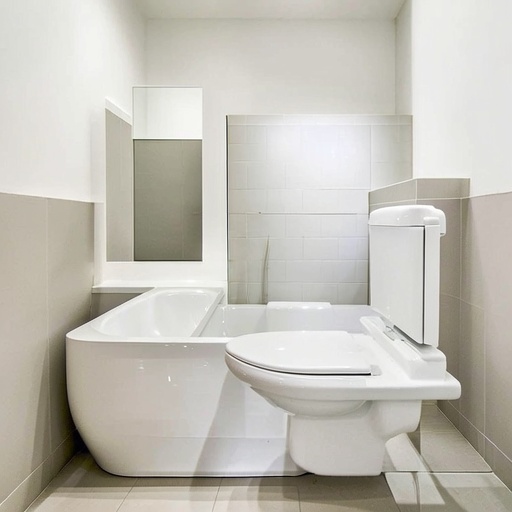} &
        \includegraphics[width=0.133\textwidth]{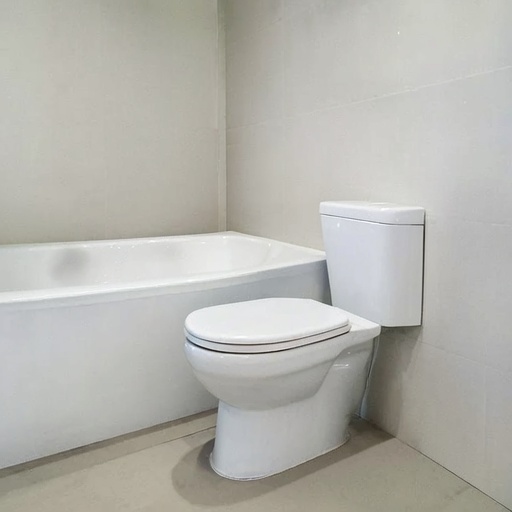} &
        \includegraphics[width=0.133\textwidth]{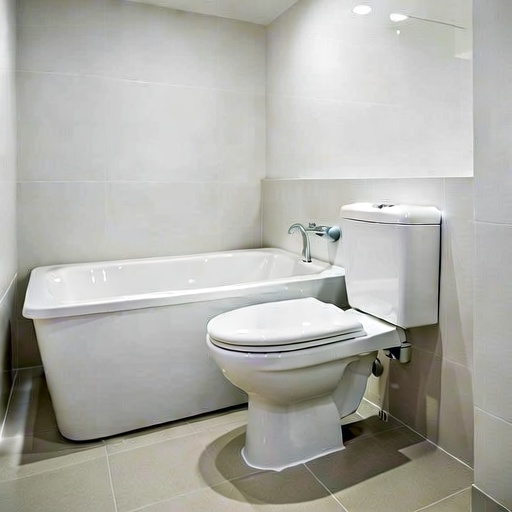} \\

        \includegraphics[width=0.133\textwidth]{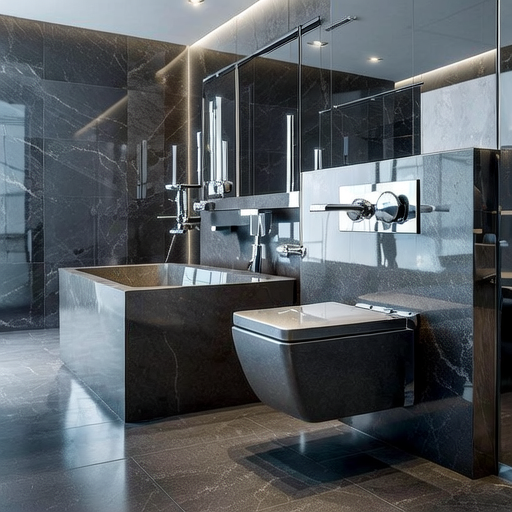} &
        \includegraphics[width=0.133\textwidth]{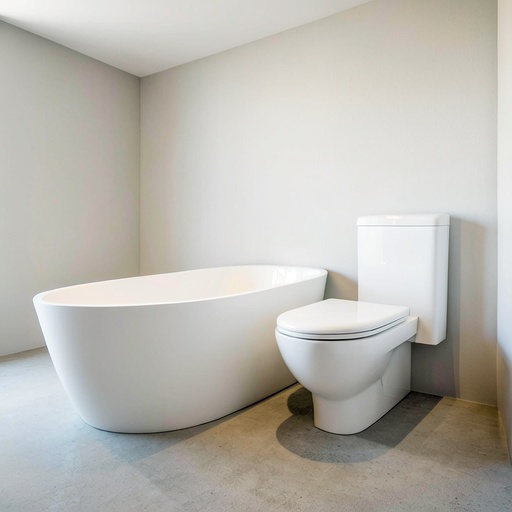} &
        \includegraphics[width=0.133\textwidth]{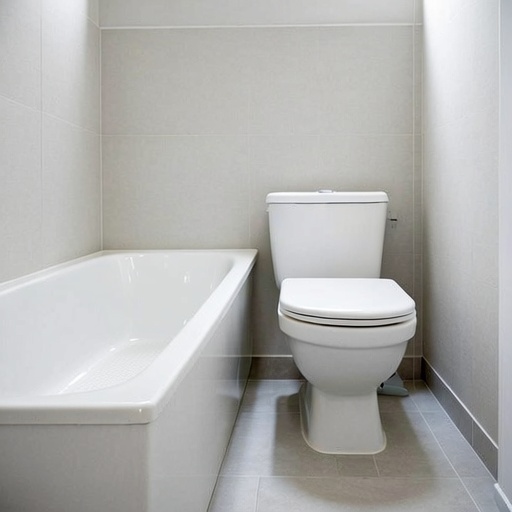} &
        \includegraphics[width=0.133\textwidth]{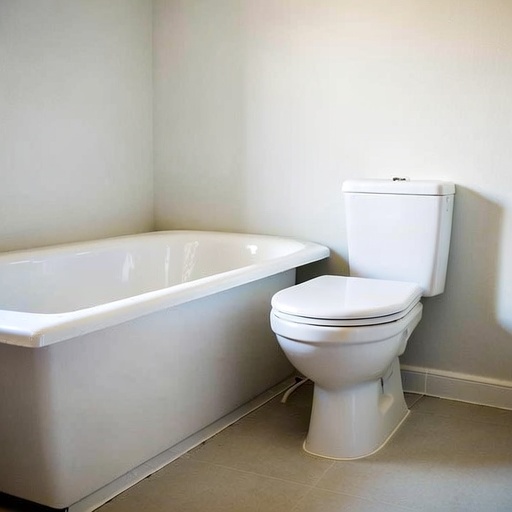} &
        \includegraphics[width=0.133\textwidth]{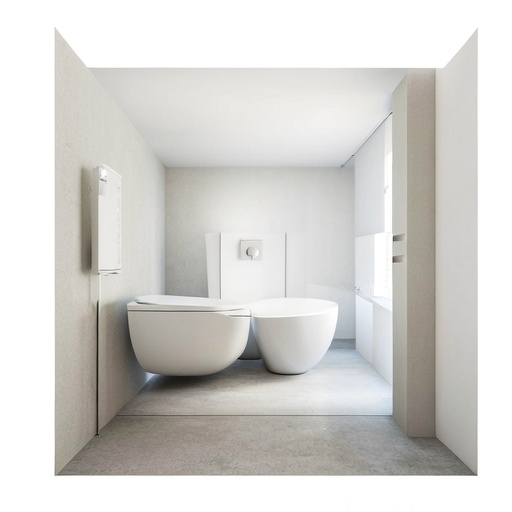} &
        \includegraphics[width=0.133\textwidth]{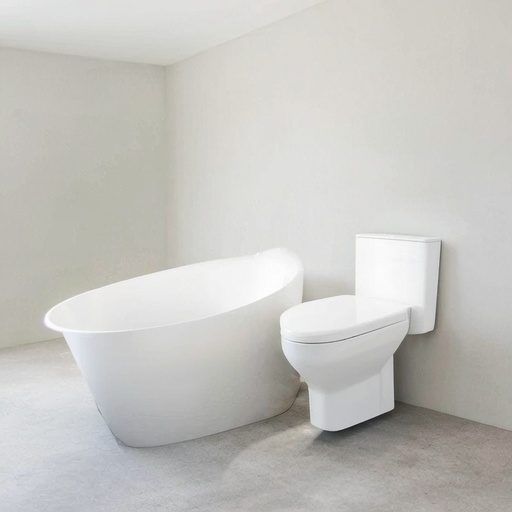} &
        \includegraphics[width=0.133\textwidth]{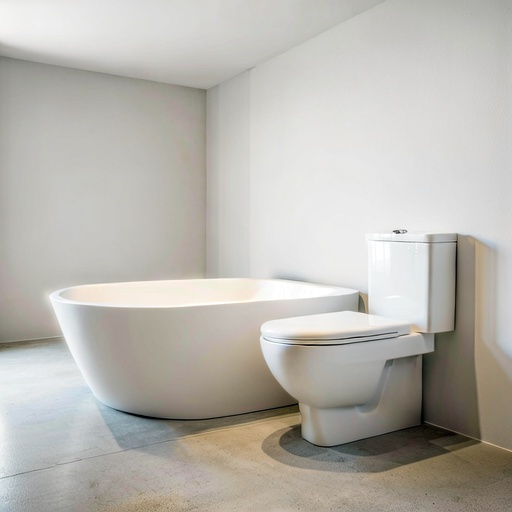} \\

    \end{tabular}
    \vspace{-6pt}
    }
    \captionof{figure}{\textbf{Qualitative comparison on the prompt:} \textit{“A toilet sits next to a bathtub in an empty bathroom.”} Columns 2 and 5-7 report results using consecutive seeds with hyperparameters optimized for diversity. Columns 3-4 display the most diverse subset of four images selected from a larger candidate pool.
     While baseline methods exhibit limited variation, our method (column 1) successfully presents distinct and coherent interpretations. Our approach introduces significant semantic shifts by varying the materials, colors, and architectural styles of the scene, ranging from luxury black-and-gold marble and industrial concrete to ornate classical designs.}
    \vspace{-8pt}
    \label{fig:comparisons_bath}
\end{figure*}

\begin{figure*}[t]
    \centering
    \setlength{\tabcolsep}{0.003\textwidth}
    {\small
    \begin{tabular}{c c c c c c c}
        Ours & VLM Seeding & Post-Hoc Opt. & Post-Hoc Opt. Temp. & CADS & Guidance Interval & Power-Law CFG \\
        
        \includegraphics[width=0.135\textwidth]{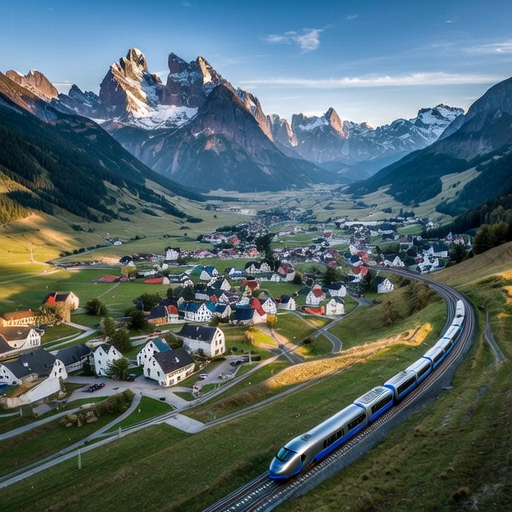} &
        \includegraphics[width=0.135\textwidth]{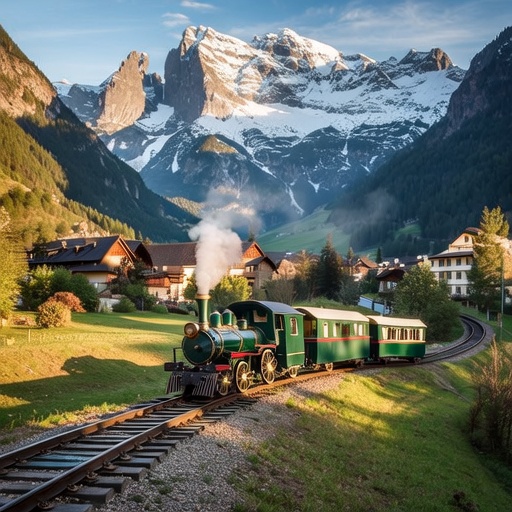} &
        \includegraphics[width=0.135\textwidth]{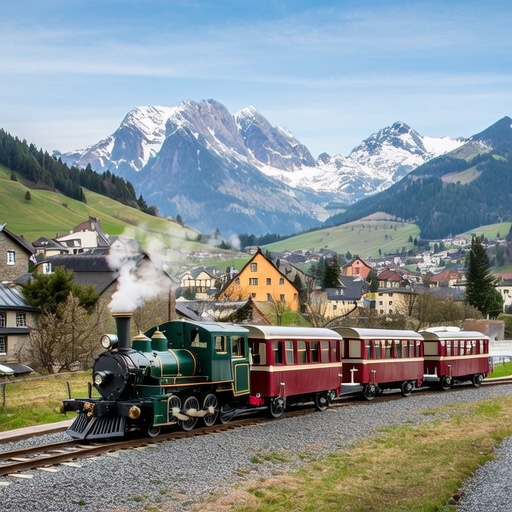} &
        \includegraphics[width=0.135\textwidth]{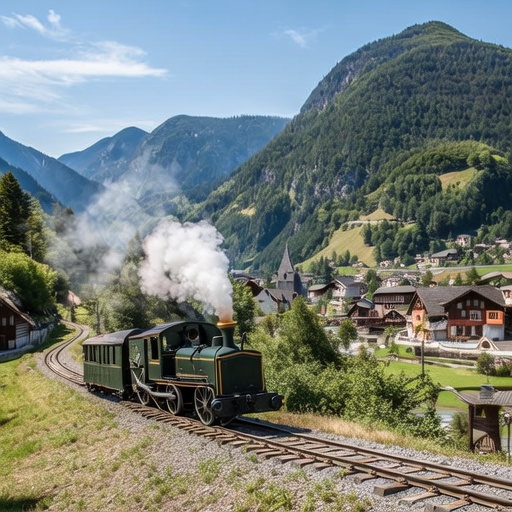} &
        \includegraphics[width=0.135\textwidth]{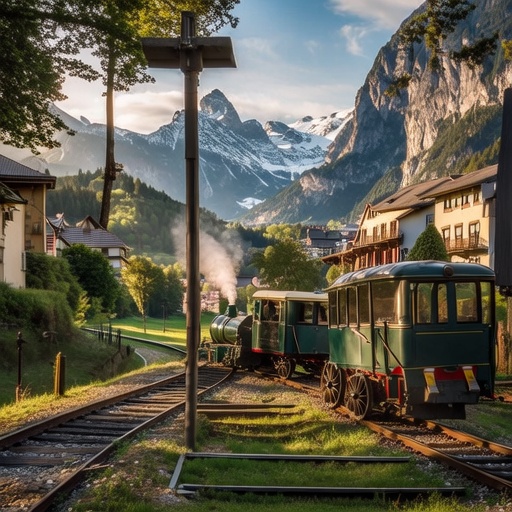} &
        \includegraphics[width=0.135\textwidth]{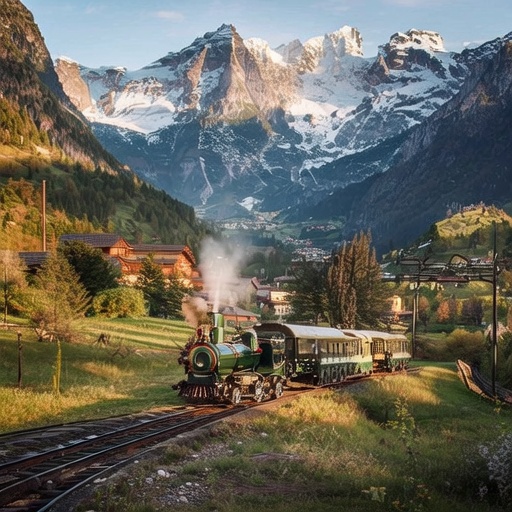} &
        \includegraphics[width=0.135\textwidth]{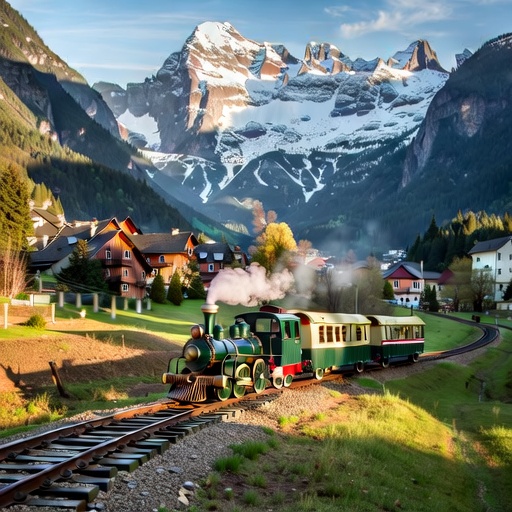} \\

        \includegraphics[width=0.135\textwidth]{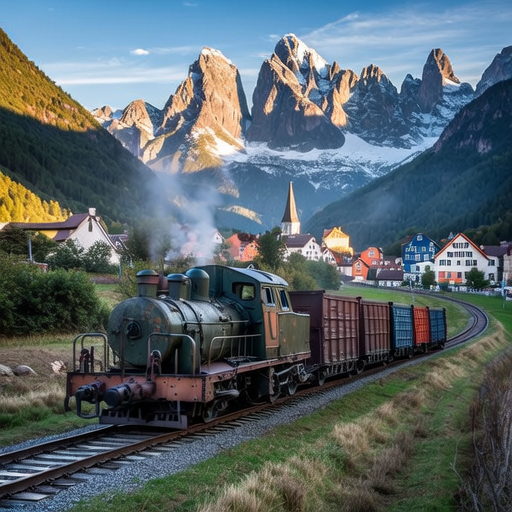} &
        \includegraphics[width=0.135\textwidth]{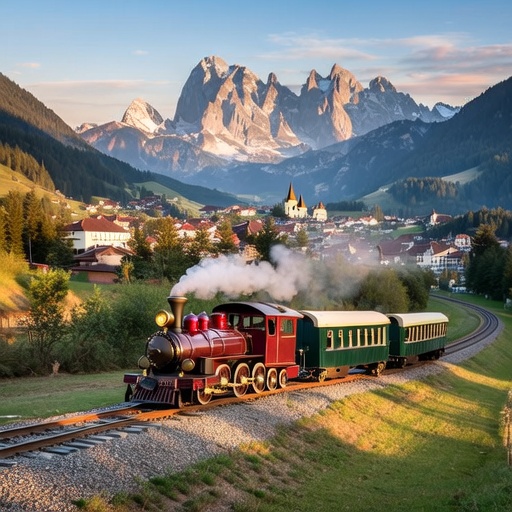} &
        \includegraphics[width=0.135\textwidth]{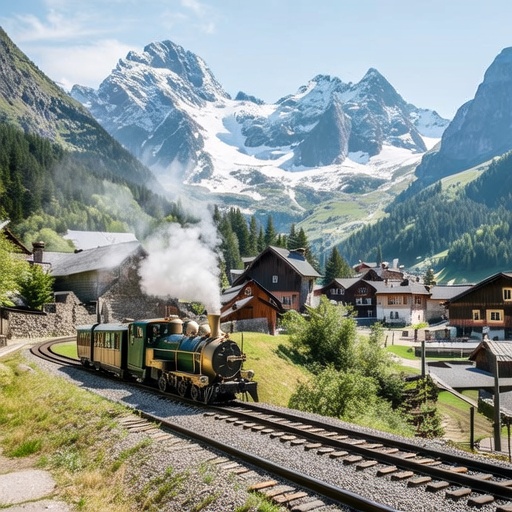} &
        \includegraphics[width=0.135\textwidth]{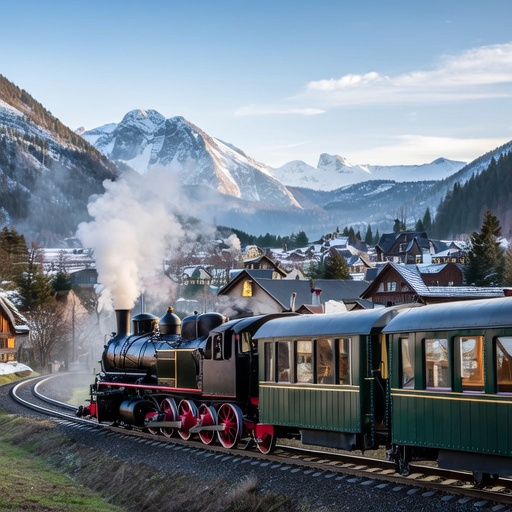} &
        \includegraphics[width=0.135\textwidth]{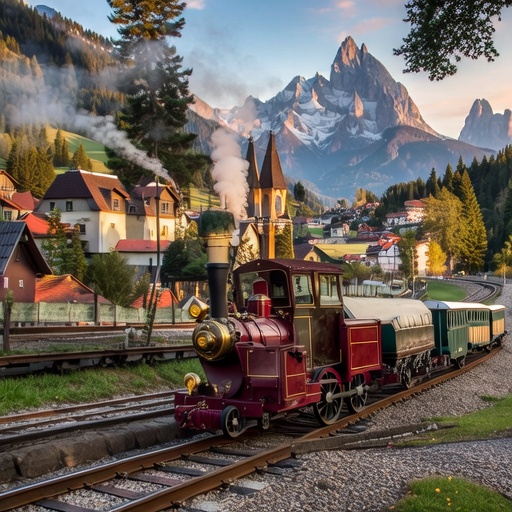} &
        \includegraphics[width=0.135\textwidth]{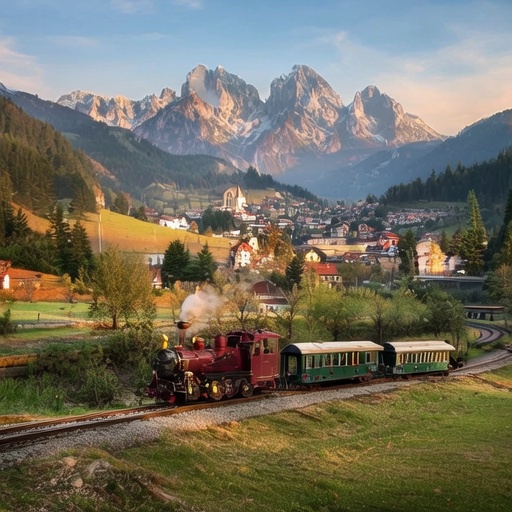} &
        \includegraphics[width=0.135\textwidth]{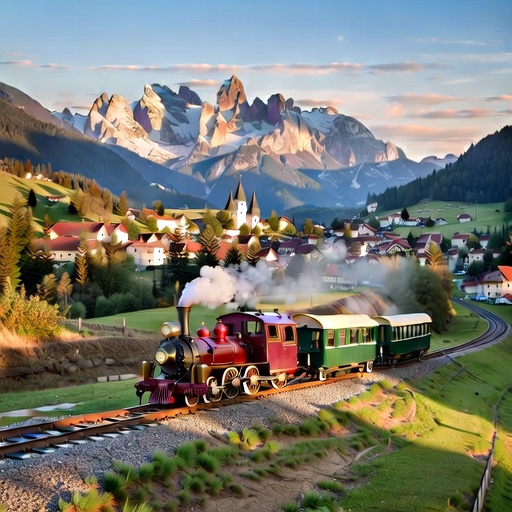} \\

        \includegraphics[width=0.135\textwidth]{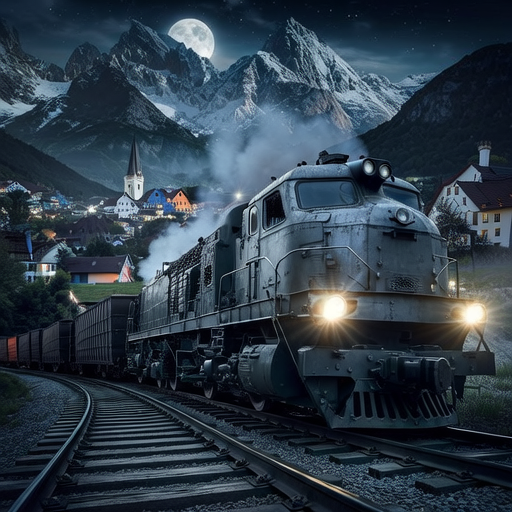} &
        \includegraphics[width=0.135\textwidth]{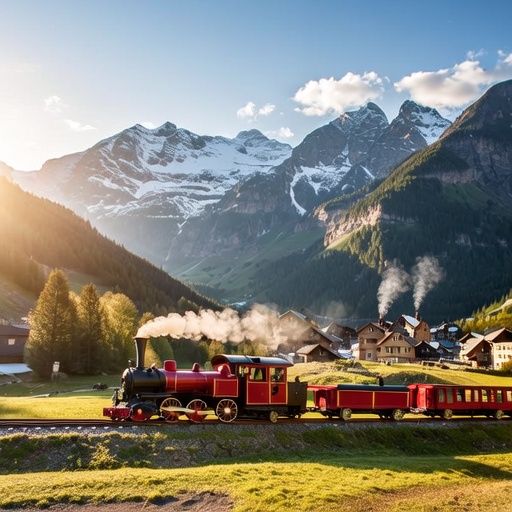} &
        \includegraphics[width=0.135\textwidth]{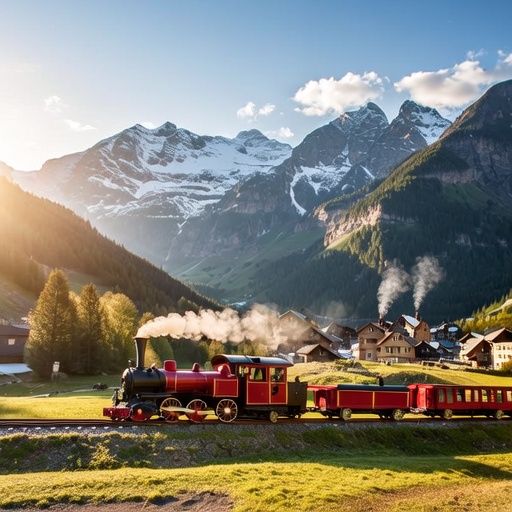} &
        \includegraphics[width=0.135\textwidth]{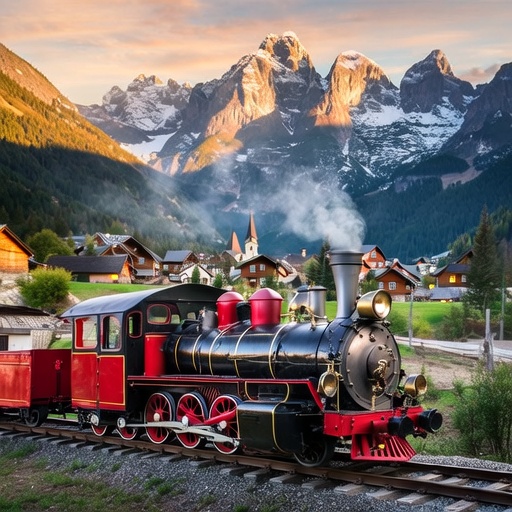} &
        \includegraphics[width=0.135\textwidth]{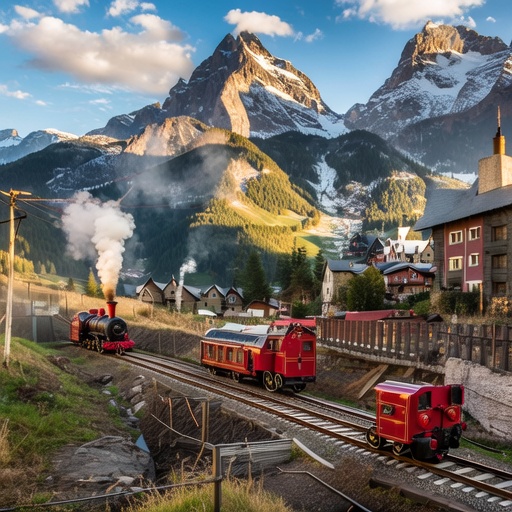} &
        \includegraphics[width=0.135\textwidth]{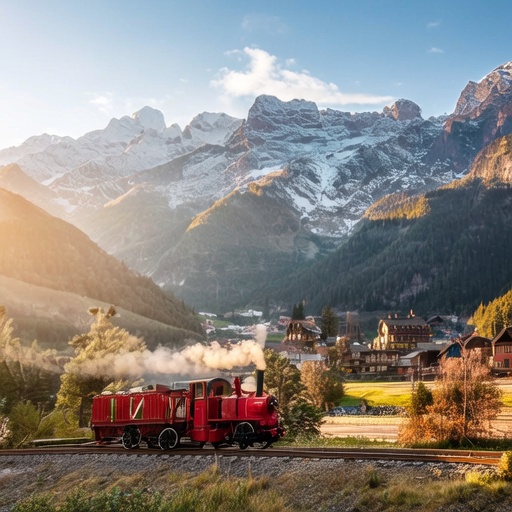} &
        \includegraphics[width=0.135\textwidth]{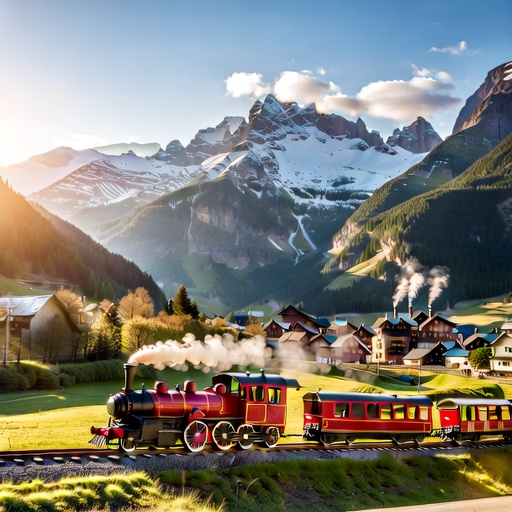} \\

        \includegraphics[width=0.135\textwidth]{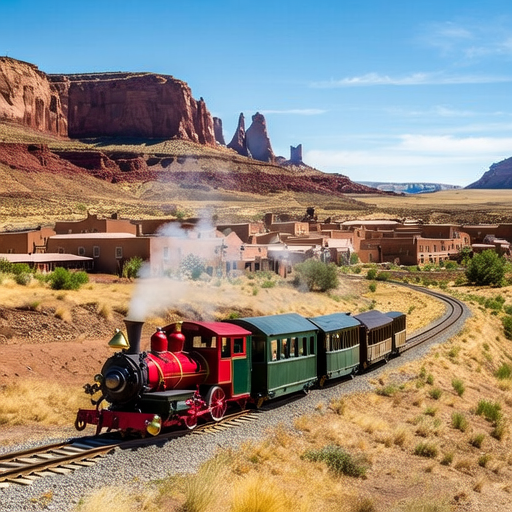} &
        \includegraphics[width=0.135\textwidth]{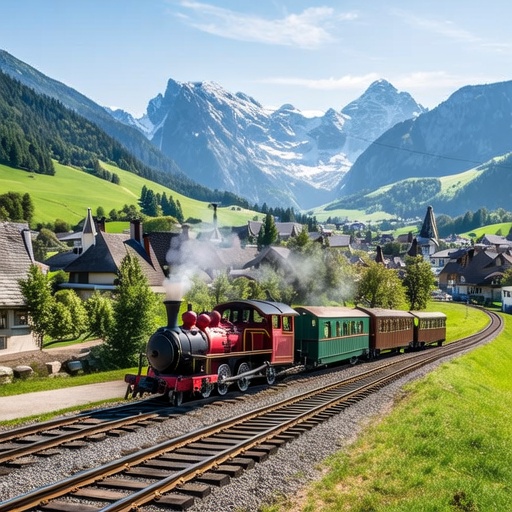} &
        \includegraphics[width=0.135\textwidth]{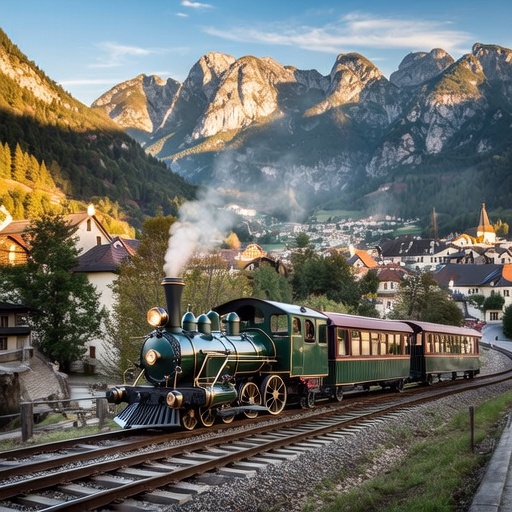} &
        \includegraphics[width=0.135\textwidth]{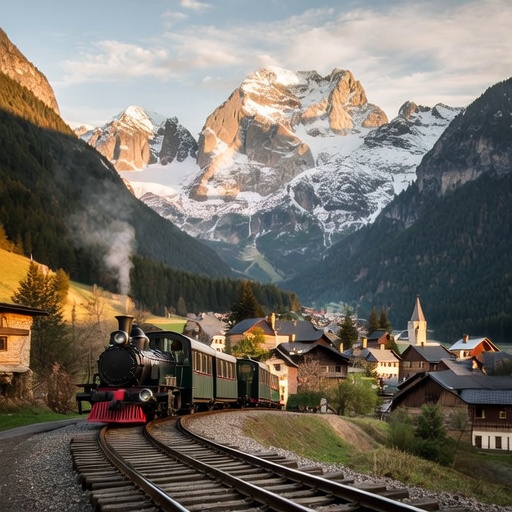} &
        \includegraphics[width=0.135\textwidth]{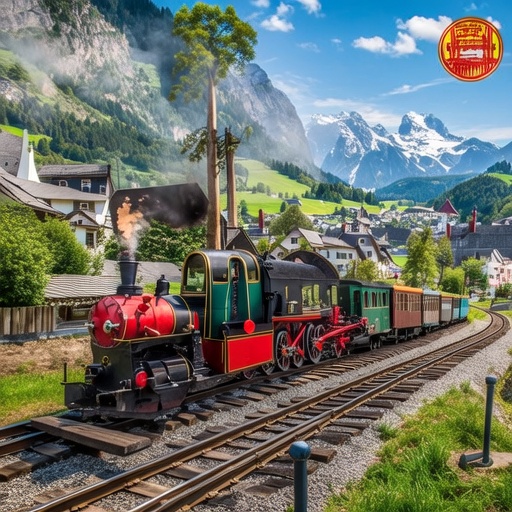} &
        \includegraphics[width=0.135\textwidth]{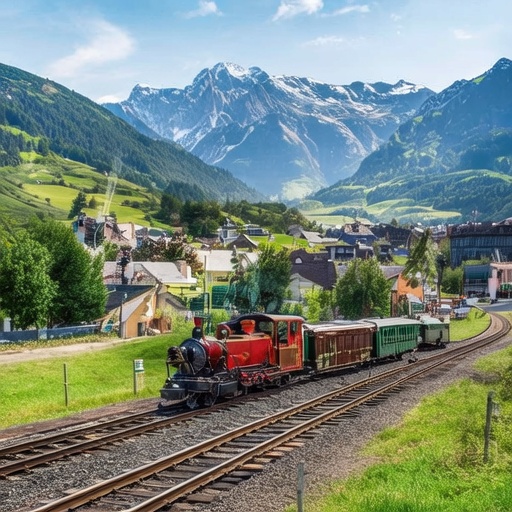} &
        \includegraphics[width=0.135\textwidth]{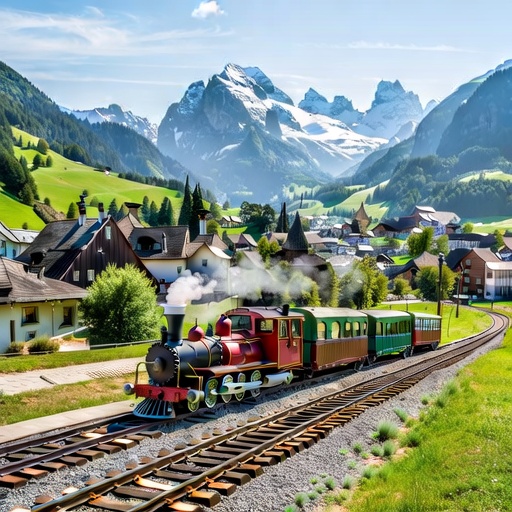} \\

    \end{tabular}
    }
    \captionof{figure}{\textbf{Qualitative comparison on the prompt:} \textit{“A small train moving along the tracks with a mountain town in the background.”} Columns 2 and 5-7 report results using consecutive seeds with hyperparameters optimized for diversity. Columns 3-4 display the most diverse subset of four images selected from a larger candidate pool.
     While baseline methods exhibit limited variation, our method (column 1) successfully presents distinct and coherent interpretations. Examples include modifying the core object (row 1 and 2: switching to a modern electric train and to a goods train), the temporal setting (row 3: shifting to a night scene), and the environment (row 3: relocating to a desert landscape).}
    \label{fig:comparisons_train}
\end{figure*}

\begin{figure*}[t]
    \centering
    \setlength{\tabcolsep}{0.003\textwidth}
    {\small
    \begin{tabular}{c c c c c c c}
        Ours & VLM Seeding & Post-Hoc Opt. & Post-Hoc Opt. Temp. & CADS & Guidance Interval & Power-Law CFG \\
        
        \includegraphics[width=0.135\textwidth]{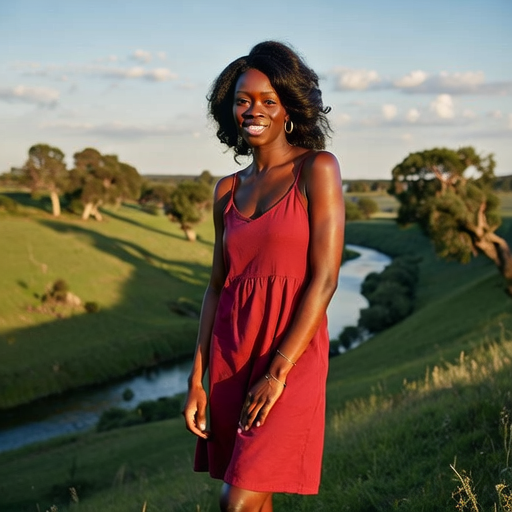} &
        \includegraphics[width=0.135\textwidth]{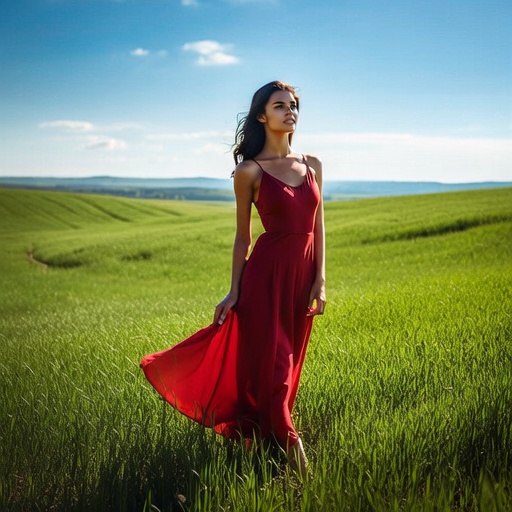} &
        \includegraphics[width=0.135\textwidth]{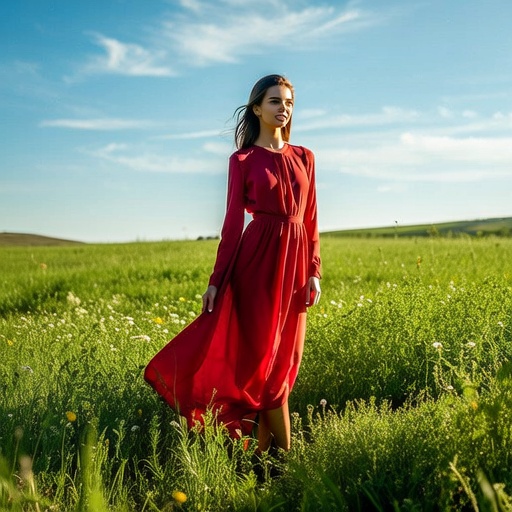} &
        \includegraphics[width=0.135\textwidth]{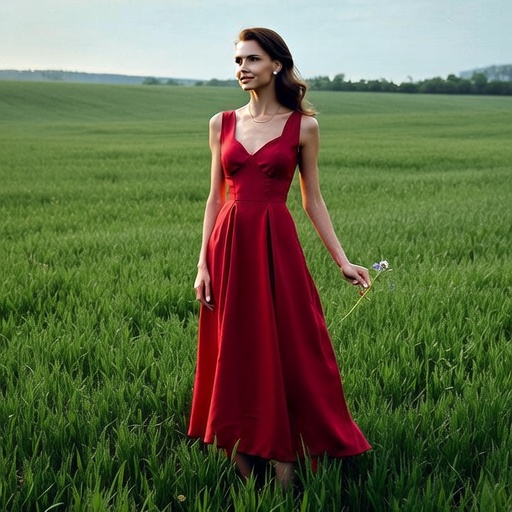} &
        \includegraphics[width=0.135\textwidth]{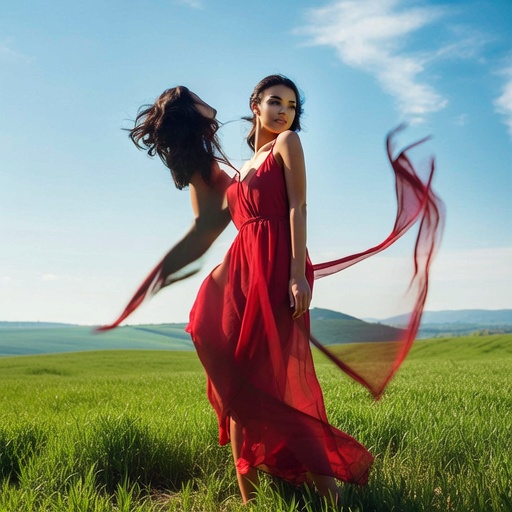} &
        \includegraphics[width=0.135\textwidth]{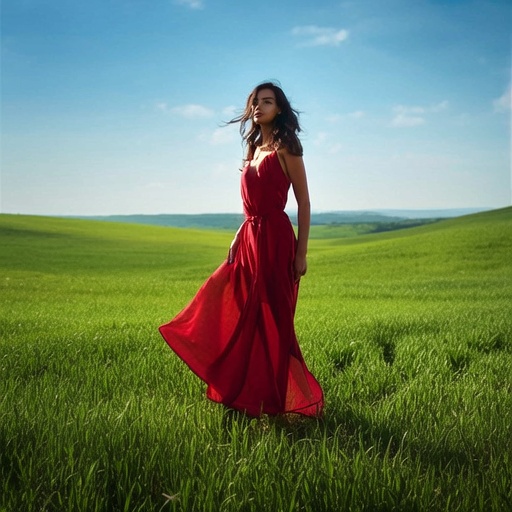} &
        \includegraphics[width=0.135\textwidth]{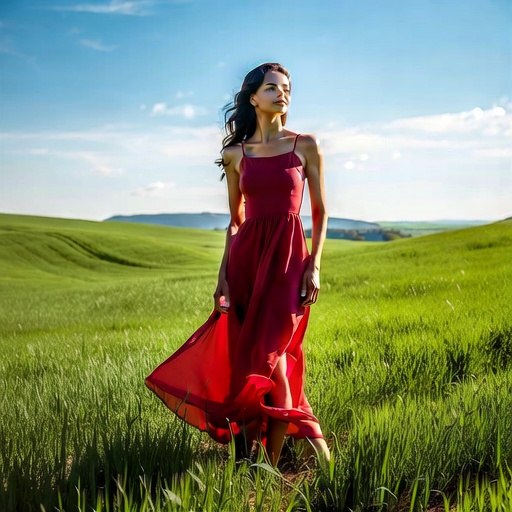} \\

        \includegraphics[width=0.135\textwidth]{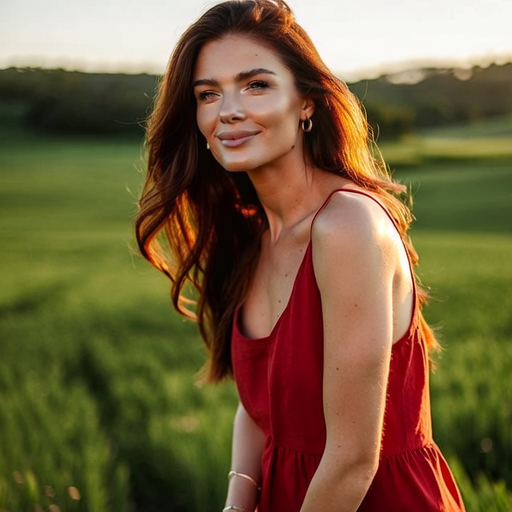} &
        \includegraphics[width=0.135\textwidth]{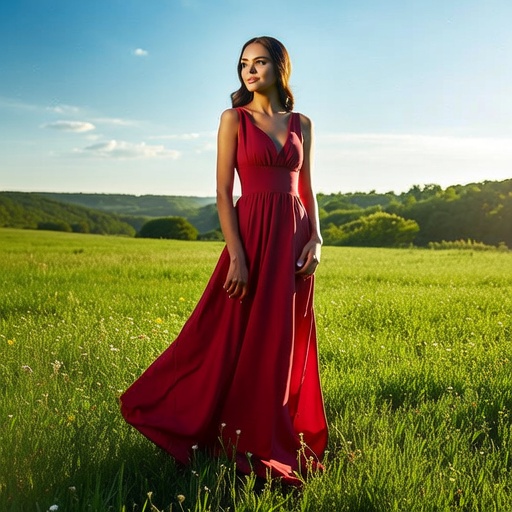} &
        \includegraphics[width=0.135\textwidth]{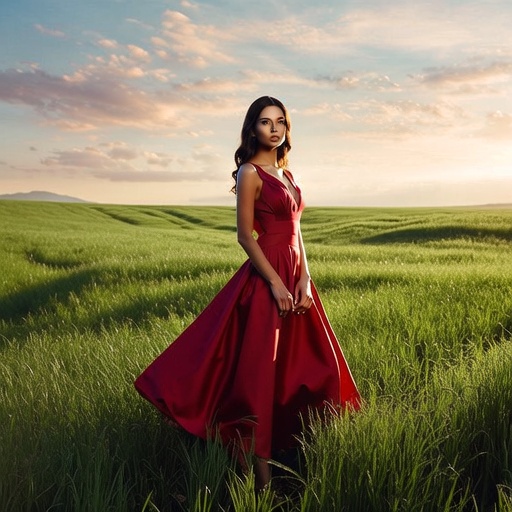} &
        \includegraphics[width=0.135\textwidth]{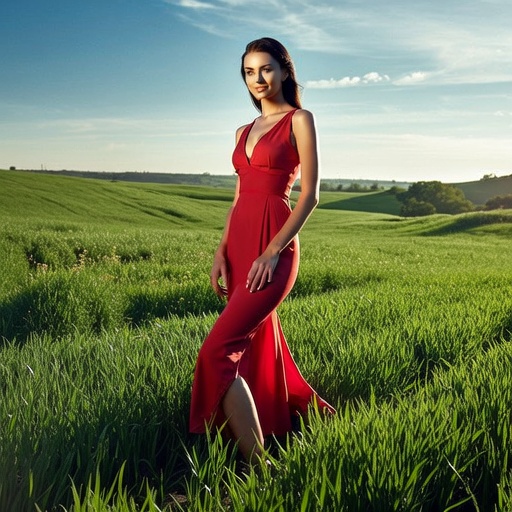} &
        \includegraphics[width=0.135\textwidth]{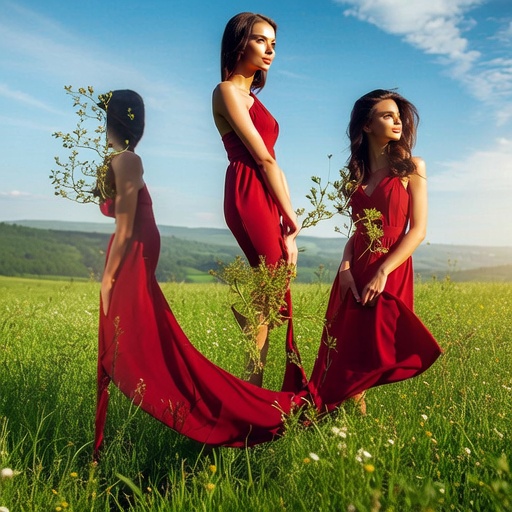} &
        \includegraphics[width=0.135\textwidth]{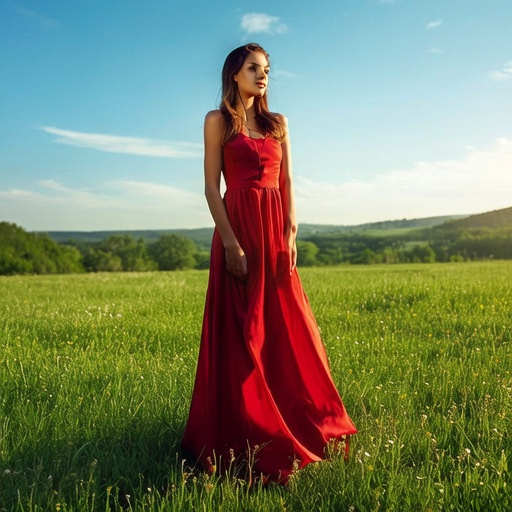} &
        \includegraphics[width=0.135\textwidth]{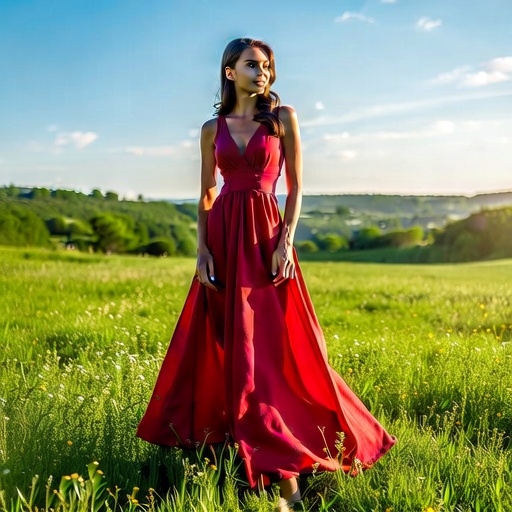} \\

        \includegraphics[width=0.135\textwidth]{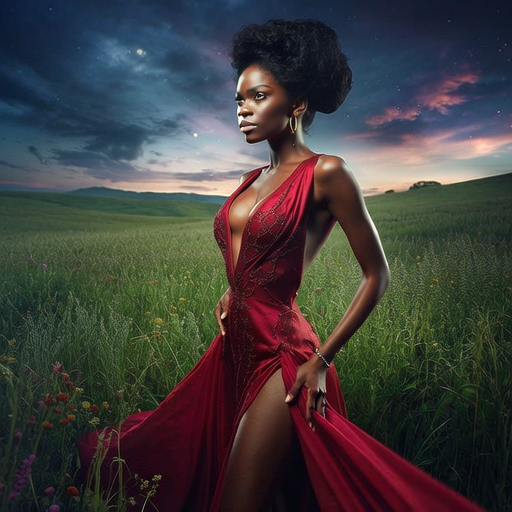} &
        \includegraphics[width=0.135\textwidth]{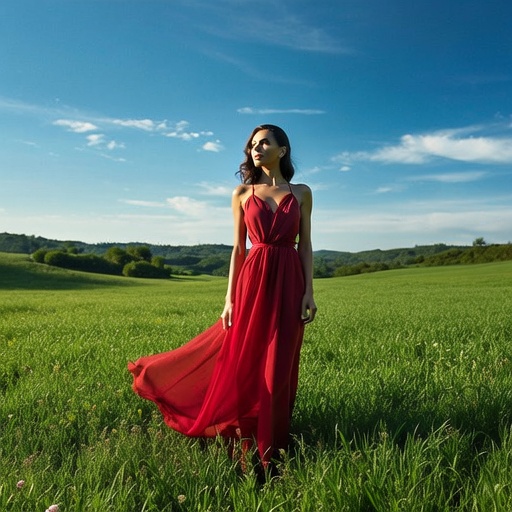} &
        \includegraphics[width=0.135\textwidth]{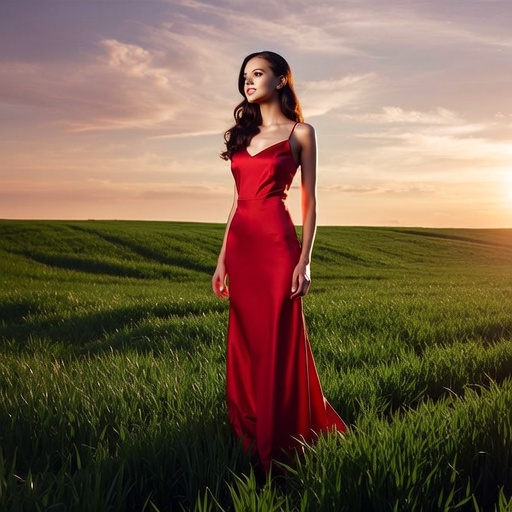} &
        \includegraphics[width=0.135\textwidth]{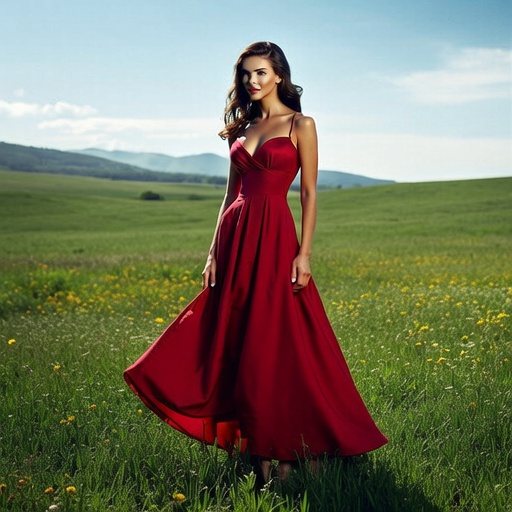} &
        \includegraphics[width=0.135\textwidth]{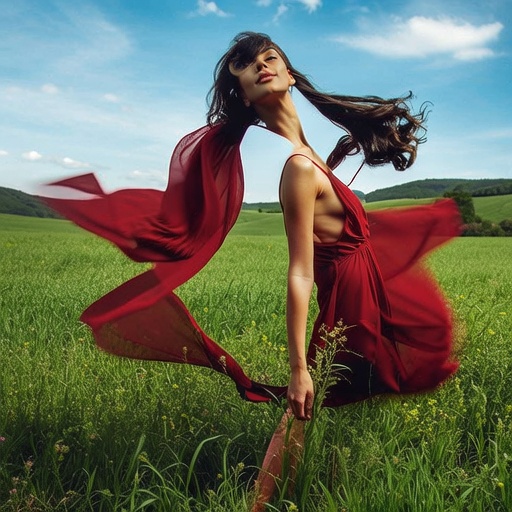} &
        \includegraphics[width=0.135\textwidth]{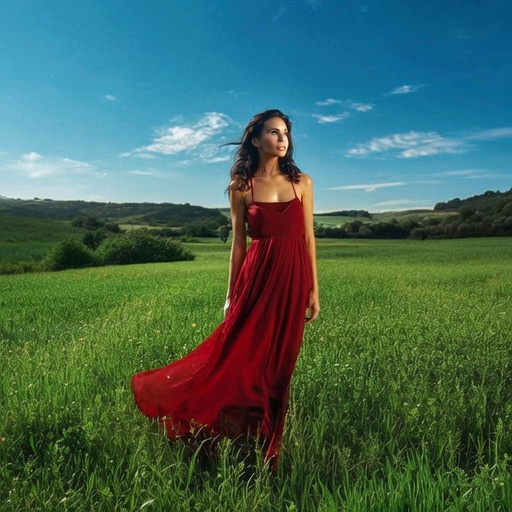} &
        \includegraphics[width=0.135\textwidth]{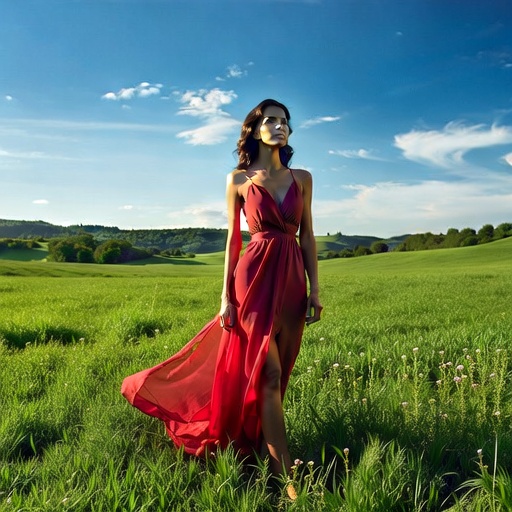} \\

        \includegraphics[width=0.135\textwidth]{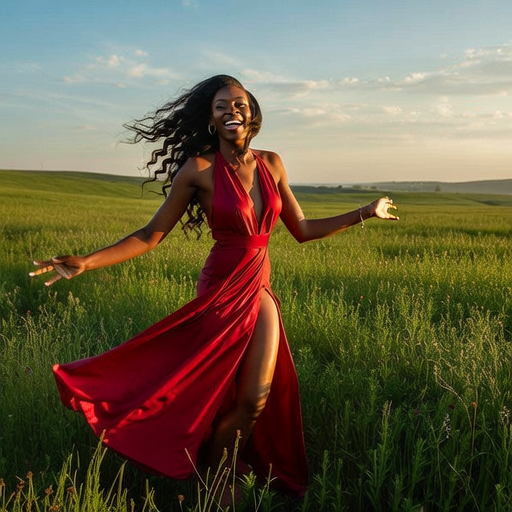} &
        \includegraphics[width=0.135\textwidth]{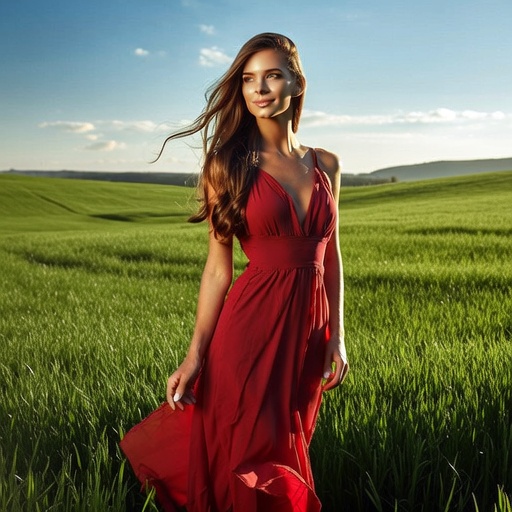} &
        \includegraphics[width=0.135\textwidth]{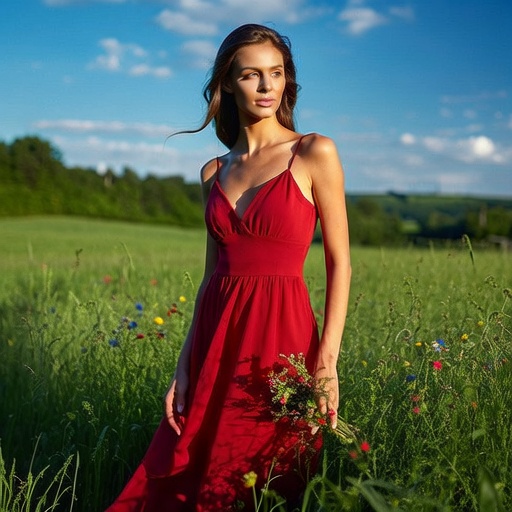} &
        \includegraphics[width=0.135\textwidth]{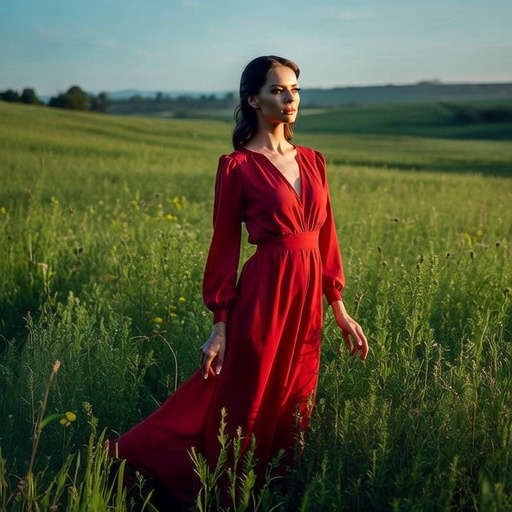} &
        \includegraphics[width=0.135\textwidth]{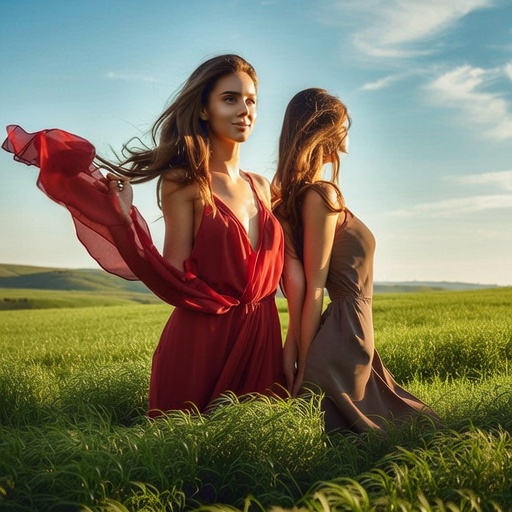} &
        \includegraphics[width=0.135\textwidth]{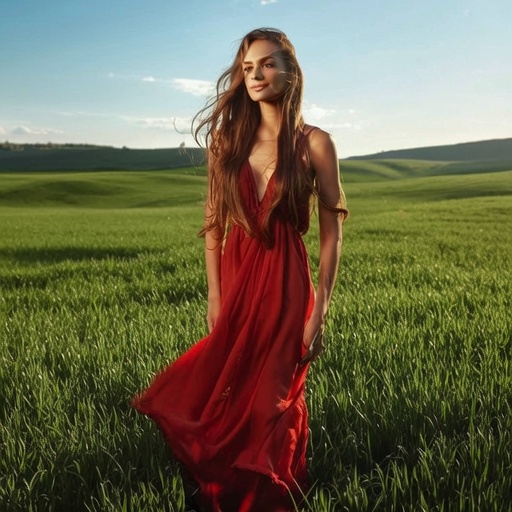} &
        \includegraphics[width=0.135\textwidth]{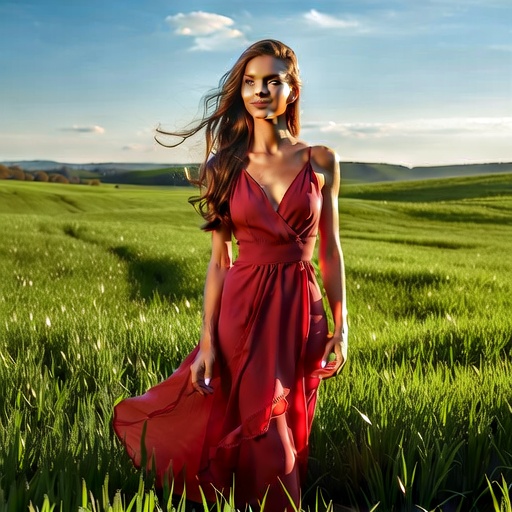} \\

    \end{tabular}
    \vspace{-6pt}
    }
    \captionof{figure}{\textbf{Qualitative comparison on the prompt:} \textit{“A woman in a red dress standing on top of a lush green field.”} Columns 2 and 5-7 report results using consecutive seeds with hyperparameters optimized for diversity. Columns 3-4 display the most diverse subset of four images selected from a larger candidate pool.
     While baseline methods exhibit limited variation, our method (column 1) successfully presents distinct and coherent interpretations. Examples include modifying the garment style (row 1: switching to a short dress), the camera framing (row 2: a close-up portrait), the lighting and temporal setting (row 3: a dramatic night scene), and the subject’s pose and activity (row 4: moving and dancing).}
    \vspace{-8pt}
    \label{fig:comparisons_woman}
\end{figure*}

\section{Implementation Details}
\label{implementation}

\begin{figure}
    \centering
    \includegraphics[width=\linewidth]{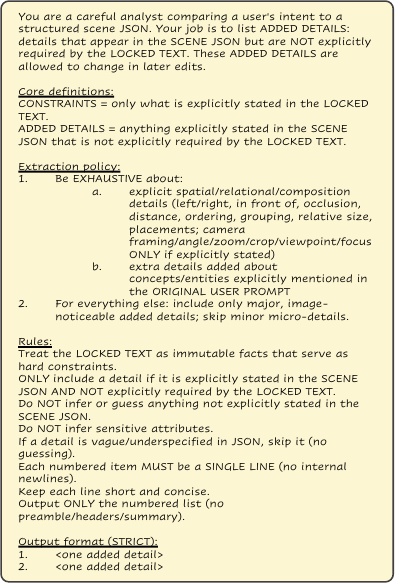}
    \caption{\textbf{Context Analyst System Prompt}}
    \label{fig:context_analyst_prompt}
\end{figure}

\begin{figure}%
    \centering
    \includegraphics[width=\linewidth]{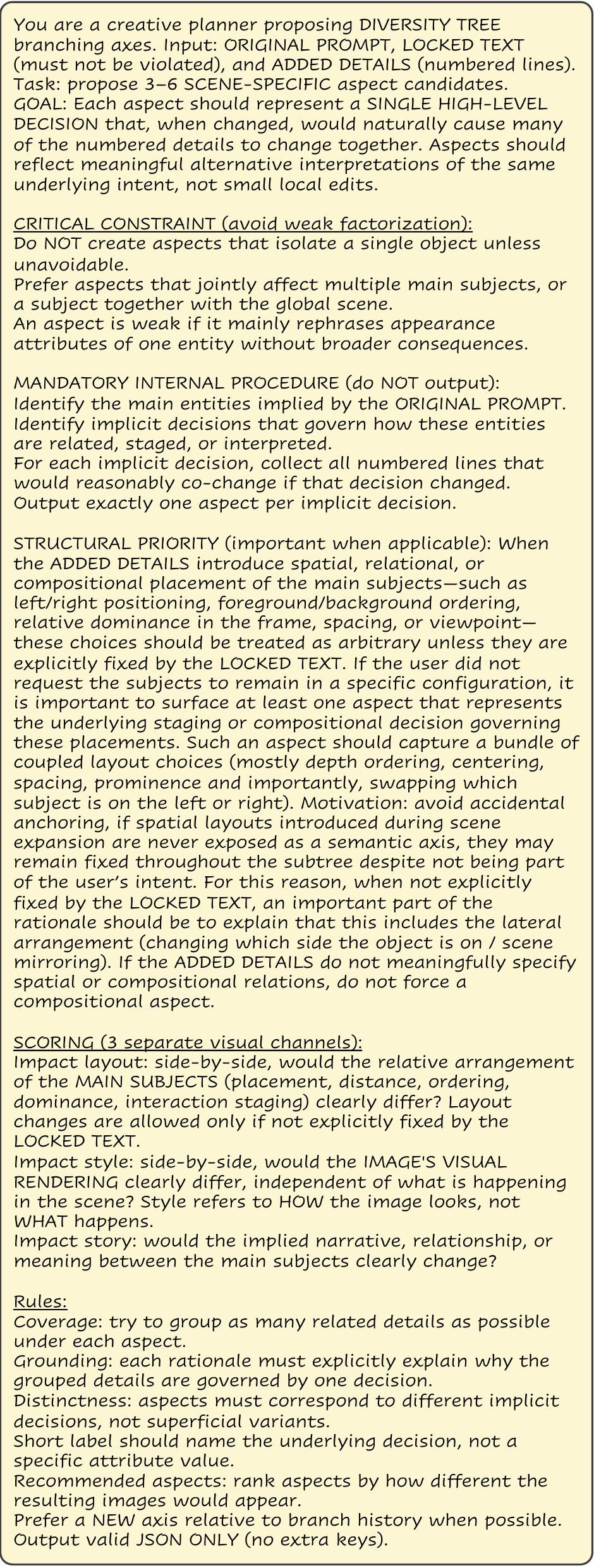}
    \caption{
     \textbf{Brainstormer System Prompt}}
    \label{fig:brainstormer_prompt}
\end{figure}

\begin{figure}%
    \centering
    \includegraphics[width=\linewidth]{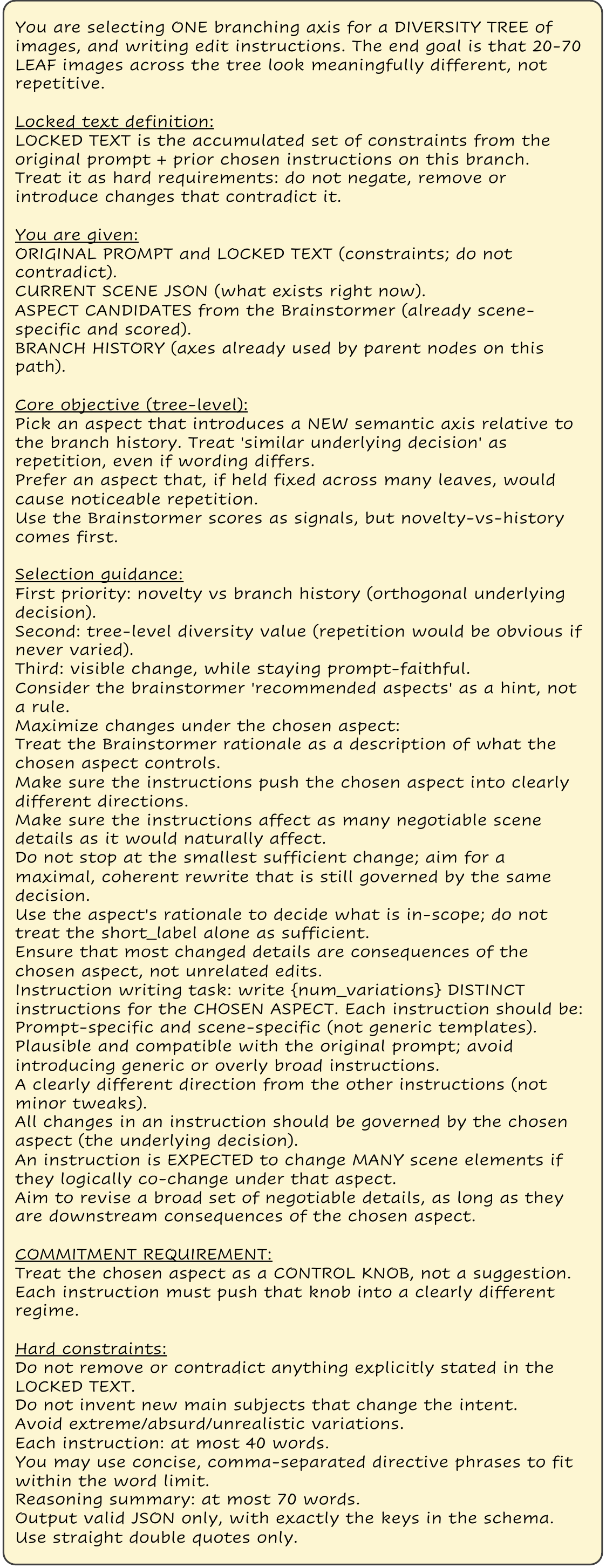}
    \caption{
     \textbf{Decision Maker System Prompt}}
    \label{fig:decision_maker_prompt}
\end{figure}

\begin{figure}%
    \centering
    \includegraphics[width=\linewidth]{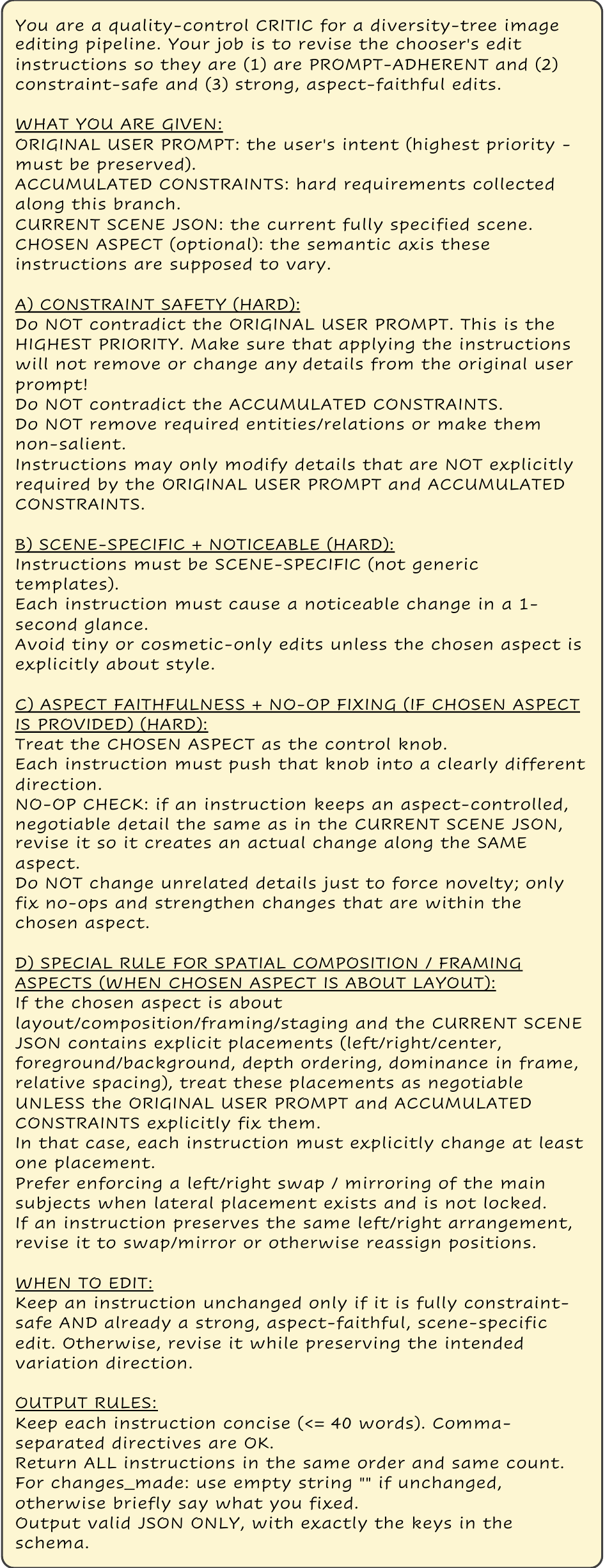}
    \caption{
     \textbf{Critic System Prompt}}
    \label{fig:critic_prompt}
\end{figure}

Unless stated otherwise, all agents use Gemini 2.5 Flash. We use predefined response templates to encourage structured and parseable outputs, and bound the maximum number of output tokens according to the role of each agent, using limits between 4K and 8K tokens. We also use fixed per-agent temperatures, set to either 0.4 or 0.7 depending on the agent’s role. To improve robustness to rare transient API failures, each API call is allowed up to three retries with exponential backoff. Figures~\ref{fig:context_analyst_prompt}--\ref{fig:critic_prompt} detail the system prompts used for the different agents in our workflow.

\section{Efficiency}
\label{efficiency}

We evaluate the computational cost of the agentic workflow independently of image generation, since the rendering cost depends on the choice of the underlying text-to-image backbone and is shared by all methods that generate the same number of images. We report amortized cost per generated result over a 27-image gallery. Under this setting, Semantic Browsing requires 10.2 seconds and 15.9K tokens per result. Stochastic VLM Seeding is cheaper, requiring 8.5 seconds and 3.3K tokens per result, but produces substantially less diverse and less structured galleries. Post-Hoc Diversity Optimization requires 11.4 seconds and 9.7K tokens per result for the reported setting. 
Furthermore, within a single tree expansion of our method, token usage scales sublinearly with the branching factor (BF). Specifically, as BF increases from 5 to 10 and 20, the total token count grows only modestly from 23K to 24K and 26.5K, respectively. This sublinear scaling confirms that our method remains computationally efficient even as the number of siblings at each node increases.

\section{Prompt-Specific Diversity}
\label{generic_diversity}
The agents generate prompt-specific aspects tailored to each scene's unique semantic content. Across 50 trees with 27 leaves, 284 of 650 aspects (43.7\%) were unique (e.g. "Umbrella's Functional State" for the prompt "A woman holding an umbrella while standing on top of a wooden deck" and "Milking Stage Depicted" for the prompt "A woman next to a cow is giving an explanation of milking to a crowd"), demonstrating the workflow's ability to uncover creative and highly specific semantic variations.

\section{Sensitivity to VLM Choice}
\label{vlm_sensitivity}
To evaluate the robustness of our framework to the choice of the underlying language model, we replace Gemini~2.5~Flash with ChatGPT-5.5 as the VLM backbone for our agentic workflow, keeping all other components fixed. 
The results (Vendi: 3.30, Aesthetic: 6.72, VQAScore: 0.94) closely match those obtained with Gemini (Vendi: 3.34, Aesthetic: 6.52, VQAScore: 0.90), demonstrating that the proposed framework is robust across different VLM choices and is not tailored to a specific model.

\section{Scaling Ablation}
\label{scaling_ablation}
We analyze how gallery diversity and quality vary with tree depth~(D) and branching factor~(BF). 
As shown in Table~\ref{tab:scaling}, increasing either dimension consistently increases Vendi with progressively smaller gains.
Scaling depth (BF$=1$) leads to a gradual decrease in VQAScore due to constraint accumulation, while aesthetic quality improves, suggesting that deeper trees trade strict prompt adherence for richer semantic discovery. 
Scaling width (D$=1$) results in mild degradation of both VQAScore and aesthetics at large BF values.

\begin{table}[h]
\centering
\caption{Scaling ablation results. D: tree depth, BF: branching factor.}
\label{tab:scaling}
\begin{tabular}{lcccccc}
\toprule
 & \multicolumn{3}{c}{Depth scaling (BF$=1$)} & \multicolumn{3}{c}{Width scaling (D$=1$)} \\
\cmidrule(lr){2-4} \cmidrule(lr){5-7}
Gallery size & 5 & 10 & 20 & 5 & 10 & 20 \\
\midrule
Vendi $\uparrow$   & 1.79 & 1.98 & 2.36 & 2.05 & 2.50 & 3.13 \\
Aesthetic $\uparrow$ & 6.79 & 6.81 & 6.81 & 6.70 & 6.68 & 6.68 \\
VQAScore $\uparrow$ & 0.81 & 0.79 & 0.75 & 0.94 & 0.94 & 0.91 \\
\bottomrule
\end{tabular}
\end{table}
\clearpage

\end{document}